\documentclass[preprint,12pt]{elsarticle}

\usepackage{amssymb}
\usepackage{amsmath}

\usepackage{algorithm}
\usepackage{algorithmic}
\usepackage{setspace}
\usepackage{bm}
\usepackage{pifont}

\usepackage{hyperref} 

\usepackage{threeparttable}
\usepackage{booktabs} 
\usepackage{float} 
\usepackage{multirow} 

\usepackage[utf8]{inputenc}  
\usepackage[final,tracking=true,kerning=true,spacing=true]{microtype} 

\graphicspath{{Fig/}} 
\usepackage{graphicx}

\journal{Machine Learning with Applications} 

\begin{document}

\begin{frontmatter}



\title{Statistical-Neural Interaction Networks for Interpretable Mixed-Type Data Imputation}


\author[label1]{Ou Deng\corref{cor1}} 
\ead{dengou@toki.waseda.jp}

\affiliation[label1]{organization={Graduate School of Human Sciences, Waseda University},
            addressline={2-579-15 Mikajima}, 
            city={Tokorozawa},
            postcode={359-1192}, 
            state={Saitama},
            country={Japan}}

\author[label2]{Shoji Nishimura}
\author[label2]{Atsushi Ogihara}
\author[label2]{Qun Jin\corref{cor1}} 
\ead{jin@waseda.jp}

\affiliation[label2]{organization={Faculty of Human Sciences, Waseda University},
            addressline={2-579-15 Mikajima}, 
            city={Tokorozawa},
            postcode={359-1192}, 
            state={Saitama},
            country={Japan}}

\cortext[cor1]{Corresponding authors}


\begin{abstract}
Real-world tabular databases routinely combine continuous measurements and categorical records, yet missing entries are pervasive and can distort downstream analysis. 
We propose Statistical--Neural Interaction (SNI), an interpretable mixed-type imputation framework that couples correlation-derived statistical priors with neural feature attention through a Controllable-Prior Feature Attention (CPFA) module. 
CPFA learns head-wise prior-strength coefficients $\{\lambda_h\}$ that softly regularize attention toward the prior while allowing data-driven deviations when nonlinear patterns appear to be present in the data. 
Beyond imputation, SNI aggregates attention maps into a directed feature-dependency matrix that summarizes which variables the imputer relied on, without requiring post-hoc explainers.
We evaluate SNI against six baselines (Mean/Mode, MICE, KNN, MissForest, GAIN, MIWAE) on six datasets spanning ICU monitoring, population surveys, socio-economic statistics, and engineering applications. 
Under MCAR/strict-MAR at 30\% missingness, SNI is generally competitive on continuous metrics but is often outperformed by accuracy-first baselines (MissForest, MIWAE) on categorical variables; in return, it provides intrinsic dependency diagnostics and explicit statistical--neural trade-off parameters. 
We additionally report MNAR stress tests (with a mask-aware variant) and discuss computational cost, limitations---particularly for severely imbalanced categorical targets---and deployment scenarios where interpretability may justify the trade-off.
\end{abstract}

\begin{graphicalabstract}
\includegraphics[width=1.0\textwidth]{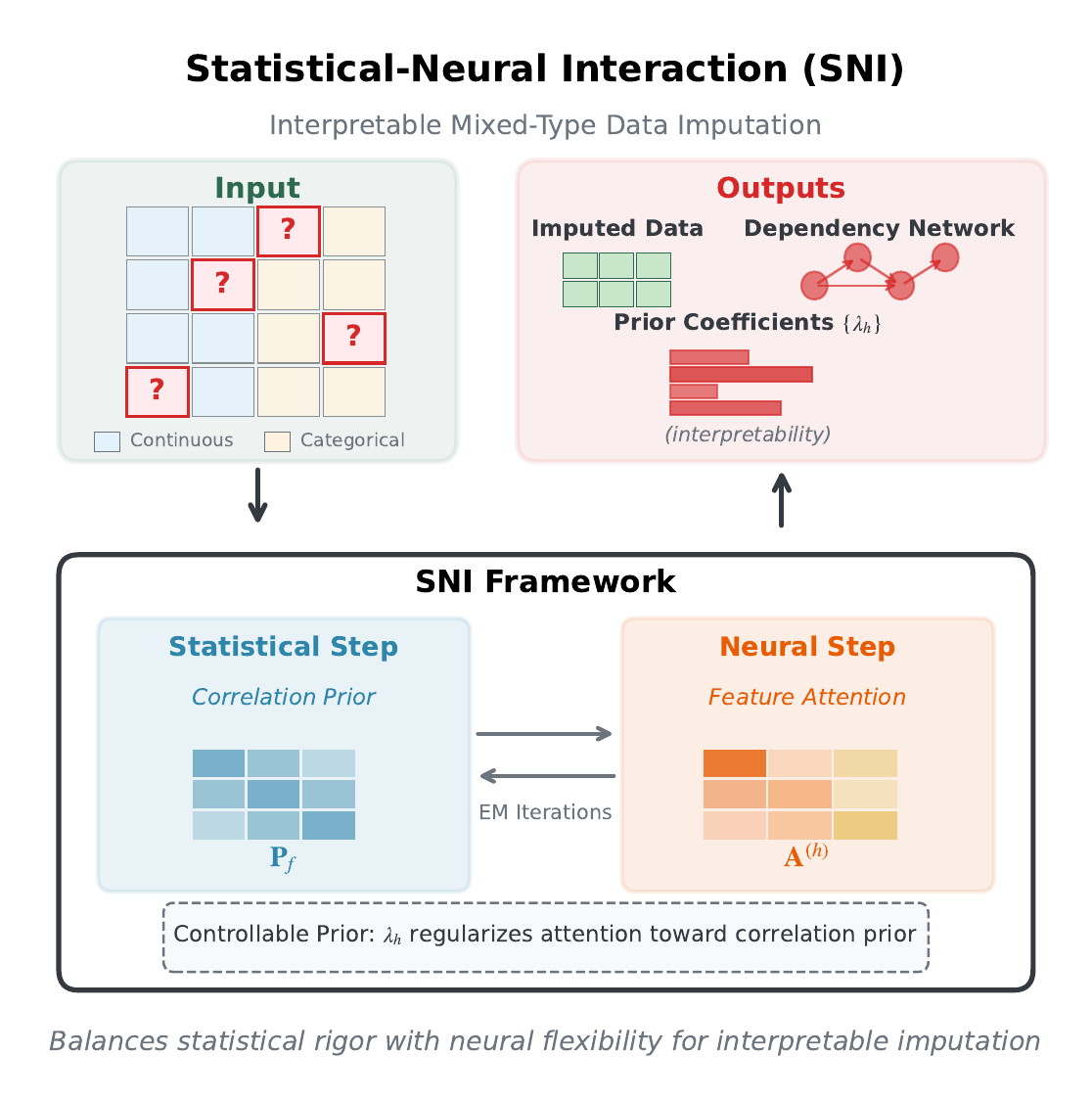} 
\end{graphicalabstract}

\begin{highlights}
\item Statistical–Neural Interaction couples correlation priors with attention for imputation
\item Learnable head-wise coefficients quantify how strongly attention follows the prior
\item SNI outputs a directed dependency network as intrinsic diagnostics without post-hoc tools
\item Benchmarks reveal accuracy–interpretability trade-offs on clinical and engineering data
\end{highlights}

\begin{keyword}
Mixed-type data \sep Imputation \sep Interpretable machine learning \sep Healthcare AI \sep Industrial data science
\end{keyword}

\end{frontmatter}


\section{Introduction}
\label{sec:intro}

Missing data pervade clinical registries, population surveys, and industrial sensor logs. Classical theory offers rigorous tools, notably the EM algorithm under Missing At Random (MAR)~\cite{Rubin1976,Dempster1977}, but linear models fail when variables interact non-linearly. Deep neural networks capture such interactions, yet recent surveys document their inconsistent tabular performance and opacity~\cite{Borisov2024,Tjoa2021}. This tension between {statistical traceability} and {representational capacity} critically impedes practical imputation in high-stakes domains.

\begin{figure}[t]
\centering
\includegraphics[width=\linewidth]{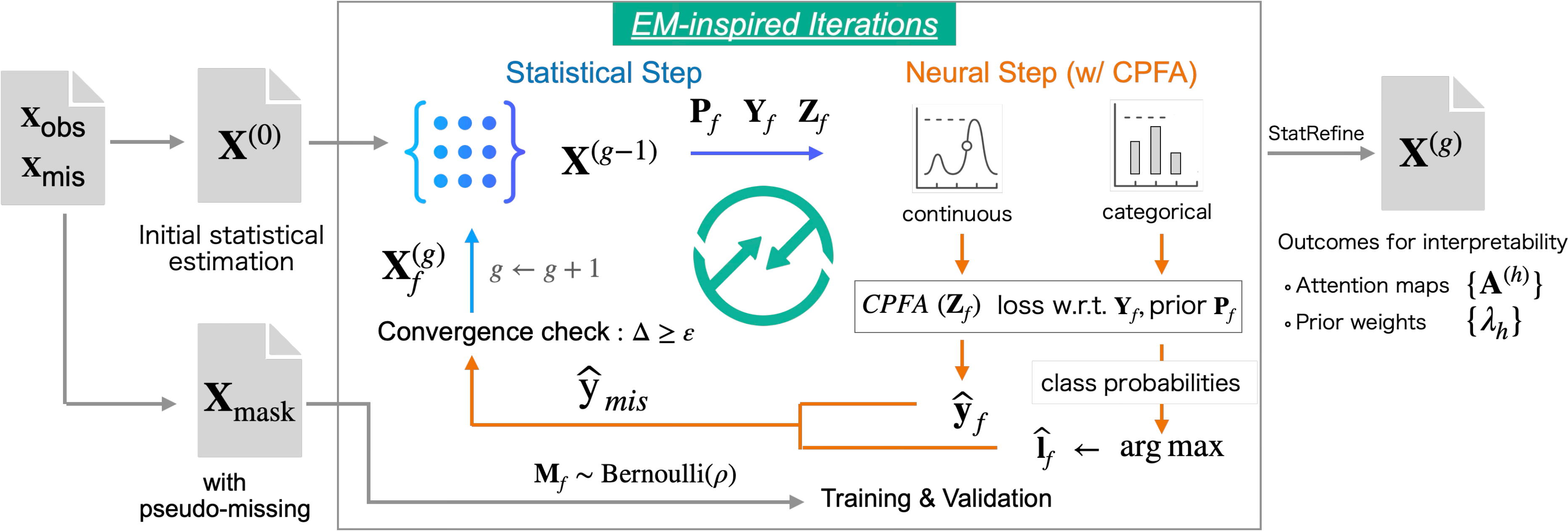}
\caption{Schematic of the Statistical-Neural Interaction (SNI) workflow. Grey icons show the observed/missing split $(\mathbf{X}_{\mathrm{obs}},\mathbf{X}_{\mathrm{mis}})$ and initial estimate $\mathbf{X}^{(0)}$. Each EM-inspired iteration comprises: \textbf{(i)} a {Statistical step} (blue) computing prior $\mathbf{P}_f$ from $\boldsymbol{\Sigma}^{(g-1)}$; \textbf{(ii)} {Pseudo-masking} via $\mathbf{M}_f\!\sim\!\mathrm{Bernoulli}(\rho)$; and \textbf{(iii)} a {Neural step} (orange) where CPFA regresses continuous targets or classifies categorical ones. Iterations terminate when relative change $\Delta<\varepsilon$. Outputs include the completed matrix $\mathbf{X}^{(g)}$, attention maps $\{\mathbf{A}^{(h)}\}$, and prior-confidence weights $\{\lambda_h\}$.}
\label{fig:SNI_architecture}
\end{figure}

\begin{figure}[t]
\centering
\includegraphics[width=0.55\textwidth]{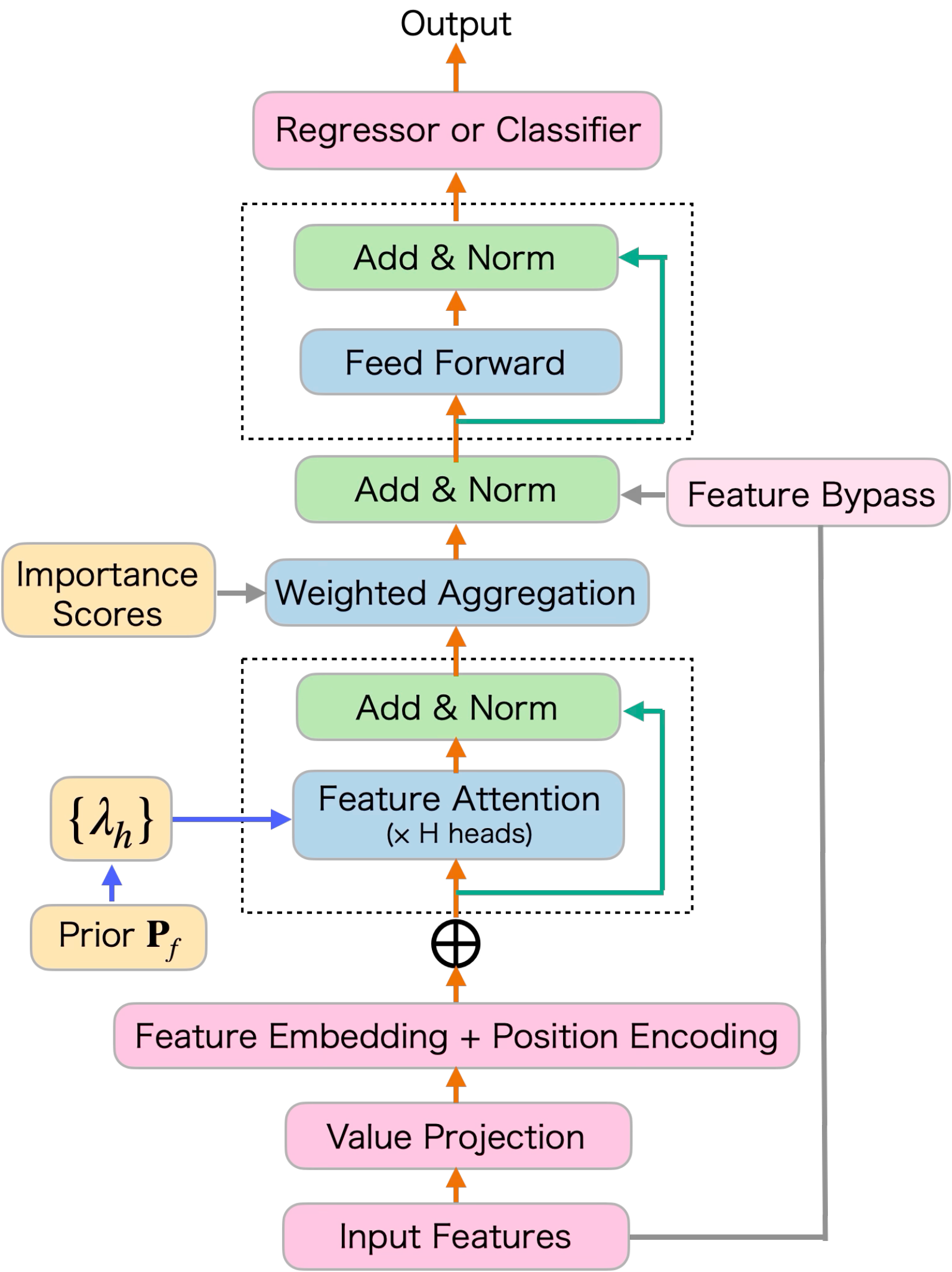} 
\caption{Architecture of the Controllable-Prior Feature Attention (CPFA) module. Input features are value-projected and position-encoded (pink) before multi-head feature attention (blue), where each head $h$ produces weights $\mathbf{A}^{(h)}$ regularized toward prior $\mathbf{P}_f$ via $\lambda_h=\mathrm{softplus}(\theta_{\lambda,h})$. The forward path (orange) passes through residual Add~\&~Norm and Feed-Forward layers (green), then weighted aggregation feeds the final Predictor. Grey arrows denote auxiliary components: importance-score extraction and a feature bypass preserving linear information. See Algorithm~\ref{alg:CPFA} for formal definition.}
\label{fig:CPFA}
\end{figure}

We address this dichotomy with \textbf{Statistical-Neural Interaction} (SNI), a method designed to balance classical inference with neural representation learning, shown in Figure~\ref{fig:SNI_architecture} and Algorithm~\ref{alg:SNI_MainFlow}. Central to SNI is \textbf{Controllable-Prior Feature Attention} (CPFA), which introduces learnable confidence coefficients mediating between empirical correlations and data-driven patterns, shown in Figure~\ref{fig:CPFA} and Algorithm~\ref{alg:CPFA}. For each feature $f$, CPFA receives observed covariates $\mathbf{Z}_f$ and correlation-derived prior $\mathbf{P}_f$. Soft-plus transformation yields non-negative coefficients $\lambda_h = \mathrm{softplus}(\theta_{\lambda,h})$ that penalize head-level attention deviations from $\mathbf{P}_f$. Large $\lambda_h$ encourages adherence to linear correlations; small $\lambda_h$ allows for higher-order modeling. Through EM-inspired iterations recomputing $\mathbf{P}_f$, statistical priors and neural attention co-evolve.

Specifically, for continuous features, CPFA regresses the standardized targets $\mathbf{y}_f$ from inputs $\mathbf{Z}_f$ under the prior $\mathbf{P}_f$, yielding predictions $\widehat{\mathbf{y}}_{f}$; for categorical features, CPFA produces class-probability vectors converted to discrete labels by $\arg\max$. The imputed values $\widehat{\mathbf{y}}_{\mathrm{mis}}$ replace the masked entries to form $\mathbf{X}^{(g)}$, optionally refined by a statistical post-processing step (\textbf{StatRefine}).

SNI is designed to address mixed-type imputation with an explicit accuracy--interpretability trade-off. 
It provides three practical outputs: 
(i) competitive imputation quality on continuous variables across heterogeneous domains without per-dataset tuning, 
(ii) intrinsic diagnostics---a directed dependency matrix/network and head-wise prior-strength coefficients $\{\lambda_h\}$---that summarize which columns the imputer relied on, and 
(iii) a controllable statistical--neural interaction mechanism that makes the influence of correlation priors explicit.
We additionally report MNAR experiments as {stress tests} (Supplementary Tables~S13--S19) and evaluate a mask-aware variant (SNI-M) as a pragmatic mitigation; these experiments are not intended to claim identifiability under arbitrary MNAR mechanisms.

We evaluate these aspects through validation on six real-world datasets under MCAR, MAR, and MNAR missingness mechanisms, benchmarking against widely used classical and deep baselines (Mean/Mode, KNN, MICE, MissForest, MIWAE, and GAIN). Across settings, SNI provides competitive continuous imputation quality while exposing a directed feature-dependency structure through its controllable-prior feature attention (CPFA), aiming to support both imputation quality and interpretable dependency analysis.

This work presents three contributions: (i) we introduce SNI, a controllable-prior feature-attention model for mixed-type imputation that seeks to balance data-driven attention with structured priors via head-wise learnable coefficients; (ii) we propose an attention-to-dependency mapping that aggregates controllable-prior feature attention into a directed dependency matrix and network, which may facilitate interpretability without requiring post-hoc explainers; (iii) we present a comprehensive evaluation on six real-world datasets under MCAR/MAR/MNAR regimes (with ablations and runtime analysis), suggesting that SNI can be competitive with strong baselines on continuous variables while providing explicit dependency insights.

The paper presents the method (Section~\ref{sec:Methodology}), comprehensive experiments (Section~\ref{sec:Experiment}), theoretical implications (Section~\ref{sec:discussion}), and conclusions with future directions (Section~\ref{sec:conclusion}).

\section{Related Work}
\label{sec:Related}

\textbf{Statistical foundations of missing-data handling.}
The taxonomy of data missing mechanisms---MCAR, MAR, and MNAR (Missing Completely at Random; Missing at Random; Missing Not at Random)---originates from~\cite{Rubin1976} and the subsequent monographs of Rubin and Little~\cite{Rubin1987,Little2019}. Likelihood-based EM inference~\cite{Dempster1977} and its refinements~\cite{Schafer1997,LittleRubin2019} remain the cornerstone of principled imputation. In applied disciplines, particularly within clinical and psychological research~\cite{Enders2017,Janssen2010}, multiple-imputation engines such as MICE~\cite{vanBuuren2018,Raghunathan2001,Azur2011} and its practical guidelines~\cite{White2011,Dong2013} have become widely adopted. While classical methods offer rigorous theoretical guarantees~\cite{Schafer2002}, they often struggle with high-dimensional interactions. Tree ensembles (e.g., MissForest~\cite{Stekhoven2011}) improve non-linear accuracy but retain the ``black-box'' opacity of modern learners. Because our variational view is derived under MAR, we treat MNAR evaluations as stress tests and do not claim identifiability under arbitrary MNAR mechanisms. We nevertheless report MNAR results and a mask-aware variant to probe sensitivity to distribution shifts (Supplementary Tables~S13--S19).

\textbf{Deep imputation models and tabular DNNs.}
Recent years have seen a surge in neural imputation strategies. Generative imputers---GAN (GAIN~\cite{Yoon2018}), VAE (MIWAE~\cite{Mattei2019}, HI-VAE~\cite{Nazabal2020}), and auto-encoder variants (MIDA~\cite{Gondara2018}, Denoising AE~\cite{Beaulieu-Jones2017})---push reconstruction error lower but often sacrifice statistical traceability. In the domain of temporal data, recurrent architectures like BRITS~\cite{Cao2018} and GRU-D~\cite{Che2016}, along with GAN-based time-series approaches~\cite{Luo2018}, have set precedents for utilizing missingness patterns directly.
For tabular data specifically, scalable approaches such as MIDAS~\cite{Lall2021} and recent end-to-end frameworks like MFAN~\cite{Chang2025} demonstrate the potential of deep learning, yet a recent survey underscores the tension between expressive power and tabular over-fitting~\cite{Borisov2024}. Moreover, specific architectures for tabular data, such as TabTransformer~\cite{Huang2020} and TabNet~\cite{Arik2021}, have begun to challenge tree-based dominance. SNI attempts to bridge these paradigms by anchoring its attention weights in a closed-form correlation prior, seeking to combine the expressiveness of deep tabular learners with elements of classical statistical reasoning.

\textbf{Attention mechanisms, interpretability and XAI.}
Since the introduction of the Transformer~\cite{Vaswani2017}, attention mechanisms have revolutionized representation learning. In the tabular domain, models like FT-Transformer~\cite{Kossen2021} and SAINT~\cite{Pilaluisa2022} have flourished~\cite{Borisov2024,Tay2022}, yet most remain purely data-driven. Explainable-AI (XAI) surveys, particularly in medical domains~\cite{Xiao2018,Fridgeirsson2023}, emphasize the urgent need for \textit{intrinsic} rather than post-hoc explanations~\cite{Wiegreffe2019,Molnar2020,Tjoa2021}. Rudin~\cite{Rudin2019} notably argues for models that are interpretable by design for high-stakes decisions. Furthermore, debates regarding whether ``attention is explanation''~\cite{Serrano2019,Abnar2020} highlight the necessity of structured constraints. By constraining each head via $\mathbf P_f$ and exposing the temperature-like $\lambda_h$, CPFA aims to provide head-level explanations that address some of these XAI desiderata without relying on external attribution methods.

\textbf{Posterior regularization and soft constraints.}
Posterior-regularization (PR) provides a framework for injecting structured domain knowledge into differentiable models~\cite{Ganchev2010,Hu2018}. SNI instantiates PR in tabular attention: the soft-plus mapping $\lambda_h=\text{softplus}(\theta_{\lambda,h})$ balances a PR term with the CPFA reconstruction loss. This approach echoes constraint-driven vision transformers~\cite{Li2021} and physics-guided supervision~\cite{Stewart2017}. The mechanism can be viewed as a deterministic-annealing schedule~\cite{Rose1998,Ueda1998} that gradually relaxes statistical rigidity, a strategy long exploited in curriculum learning~\cite{Bengio2009} and optimization~\cite{Li2024}.

\textbf{Sparse Bayesian perspectives and head specialization.}
The ability of CPFA to deactivate redundant heads resonates with sparse Bayesian learning (ARD~\cite{Tipping2001}) and global-local shrinkage priors such as the Horseshoe~\cite{Carvalho2009,Piironen2017,Bhadra2019}. Empirical analyses of language transformers show that many heads can be pruned with negligible loss~\cite{Michel2019,Voita2019}; CPFA operates similarly but under an explicit, data-dependent prior rather than a purely sparsity-inducing one. This connects to the ``product of experts" philosophy~\cite{Hinton1999}, where individual heads (experts) specialize in distinct dependency patterns.

\textbf{Self-supervised denoising and pseudo-masking.}
Our pseudo-mask scheme follows the denoising-auto-encoder paradigm~\cite{Vincent2008,Bengio2013} and its tabular variant MIDA~\cite{Gondara2018}, but preserves MAR identifiability by re-drawing the mask each EM cycle. Score-matching links between denoising and likelihood estimation~\cite{Alain2014} suggest a potential connection between CPFA's reconstruction objective and missing-data score estimation, paralleling recent work on masked language modeling~\cite{Devlin2019}.

\textbf{Complementary threads.}
Hybrid gradient and gradient-free optimization~\cite{Majid2024} offers an intriguing route to further tune CPFA's discrete hyper-parameters via evolutionary search. Additionally, addressing class imbalance in missing data scenarios~\cite{He2009} remains a critical frontier. While efficient attention mechanisms~\cite{Tay2022} motivate future sparse CPFA variants for ultra-wide registries, current work focuses on establishing the statistical-neural synergy.

Collectively, these strands illustrate how SNI attempts to connect classical statistical principles with the representational capacity---and explainability considerations---of contemporary tabular deep learning.

\section{Methodology}  
\label{sec:Methodology}

\subsection{Statistical-Neural Interaction Method}
\label{subsec:SNI_framework}

The SNI method seeks to navigate the tension between theoretically grounded statistical imputation and flexible neural modeling. Let $\mathbf{X}\in\mathbb{R}^{n\times d}$ be the incomplete data matrix, partitioned into continuous features $\mathcal{C}$ and categorical features $\mathcal{K}$. For correlation computation, we build a correlation-design matrix where continuous columns are standardized and categorical columns are one-hot encoded; Pearson correlation on this representation yields a unified association matrix used only to construct priors. CPFA itself operates on the original feature set, using regression for continuous targets and classification for categorical targets. We denote this correlation-design matrix by $\tilde{\mathbf{X}}$.

Initialization employs a standard statistical imputation (e.g., MICE) to obtain $\mathbf{X}^{(0)}=\mathrm{StatImpute}(\mathbf{X})$. At each iteration $g$, we compute

\begin{equation}
\boldsymbol{\Sigma}^{(g-1)} = \mathrm{corr}\bigl(\tilde{\mathbf{X}}^{(g-1)}_{\mathcal{I}_{\mathrm{tr}}}\bigr),
\end{equation}

optionally applying Fisher's $z$-transformation for variance stabilization. For each feature $f$ with missing entries, we extract

\begin{equation}
\mathbf{y}_f = \mathbf{X}^{(g-1)}_{:,f}, \quad \mathbf{Z}_f = \mathbf{X}^{(g-1)}_{:,\neg f},
\end{equation}

and construct a feature-specific prior $\mathbf{P}_f$ as:

\begin{equation}
\label{eq:p_f}
   \mathbf{P}_f 
   = \mathrm{normalize}\!\left(\left[\mathrm{aggregate}\Bigl(\bigl|\boldsymbol{\Sigma}^{(g-1)}_{:,\mathcal{J}_f}\bigr|\Bigr)\right]_{\neg f}\right),
\end{equation}

where $\mathcal{J}_f$ indexes the correlation-design columns associated with feature $f$ (a singleton for continuous features and the one-hot group for categorical features), $\mathrm{aggregate}(\cdot)$ collapses one-hot groups into per-feature scores (mean absolute correlation in our implementation), $[\cdot]_{\neg f}$ removes the target-feature entry, and $\mathrm{normalize}(\cdot)$ rescales the resulting nonnegative vector to sum to one (implementation details: Supplementary Section~S2.5). A Bernoulli mask $\mathbf{M}_f\sim\mathrm{Bernoulli}(\rho)$ is drawn over $\mathcal{I}_{\mathrm{tr}}$ to generate pseudo-missing entries for self-supervision.

For each attention head $h$ we introduce a non-negative prior-confidence coefficient:

\begin{equation}
\label{eq:lambda_def}
   \lambda_h = \operatorname{softplus}(\theta_{\lambda,h}),
  \qquad\lambda_h\in(0,\infty).   
\end{equation}

Large $\lambda_h$ forces the head-level mean attention $\overline{\mathbf A}^{(h)}$ to stay close to the statistical prior $\mathbf P_f$, whereas small values allow the head to deviate and capture non-linear interactions; more details are provided in Equation~\eqref{eq:L_prior}.

For continuous features ($f\in\mathcal{C}$), both the target values $\mathbf{y}_f$ and input features $\mathbf{Z}_f$ are standardized using statistics computed from observed data only. The CPFA subnetwork for regression is then trained by:

\begin{equation}
\label{eq:theta_f_reg}
\begin{split}
\theta_{f}^{\text{(reg)}} = \arg\min_{\theta} \bigg[ &\frac{1}{|\mathcal{I}_{\mathrm{train}}|}\sum_{i\in\mathcal{I}_{\mathrm{train}}} \bigl(y_{\mathrm{standardized},i} - \mathrm{CPFA}(\mathbf{Z}_f)_i\bigr)^2 \\
&+ \alpha^{(g-1)}\sum_{h=1}^H\lambda_h\bigl\|\overline{A}^{(h)} - \mathbf{P}_f\bigr\|_2^2 \bigg]
\end{split}
\end{equation}

where $\{\lambda_h,\overline{A}^{(h)}\}_{h=1}^H$ are the CPFA attention parameters controlling prior regularization. Missing values are imputed by applying CPFA to standardized inputs and then inverse-transforming the predictions back to the original scale.

For categorical features ($f\in\mathcal{K}$), the categorical values are label-encoded to integer labels $\mathbf{l}_f$ for training the classification CPFA:

\begin{equation}
\label{eq:theta_f_clf}
\begin{split}
\theta_{f}^{\text{(clf)}} = \arg\min_{\theta} \bigg[ &\frac{1}{|\mathcal{I}_{\mathrm{train}}|}\sum_{i\in\mathcal{I}_{\mathrm{train}}} -\log \Pr\Bigl(l_{f,i}\mid \mathrm{CPFA}(\mathbf{Z}_f)_i\Bigr) \\
&+ \alpha^{(g-1)}\sum_{h=1}^H\lambda_h\bigl\|\overline{A}^{(h)} - \mathbf{P}_f\bigr\|_2^2 \bigg]
\end{split}
\end{equation}

In experiments we use a focal-loss variant of cross-entropy (with $\gamma=2$) for categorical targets to partially mitigate severe class imbalance (Supplementary Table~S4).

Then $\arg\max\,\mathrm{CPFA}(\mathbf{Z}_f[\mathrm{mis}];\theta_f^{(\mathrm{clf})})$ performs the imputation.

After all features are imputed, we refine $\mathbf{X}^{(g)} = \mathrm{StatRefine}\bigl(\mathbf{X}^{(g-1)}\bigr)$, and check convergence via $\|\mathbf{X}^{(g)} - \mathbf{X}^{(g-1)}\|_F / \|\mathbf{X}^{(g-1)}\|_F < \varepsilon$.

The variational derivation of our EM procedure and the hierarchical Bayesian prior placed on the confidence coefficients $\lambda_h$ are provided in the Supplementary Material.

\begin{algorithm}[h]
\small
\caption{Statistical–Neural Interaction (SNI) Imputation Scheme}
\label{alg:SNI_MainFlow}
\begin{algorithmic}[1]
\REQUIRE Incomplete data $\mathbf{X} \in \mathbb{R}^{n\times d}$; Continuous $\mathcal{C}$ \& Categorical $\mathcal{K}$ indices; \\
         \hspace{1.2em} Mask ratio $\rho$; Threshold $\varepsilon$; Max iterations $G$.
\ENSURE Imputed matrix $\widehat{\mathbf{X}}$.

\STATE \textbf{Partition} $\mathcal{I} = \mathcal{I}_{\mathrm{tr}} \cup \mathcal{I}_{\mathrm{val}} \cup \mathcal{I}_{\mathrm{test}}$
\STATE \textbf{Initialize} $\mathbf{X}^{(0)} \leftarrow \mathrm{MeanModeImpute}(\mathbf{X})$
\REPEAT
    \STATE $g \leftarrow g + 1$
    \STATE Update correlation $\boldsymbol{\Sigma}^{(g-1)} \leftarrow \mathrm{corr}(\tilde{\mathbf{X}}^{(g-1)}_{\mathcal{I}_{\mathrm{tr}}})$
    
    \FOR{each feature $f \in \mathcal{C} \cup \mathcal{K}$ with missing entries}
        \STATE Set target $\mathbf{y}_f = \mathbf{X}^{(g-1)}_{:,f}$ and predictors $\mathbf{Z}_f = \mathbf{X}^{(g-1)}_{:,\neg f}$
        \STATE Compute prior $\mathbf{P}_f$ using Eq.~\eqref{eq:p_f} based on $\boldsymbol{\Sigma}^{(g-1)}$
        \STATE Sample mask $\mathbf{M}_f \sim \mathrm{Bernoulli}(\rho)$ on $\mathcal{I}_{\mathrm{tr}}$
        
        \IF{$f \in \mathcal{C}$}
            \STATE $\theta_f^* \leftarrow \arg\min_{\theta} \mathcal{L}_{\mathrm{reg}}\bigl(\mathbf{Z}_f[\mathrm{obs}], \mathbf{y}_f[\mathrm{obs}]; \theta \bigr)$
            \STATE $\mathbf{X}^{(g)}[\mathrm{mis}, f] \leftarrow \mathrm{CPFA}(\mathbf{Z}_f[\mathrm{mis}]; \theta_f^*)$
        \ELSE
            \STATE $\theta_f^* \leftarrow \arg\min_{\theta} \mathcal{L}_{\mathrm{clf}}\bigl(\mathbf{Z}_f[\mathrm{obs}], \mathrm{LabelEncode}(\mathbf{y}_f[\mathrm{obs}]); \theta \bigr)$
            \STATE $\mathbf{X}^{(g)}[\mathrm{mis}, f] \leftarrow \mathrm{Decode}\bigl( \arg\max \mathrm{CPFA}(\mathbf{Z}_f[\mathrm{mis}]; \theta_f^*) \bigr)$
        \ENDIF
    \ENDFOR
    
    \STATE $\mathbf{X}^{(g)} \leftarrow \mathrm{PostProcess}(\mathbf{X}^{(g)})$ \COMMENT{e.g., clip bounds}
\UNTIL{$ \|\mathbf{X}^{(g)} - \mathbf{X}^{(g-1)}\|_F / \|\mathbf{X}^{(g-1)}\|_F < \varepsilon$ \textbf{or} $g \ge G$}
\RETURN $\widehat{\mathbf{X}} = \mathbf{X}^{(g)}$
\end{algorithmic}
\end{algorithm}

\begin{algorithm}[h]
\small
\caption{Controllable-Prior Feature Attention (CPFA) Module}
\label{alg:CPFA}
\begin{algorithmic}[1]
\REQUIRE Predictors $\mathbf{Z} \in \mathbb{R}^{B \times (d-1)}$; Target $\mathbf{y} \in \mathbb{R}^{B}$; \\
         \hspace{1.2em} Prior vector $\mathbf{P} \in \mathbb{R}^{d-1}$; Regularizer weights $\{\alpha^{(t)}\}$.
\ENSURE Learned parameters $\theta$, Attention weights $\mathbf{A}$.

\FOR{epoch $t = 1, \dots, T$}
    \STATE \textbf{Forward Pass}:
    \FOR{head $h = 1, \dots, H$}
        \STATE Compute attention scores $\mathbf{A}^{(h)}$ via Eq.~\eqref{eq:A_h}
        \STATE Compute confidence $\lambda_h = \mathrm{softplus}(\theta_{\lambda, h})$ via Eq.~\eqref{eq:lambda_def}
    \ENDFOR
    
    \STATE \textbf{Loss Computation}:
    \STATE $\mathcal{L}_{\mathrm{recon}} \leftarrow \text{MSE}(\hat{\mathbf{y}}, \mathbf{y})$ \textbf{or} $\text{CrossEntropy}(\hat{\mathbf{y}}, \mathbf{y})$
    \STATE $\mathcal{L}_{\mathrm{prior}} \leftarrow \sum_{h} \alpha^{(t)} \cdot \lambda_h \bigl\|\overline{\mathbf{A}}^{(h)} - \mathbf{P}\bigr\|_2^2$
    
    \STATE \textbf{Update}: $\theta \leftarrow \theta - \eta \nabla_{\theta} (\mathcal{L}_{\mathrm{recon}} + \mathcal{L}_{\mathrm{prior}})$
\ENDFOR
\RETURN Parameters $\theta$ and aggregated attention $\mathbf{A}$
\end{algorithmic}
\end{algorithm}

\subsection{Controllable-Prior Feature Attention}

The CPFA mechanism integrates statistical priors $\mathbf{P}_f$ into a multi-head attention architecture. Given $\mathbf{Z}\in\mathbb{R}^{n\times(d-1)}$ and target $y$, CPFA computes $H$ attention maps as:

\begin{equation}
\label{eq:A_h}
   \mathbf{A}^{(h)} = \mathrm{Attention}\bigl(\mathbf{Z}\bigr),\quad h=1,\ldots,H.
\end{equation}

Each head has an associated confidence parameter $\lambda_h = \mathrm{softplus}(\theta_{\lambda,h})$, so that the average attention $\overline{\mathbf{A}}^{(h)}$ is regularized toward $\mathbf{P}_f$. Specifically, $\mathcal{L}_{\mathrm{prior}}$ is described as:

\begin{equation}
\label{eq:L_prior}
   \mathcal{L}_{\mathrm{prior}}
   = \alpha^{(g-1)} \sum_{h=1}^H \lambda_h \bigl\|\overline{\mathbf{A}}^{(h)} - \mathbf{P}_f\bigr\|_2^2,
\end{equation}

where $\overline{\mathbf{A}}^{(h)}\in\mathbb{R}^{d-1}$ is the mean attention vector for head $h$.

When $y_f=\mathbf{y}$ (standardized continuous values) or $y_f=\mathbf{l}$ (label-encoded categorical), the reconstruction loss is:

\begin{equation}
\label{eq:L_recon}
   \mathcal{L}_{\mathrm{recon}}
   = 
   \begin{cases}
     \frac{1}{n_{\mathrm{tr}}}\sum_{i\in\mathcal{I}_{\mathrm{tr}}}
     \bigl(\widehat{y}_i - y_i\bigr)^2, 
     & \text{if } y_f=\mathbf{y},\\[6pt]
     -\frac{1}{n_{\mathrm{tr}}}\sum_{i\in\mathcal{I}_{\mathrm{tr}}} 
     \log \bigl[\Pr\bigl(l_i \mid \widehat{l}_i\bigr)\bigr], 
     & \text{if } y_f=\mathbf{l}.
   \end{cases}
\end{equation}

The total loss is $\mathcal{L}_{\mathrm{total}} = \mathcal{L}_{\mathrm{recon}} + \mathcal{L}_{\mathrm{prior}}$.

For continuous features, the predictor directly outputs the imputed value $\widehat{\mathbf{y}}_{f}$, which is inverse-standardized to recover the original scale. For categorical features, the predictor outputs class-probability vectors that are converted to discrete labels by $\arg\max$. The feature bypass pathway enables raw embeddings to shortcut the deep attention blocks, preserving linear information that might otherwise be attenuated.

The learned $\lambda_h$ values quantify each head's reliance on linear correlations versus non-linear patterns. A large $\lambda_h$ indicates that $\overline{\mathbf{A}}^{(h)}\approx\mathbf{P}_f$, whereas a small $\lambda_h$ allows the head to deviate and capture complex interactions. This mechanism is intended to support interpretability: monitoring $\{\lambda_h\}$ may indicate which features are well explained by correlation and which may benefit from richer representations.

\subsection{Self-Supervised Pseudo-Missing Strategy}

For each feature $f$, a pseudo-missing mask $\mathbf{M}_f\sim\mathrm{Bernoulli}(\rho)$ is redrawn each iteration. This transforms imputation into a self-supervised reconstruction task: the model learns to predict masked entries from $\mathbf{Z}_f$. Because $\mathbf{M}_f$ is applied uniformly across $\mathcal{C}$ and $\mathcal{K}$, the training signal remains balanced even when missingness rates differ. The stochastic masking also acts as a form of data augmentation, which may help improve generalization while aiming to preserve statistical validity.

\subsection{Theoretical Foundations and Practical Implications}

By combining Fisher-transformed correlation priors with feature attention, SNI makes the statistical--neural interaction explicit. In the EM-inspired outer loop (Algorithm~\ref{alg:SNI_MainFlow}), priors are recomputed from the current completed data and CPFA parameters are updated by minimizing a reconstruction loss plus a prior regularizer (Section~\ref{sec:Methodology}). Because the neural step is optimized approximately by stochastic gradient descent, we do not claim monotonic improvement at every iteration; instead, we monitor relative changes and empirically observe generally stable convergence behavior across seeds and datasets in our experiments. For categorical targets we use a focal-loss variant to partially mitigate imbalance, although extreme class imbalance remains a limitation (Section~\ref{sec:discussion}).

For mixed-type data, continuous variables are standardized and categorical variables are handled by feature embeddings and a classification head in CPFA, while the statistical prior is derived from correlations computed on a standardized/one-hot correlation-design matrix. This avoids imposing an artificial numerical ordering on categories when constructing priors. Training is performed in mini-batches and each predictor only conditions on the remaining $(d-1)$ features, which is compatible with moderate-to-large sample sizes; the main computational bottleneck is the feature-attention component. 

In summary, SNI approaches missing data imputation by embedding statistical priors into a neural architecture. Through direct handling of standardized continuous values, categorical encoding via embeddings, and controllable attention, the framework aims to balance theoretical grounding with practical applicability.

\section{Experiments}
\label{sec:Experiment}

\subsection{Cross-Domain Datasets and Missing Mechanism}
We consider six real-world tabular datasets that span medical and industrial/engineering domains: MIMIC-IV (ICU vitals and alarms), eICU (ICU physiology and interventions), NHANES (survey and laboratory measurements), Concrete Compressive Strength (materials engineering), AutoMPG (vehicle specifications and fuel economy), and Communities \& Crime (socio-economic indicators). The datasets contain mixed continuous and categorical variables (Concrete is fully continuous). We simulate missingness using three mechanisms: MCAR, strict MAR, and MNAR. Unless otherwise stated, the main manuscript reports MCAR/MAR with 30\% missingness, repeating each configuration over five seeds and reporting mean $\pm$ SD. To promote fair comparison, we use publicly available implementations and recommended hyperparameters for deep baselines (notably GAIN and MIWAE) and keep them fixed across datasets; details are in Supplementary Table~S5.

The Supplementary Material provides the complete benchmark tables and configurations: average ranks under MCAR/MAR (30\%) are summarized in Table~S6 with per-dataset results in Tables~S7--S12; MNAR stress tests are summarized in Table~S13 with per-dataset results in Tables~S14--S19; hyperparameters are listed in Tables~S4--S5; and ablations are reported in Table~S20.

\begin{table}[h]
\footnotesize
\centering
\caption{Continuous variable performance on MIMIC-IV ICU subset, MAR 30\% missingness. Values are mean $\pm$ SD over five seeds.\label{tab:mimic_MAR30per_cont}}
\begin{tabular}{lccccc}
\toprule
Method & NRMSE$\downarrow$ & MAE$\downarrow$ & MB$\downarrow$ & $R^2\uparrow$ & Spearman $\rho\uparrow$ \\
\midrule
SNI & 0.056$\pm$0.001 & 2.082$\pm$0.029 & 0.247$\pm$0.067 & 0.773$\pm$0.007 & 0.825$\pm$0.006 \\
MissForest & 0.045$\pm$0.000 & \textbf{1.633$\pm$0.010} & 0.036$\pm$0.016 & 0.851$\pm$0.003 & \textbf{0.880$\pm$0.002} \\
MIWAE & \textbf{0.042$\pm$0.000} & 1.691$\pm$0.023 & 0.126$\pm$0.115 & \textbf{0.856$\pm$0.004} & 0.879$\pm$0.002 \\
GAIN & 0.187$\pm$0.018 & 10.188$\pm$0.605 & -0.598$\pm$1.960 & -1.547$\pm$0.457 & 0.350$\pm$0.027 \\
KNN & 0.085$\pm$0.000 & 3.517$\pm$0.000 & \textbf{-0.018$\pm$0.000} & 0.537$\pm$0.000 & 0.714$\pm$0.000 \\
MICE & 0.095$\pm$0.002 & 3.569$\pm$0.045 & -0.045$\pm$0.142 & 0.312$\pm$0.023 & 0.649$\pm$0.014 \\
Mean/Mode & 0.129$\pm$0.000 & 6.722$\pm$0.000 & 0.975$\pm$0.000 & -0.027$\pm$0.000 & 0.000$\pm$0.000 \\
\bottomrule
\end{tabular}
\end{table}

\begin{table}[h]
\footnotesize
\centering
\caption{Categorical variable performance on MIMIC-IV ICU subset, MAR 30\% missingness. Values are mean $\pm$ SD over five seeds.\label{tab:mimic_MAR30per_cat}}
\begin{tabular}{lccc}
\toprule
Method & Accuracy$\uparrow$ & Macro-F$_1\uparrow$ & Cohen's $\kappa\uparrow$ \\
\midrule
SNI & 0.317$\pm$0.021 & 0.301$\pm$0.018 & 0.204$\pm$0.025 \\
MissForest & \textbf{0.836$\pm$0.003} & \textbf{0.539$\pm$0.015} & \textbf{0.723$\pm$0.005} \\
MIWAE & 0.820$\pm$0.011 & 0.505$\pm$0.019 & 0.698$\pm$0.018 \\
GAIN & 0.293$\pm$0.101 & 0.127$\pm$0.022 & 0.046$\pm$0.033 \\
KNN & 0.755$\pm$0.000 & 0.409$\pm$0.000 & 0.582$\pm$0.000 \\
MICE & 0.598$\pm$0.010 & 0.317$\pm$0.010 & 0.348$\pm$0.014 \\
Mean/Mode & 0.506$\pm$0.000 & 0.100$\pm$0.000 & 0.000$\pm$0.000 \\
\bottomrule
\end{tabular}
\end{table}

\subsection{Overall Results}
Tables~\ref{tab:mimic_MAR30per_cont}--\ref{tab:summary_additional} report the main MCAR/strict-MAR results at 30\% missingness, while the Supplementary Material provides the full MCAR/MAR/MNAR tables (Tables~S6--S20) including missing-rate sweeps and ablations.

\paragraph{Accuracy on continuous variables}
Across six datasets and two mechanisms (MCAR/MAR at 30\% missingness), MissForest is the strongest classic baseline on average, with MIWAE also competitive in multiple settings.
SNI is typically not the top performer, but it remains competitive on continuous metrics and is closest to MissForest on several settings (e.g., AutoMPG and ComCri under MAR, where the relative NRMSE gaps are -5.5\% and -5.9\%; Table~\ref{tab:summary_additional}). 
These results indicate that SNI should not be viewed as an accuracy-only replacement for MissForest/MIWAE, but as an interpretable alternative when understanding the imputation logic is valuable.

\paragraph{Categorical variables and class imbalance}
On categorical targets, especially under extreme imbalance (e.g., ICU alarms and discretized SpO$_2$ bins), nonparametric baselines (MissForest, KNN) tend to dominate.
SNI uses a focal-loss variant for categorical targets (Supplementary Table~S4), yet substantial gaps remain on the most imbalanced datasets (e.g., eICU; Supplementary Tables~S8 and~S19). 
We therefore position categorical imputation under severe imbalance as an important limitation and discuss practical workarounds and future improvements in Section~\ref{sec:discussion}.

\paragraph{GAIN performance}
GAIN~\cite{Yoon2018} is included for completeness as a representative GAN-based imputer.
In our mixed-type benchmarks (especially those with discrete and/or highly imbalanced targets under strict-MAR/MNAR), its training can be sensitive and in some settings it underperforms even simple imputers (Mean/Mode).
To avoid over-interpreting a potentially unstable baseline, we do not rely on GAIN for any main conclusion; we report its fixed hyperparameters (Supplementary Table~S5) and full per-dataset results in the Supplementary Material.

\begin{figure}[t]
\centering
\includegraphics[width=0.95\textwidth]{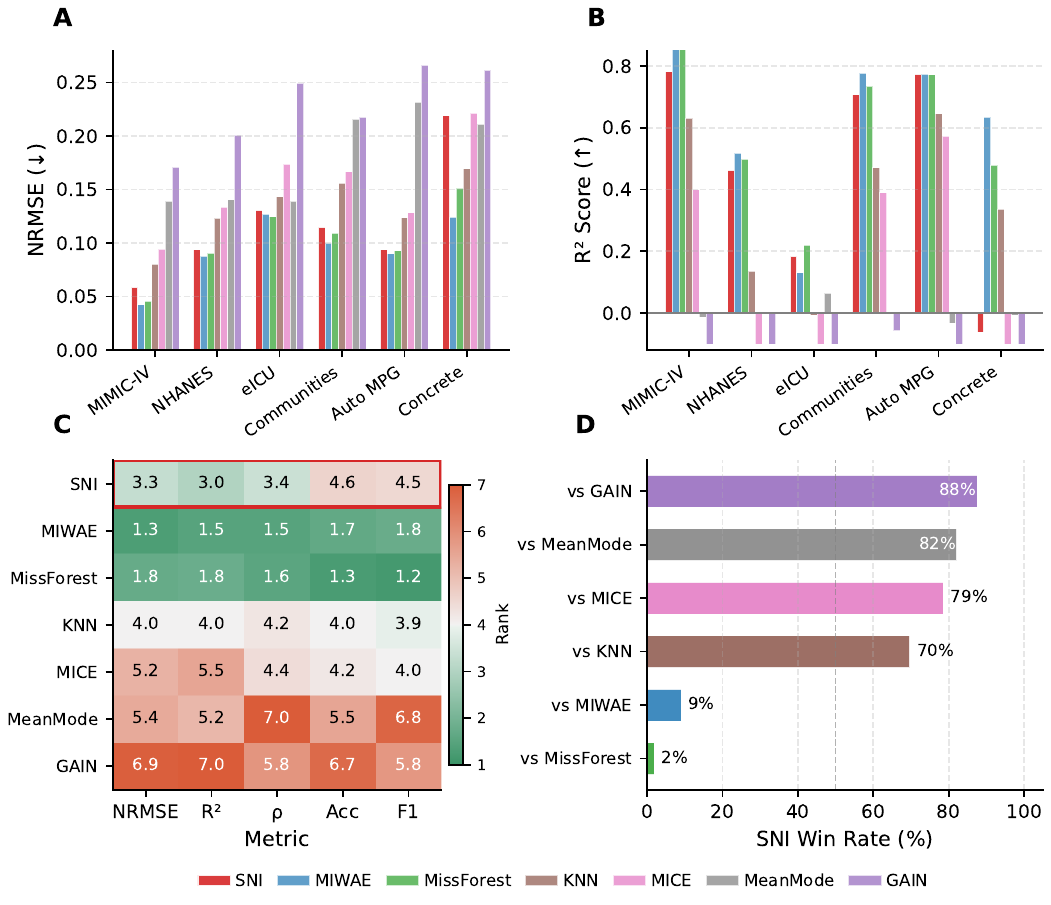}
\caption{Compact cross-domain summary of SNI versus six baselines under MCAR and strict MAR at 30\% missingness (mean over five seeds; panels A--B are further averaged over MCAR/MAR). 
(A) Continuous NRMSE by dataset (lower is better). 
(B) Continuous $R^2$ by dataset (higher is better; values below $-0.1$ are clipped for readability). 
(C) Average per-setting rank across metrics (lower is better), with SNI highlighted. 
(D) Pairwise win rate of SNI against each baseline, aggregated over all reported metrics and all dataset--mechanism settings (a ``win'' is defined as SNI outperforming the baseline on a given metric in a given setting).
Categorical metrics are computed only on datasets with categorical variables.\label{fig:combined_sni_analysis}}
\end{figure}

\paragraph{What interpretability means here}
Beyond predictive scores, SNI provides two intrinsic diagnostics derived from CPFA:
(i) a directed dependency matrix/network $D$ that summarizes which observed features were used to impute each target feature, and 
(ii) learned head-wise prior-strength coefficients $\{\lambda_h\}$ that quantify how strongly each attention head follows the correlation prior.
These outputs may support model auditing and scientific use cases where the {reliance pattern of the imputer} is itself of interest (e.g., feature screening, reliance checks, and leakage detection), even when an accuracy-first baseline achieves marginally better predictive performance.
Figure~\ref{fig:combined_sni_analysis} gives a compact cross-domain comparison across datasets, metrics, and win rates.

\subsection{Case Study: MIMIC-IV ICU Subset}
MIMIC-IV is representative of high-stakes mixed-type tabular data: continuous physiological variables (e.g., heart rate and respiratory rate) coexist with discrete targets such as discretized SpO$_2$ bins and ICU alarm categories, and the discrete labels can be strongly imbalanced. 
Under strict MAR with 30\% missingness, MIWAE and MissForest achieve the lowest continuous error (NRMSE = 0.042$\pm$0.000 and 0.045$\pm$0.000, respectively), while SNI attains NRMSE = 0.056$\pm$0.001 and $R^2$ = 0.773$\pm$0.007 (Table~\ref{tab:mimic_MAR30per_cont}). 
For the categorical variables, MissForest remains the strongest baseline (Macro-F$_1$ = 0.539$\pm$0.015), whereas SNI achieves Macro-F$_1$ = 0.301$\pm$0.018 (Table~\ref{tab:mimic_MAR30per_cat}), reflecting the difficulty of extremely imbalanced discrete targets.

\begin{figure}[h]
\centering
\includegraphics[width=1.0\textwidth]{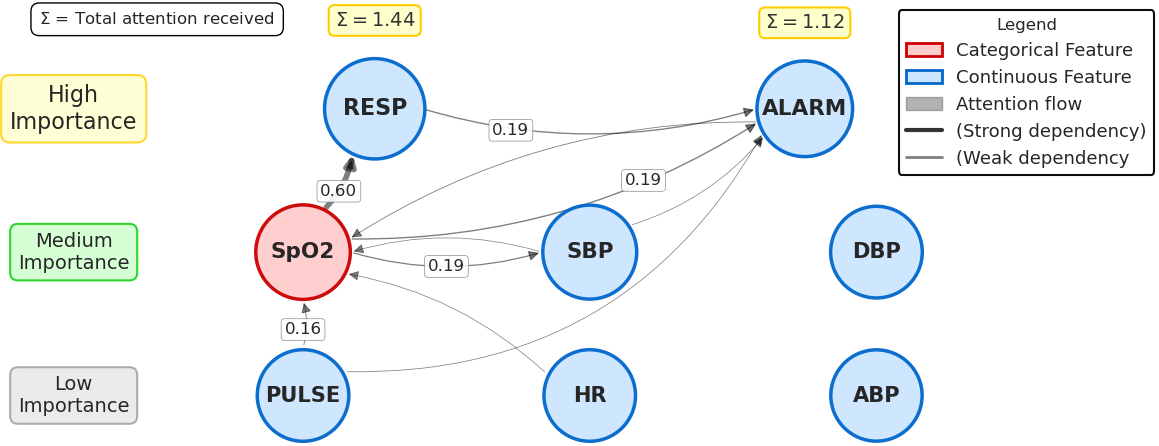}
\caption{Model-reliance dependency network derived from SNI on MIMIC-IV (strict MAR, 30\% missingness). Each directed edge $j\!\rightarrow\! i$ corresponds to an entry $D_{ij}$ and its thickness encodes the average attention mass, i.e., how strongly SNI relied on source feature $j$ when imputing target feature $i$. Node size reflects incoming mass $\Sigma_j=\sum_i D_{ij}$, highlighting globally informative source variables (RESP and ALARM in this subset). This visualization is intended as an imputation diagnostic and does not imply causal relationships. \label{fig:SNI_DepNet_MIMIC}}
\end{figure}

\paragraph{Interpretable deliverable: an imputation ``reliance map''}
Beyond accuracy, SNI produces a directed dependency matrix $D$ derived from controllable-prior feature attention (CPFA). 
Intuitively, $D_{ij}$ quantifies how much the imputer relied on source feature $j$ when imputing target feature $i$ (averaged across heads and samples). 
This should be interpreted as a {model-reliance diagnostic} rather than a causal graph.
Figure~\ref{fig:SNI_DepNet_MIMIC} visualizes $D$ as a sparse network: thicker edges $j\!\rightarrow\! i$ indicate stronger reliance of the imputer for target $i$ on source $j$. 
The incoming mass $\Sigma_j=\sum_i D_{ij}$ highlights source variables that broadly support imputing many other variables; in this subset RESP ($\Sigma=1.44$) and ALARM ($\Sigma=1.12$) appear to serve as prominent hubs.

\paragraph{From CPFA attention to $D$ and how to read it}
For each attention head $h$, CPFA produces a directed matrix $A^{(h)}\in\mathbb{R}^{d\times d}$, where row $i$ contains the normalized attention weights over source features used to impute target feature $i$. 
We aggregate heads as $D=\tfrac{1}{H}\sum_{h=1}^H A^{(h)}$ and set $D_{ii}=0$ to exclude self-dependencies. 
Thus, {columns correspond to sources and rows correspond to targets}: a large entry $D_{ij}$ means that when $i$ is missing, the imputer tends to consult $j$ strongly.

\begin{figure}[h]
\centering
\includegraphics[width=0.85\textwidth]{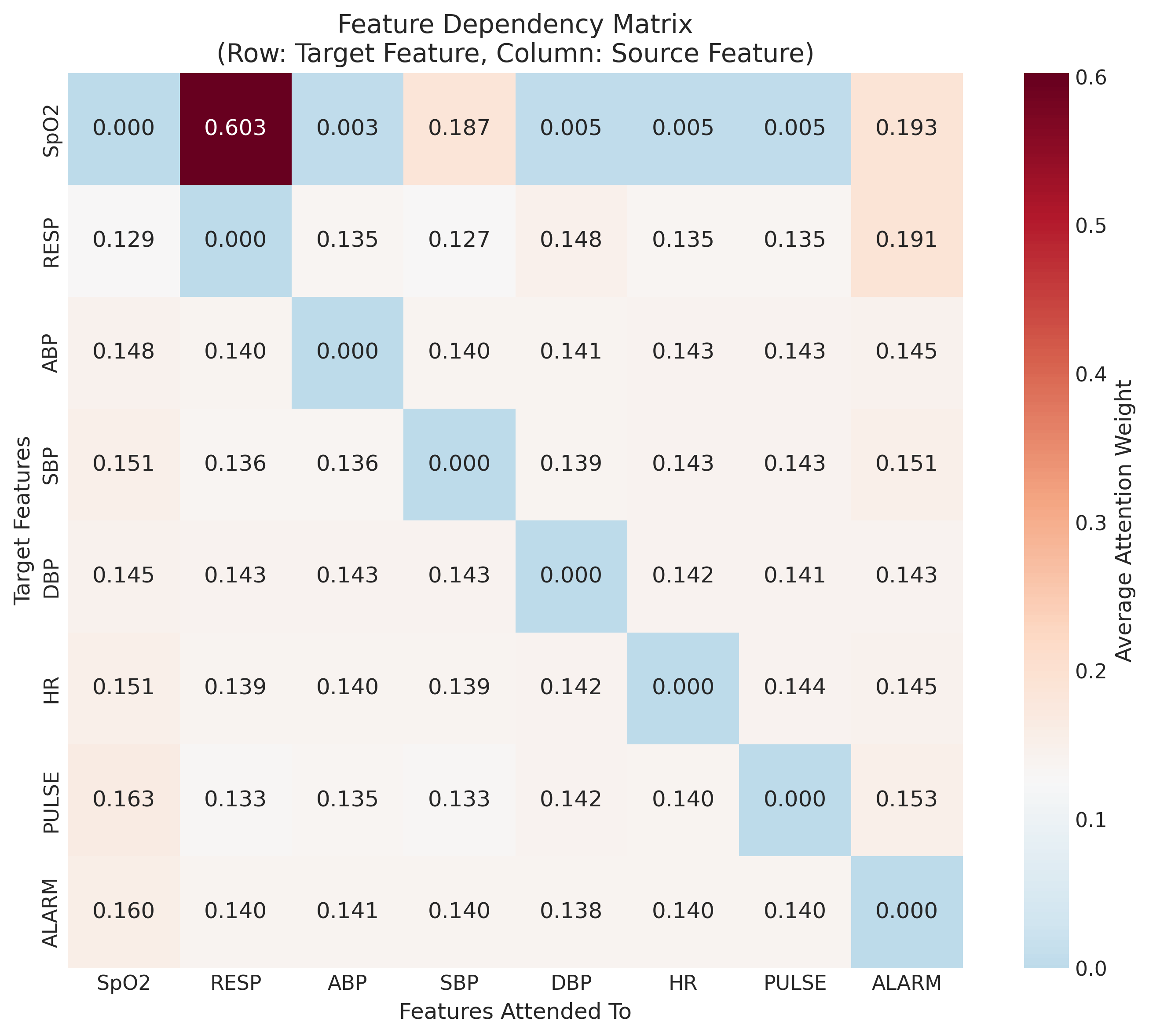}
\caption{Feature dependency matrix $D$ for MIMIC-IV derived from CPFA attention (strict MAR, 30\% missingness). $D_{ij}$ denotes the normalized contribution from source feature $j$ (column) when imputing target feature $i$ (row), with $D_{ii}=0$. The first row illustrates that imputing SpO$_2$ relies most on RESP (0.60), SBP (0.19), and ALARM (0.19). As a model-reliance view, $D$ provides a compact audit trail of which columns were used by the imputer and may help identify unexpected reliance patterns or inform feature screening, even when an accuracy-first baseline is preferred.\label{fig:SNI_DepMat_MIMIC}}
\end{figure}
\subsection{Case Study: NHANES Subset}
NHANES contains a heterogeneous mix of laboratory biomarkers and questionnaire-style variables, making it a useful case study for mixed-type imputation and interpretability. 
Under strict MAR with 30\% missingness, MIWAE achieves the strongest overall accuracy on both continuous and categorical variables ($R^2$ = 0.506$\pm$0.005, Macro-F$_1$ = 0.698$\pm$0.009), with MissForest close behind ($R^2$ = 0.485$\pm$0.004, Macro-F$_1$ = 0.693$\pm$0.003). 
SNI remains competitive ($R^2$ = 0.453$\pm$0.007, Macro-F$_1$ = 0.688$\pm$0.009; Tables~\ref{tab:NHANES_MAR30per_cont}--\ref{tab:NHANES_MAR30per_cat}) while additionally yielding interpretable reliance diagnostics.

\begin{table}[h]
\footnotesize
\centering
\caption{Continuous variable performance on NHANES subset, strict MAR, 30\% missingness. Values are mean $\pm$ SD over five seeds.\label{tab:NHANES_MAR30per_cont}}
\begin{tabular}{lccccc}
\toprule
Method & NRMSE$\downarrow$ & MAE$\downarrow$ & MB$\downarrow$ & $R^2\uparrow$ & Spearman $\rho\uparrow$ \\
\midrule
SNI & 0.094$\pm$0.000 & 10.456$\pm$0.143 & -0.895$\pm$0.458 & 0.453$\pm$0.007 & 0.675$\pm$0.002 \\
MissForest & 0.092$\pm$0.000 & 10.210$\pm$0.046 & -0.244$\pm$0.030 & 0.485$\pm$0.004 & 0.687$\pm$0.002 \\
MIWAE & \textbf{0.088$\pm$0.000} & \textbf{9.985$\pm$0.036} & -0.770$\pm$0.318 & \textbf{0.506$\pm$0.005} & \textbf{0.703$\pm$0.002} \\
GAIN & 0.203$\pm$0.007 & 26.374$\pm$3.317 & -0.644$\pm$7.005 & -1.881$\pm$0.698 & 0.213$\pm$0.044 \\
KNN & 0.128$\pm$0.000 & 13.611$\pm$0.000 & -0.899$\pm$0.000 & 0.077$\pm$0.000 & 0.379$\pm$0.000 \\
MICE & 0.134$\pm$0.001 & 15.136$\pm$0.267 & \textbf{-0.147$\pm$0.666} & -0.113$\pm$0.073 & 0.443$\pm$0.007 \\
Mean/Mode & 0.140$\pm$0.000 & 14.710$\pm$0.000 & 0.106$\pm$0.000 & -0.001$\pm$0.000 & 0.000$\pm$0.000 \\
\bottomrule
\end{tabular}
\end{table}

\begin{table}[h]
\footnotesize
\centering
\caption{Categorical variable performance on NHANES subset, strict MAR, 30\% missingness. Values are mean $\pm$ SD over five seeds.\label{tab:NHANES_MAR30per_cat}}
\begin{tabular}{lccc}
\toprule
Method & Accuracy$\uparrow$ & Macro-F$_1\uparrow$ & Cohen's $\kappa\uparrow$ \\
\midrule
SNI & 0.686$\pm$0.009 & 0.688$\pm$0.009 & 0.490$\pm$0.014 \\
MissForest & 0.694$\pm$0.003 & 0.693$\pm$0.003 & 0.493$\pm$0.006 \\
MIWAE & \textbf{0.700$\pm$0.008} & \textbf{0.698$\pm$0.009} & \textbf{0.513$\pm$0.016} \\
GAIN & 0.348$\pm$0.012 & 0.289$\pm$0.028 & 0.019$\pm$0.007 \\
KNN & 0.454$\pm$0.000 & 0.449$\pm$0.000 & 0.177$\pm$0.000 \\
MICE & 0.549$\pm$0.010 & 0.548$\pm$0.010 & 0.297$\pm$0.018 \\
Mean/Mode & 0.352$\pm$0.000 & 0.194$\pm$0.000 & 0.000$\pm$0.000 \\
\bottomrule
\end{tabular}
\end{table}

\begin{figure}[H]
\centering
\includegraphics[width=1.0\textwidth]{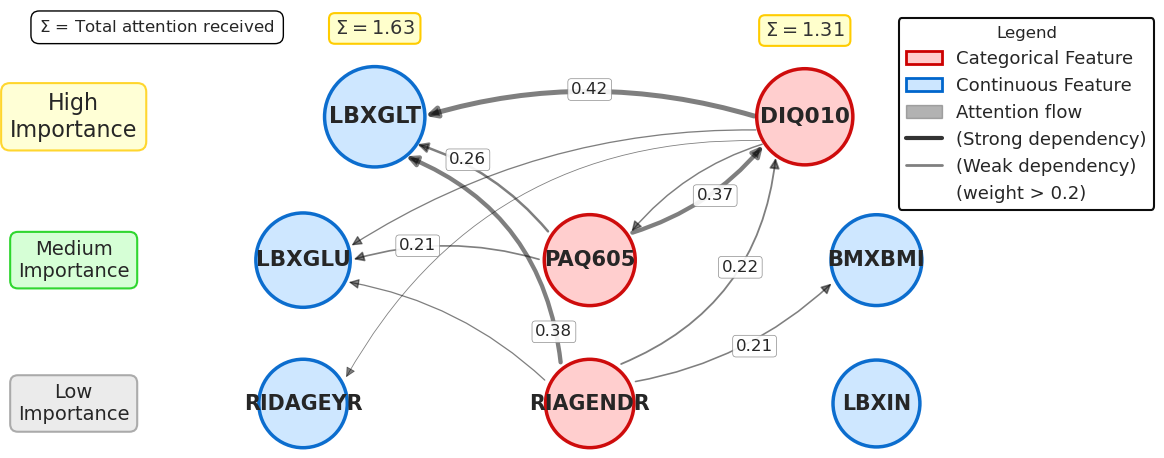}
\caption{Model-reliance dependency network derived from SNI on NHANES (strict MAR, 30\% missingness). Directed edges $j\!\rightarrow\! i$ correspond to entries of $D$ and indicate that the imputer relied on source variable $j$ when imputing target variable $i$; thickness encodes the average attention mass. Node size reflects incoming mass $\Sigma_j=\sum_i D_{ij}$, highlighting globally informative source variables. This view summarizes the imputer's reliance patterns and should not be interpreted as causal. \label{fig:SNI_NHANES_dep}}
\end{figure}

\paragraph{What the dependency network explains}
Figure~\ref{fig:SNI_NHANES_dep} summarizes the learned dependency matrix $D$ as a directed network. 
For a non-domain reader, the key point is: it makes explicit {which observed columns were used to fill in which missing columns}. 
For example, questionnaire and diet-related variables can contribute to imputing metabolic biomarkers, while highly connected hubs indicate variables that broadly support imputations across the dataset. 
This may be useful for auditing and transparency even when SNI is not the most accurate imputer, as it reveals the model's reliance structure rather than only the final filled-in table.

\begin{figure}[H]
\centering
\includegraphics[width=1.0\textwidth]{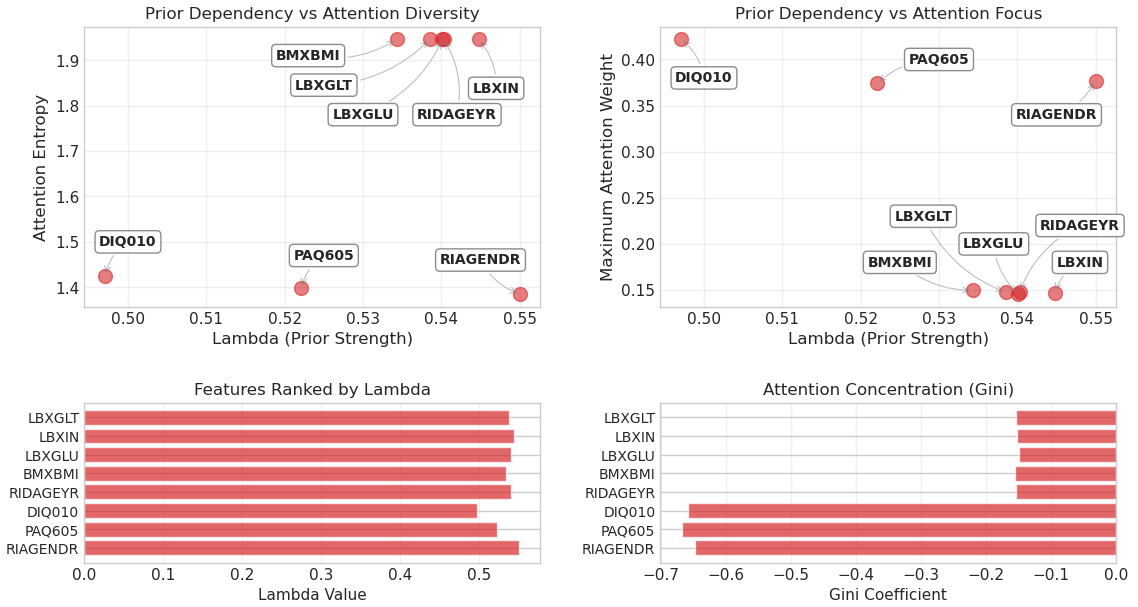}
\caption{Learned prior-strength coefficients $\lambda_h$ for CPFA heads on NHANES (strict MAR, 30\% missingness). Larger $\lambda_h$ indicates a head is more strongly regularized toward the correlation-derived prior, while smaller $\lambda_h$ indicates more data-driven attention. Heterogeneous $\lambda_h$ values suggest that different heads specialize in prior-guided versus data-driven dependencies, providing a quantitative handle on the statistical--neural trade-off. \label{fig:SNI_NHANES_lambda_att}}
\end{figure}

\paragraph{What the prior-strength coefficients explain}
Figure~\ref{fig:SNI_NHANES_lambda_att} shows the learned head-wise prior-strength coefficients $\lambda_h$. 
These values quantify the extent to which each attention head is constrained by the correlation-derived prior versus allowed to deviate in a data-driven manner. 
In practice, heterogeneous $\lambda_h$ values suggest that CPFA may use a mixture of ``prior-following'' and ``data-driven'' heads, providing a potentially useful handle for understanding (and possibly tuning) the inductive bias.

\subsection{Sanity Check: Dependency Recovery on Synthetic Data}
\label{subsec:sanity_synth}

A recurring concern is whether attention weights can be meaningfully summarized as ``feature dependencies'' (cf.\ ``attention is not explanation'').
We therefore conduct a reviewer-facing sanity check on synthetic mixed-type data where the ground-truth dependency graph is known.
Concretely, we generate incomplete tables under strict MAR (30\% missingness) from a sparse DAG and fit SNI to obtain the induced dependency matrix $D$ (row: target; column: source).
We treat each entry $D_{f,j}$ as a score for the directed dependence $j\!\to\!f$ and evaluate edge recovery using AUROC/AUPRC over all directed pairs and Precision@K/Recall@K (with $K$ set to the number of true parents of each target).
Full details and results are reported in Supplementary Section~S4 (Table~S21).

We consider three generating regimes: {linear\_gaussian}, {nonlinear\_mixed}, and an {interaction\_xor} stress test where several children are driven primarily by product/XOR terms such that {marginal} correlations can be weak even when conditional dependence is strong.
To isolate the role of statistical guidance, we compare SNI against a neural-only ablation (NoPrior) and a statistics-only diagnostic reference (PriorOnly) constructed from the same correlation-based prior used by SNI.

Table~S21 shows that SNI tends to improve over NoPrior across all regimes (e.g., nonlinear\_mixed AUROC $0.807\pm0.040$ vs.\ $0.695\pm0.084$; AUPRC $0.687\pm0.069$ vs.\ $0.555\pm0.085$), suggesting that prior regularization may help stabilize dependency recovery rather than merely post-hoc ``explaining'' a black box.
More importantly, in the interaction\_xor regime, SNI achieves the best edge-recovery performance (AUROC $0.848\pm0.017$, AUPRC $0.809\pm0.032$), exceeding PriorOnly (AUROC $0.800\pm0.048$, AUPRC $0.741\pm0.041$), which suggests that the induced $D$ is {not} simply a restatement of marginal correlations.
We emphasize that this sanity check supports interpreting $D$ as a {model-reliance diagnostic} when ground truth is available, while on real-world datasets $D$ should still be read as the imputer's internal reliance rather than causal effects.

\subsection{Additional Benchmarks: Cross-Domain Validation}
To evaluate generality beyond the two detailed mixed-type case studies (MIMIC-IV and NHANES), we additionally benchmark on AutoMPG, ComCri, Concrete, and eICU. Table~\ref{tab:summary_additional} summarizes a focused comparison between SNI and the strongest classical baseline, MissForest, under MAR 30\% missingness.

\begin{table}[h]
\footnotesize
\centering
\caption{Summary comparison on additional datasets under MAR 30\% missingness (SNI vs. MissForest). $\Delta$NRMSE is computed as $(\mathrm{NRMSE}_{\mathrm{MF}}-\mathrm{NRMSE}_{\mathrm{SNI}})/\mathrm{NRMSE}_{\mathrm{SNI}}\times 100\%$ (positive indicates SNI improves over MissForest). $\Delta$F$_1$ is computed as $(\mathrm{F1}_{\mathrm{SNI}}-\mathrm{F1}_{\mathrm{MF}})/\mathrm{F1}_{\mathrm{MF}}\times 100\%$ (positive indicates higher Macro-F$_1$ for SNI).\label{tab:summary_additional}}
\begin{tabular}{lccc ccc}
\toprule
Dataset & NRMSE$_{\mathrm{SNI}}$ & NRMSE$_{\mathrm{MF}}$ & $\Delta$NRMSE (\%) & Macro-F$_1{}_{\mathrm{SNI}}$ & Macro-F$_1{}_{\mathrm{MF}}$ & $\Delta$F$_1$ (\%) \\
\midrule
eICU & 0.135 & 0.125 & -7.4 & 0.541 & 0.605 & -10.6 \\
AutoMPG & 0.103 & 0.098 & -4.9 & 0.248 & 0.356 & -30.3 \\
ComCri & 0.119 & 0.112 & -5.9 & 0.592 & 0.633 & -6.5 \\
Concrete & 0.221 & 0.158 & -28.5 & -- & -- & -- \\
\bottomrule
\end{tabular}
\end{table}

\begin{figure}[ht]
\centering
\includegraphics[width=0.95\textwidth]{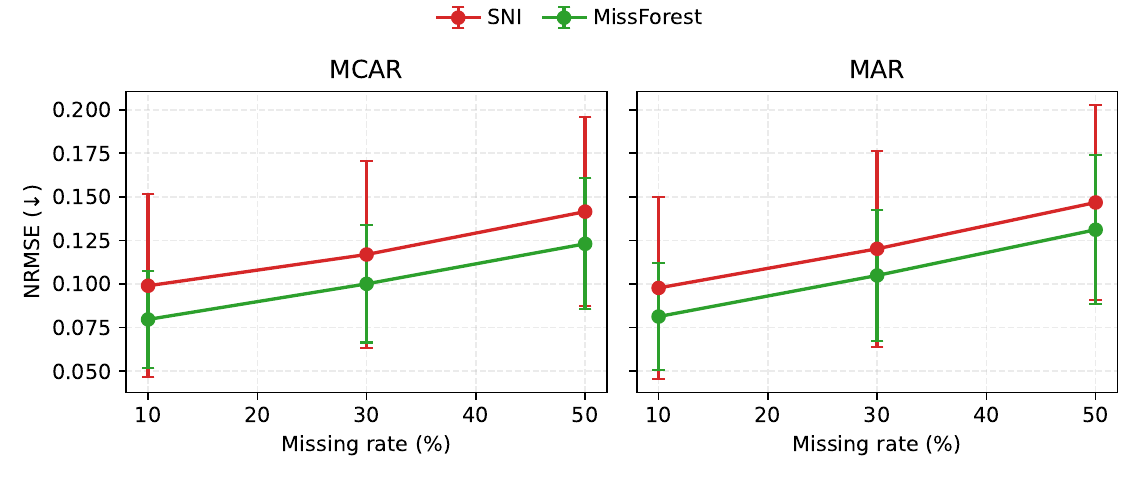}
\caption{Missing-rate sensitivity at 10\%, 30\%, and 50\% missingness. We report dataset-wise mean continuous NRMSE (lower is better) for SNI and MissForest under MCAR (left) and strict MAR (right); error bars indicate the standard deviation across datasets.\label{fig:rate_sweep_mar_nrmse}}
\end{figure}

\begin{figure}[ht]
\centering
\includegraphics[width=0.95\textwidth]{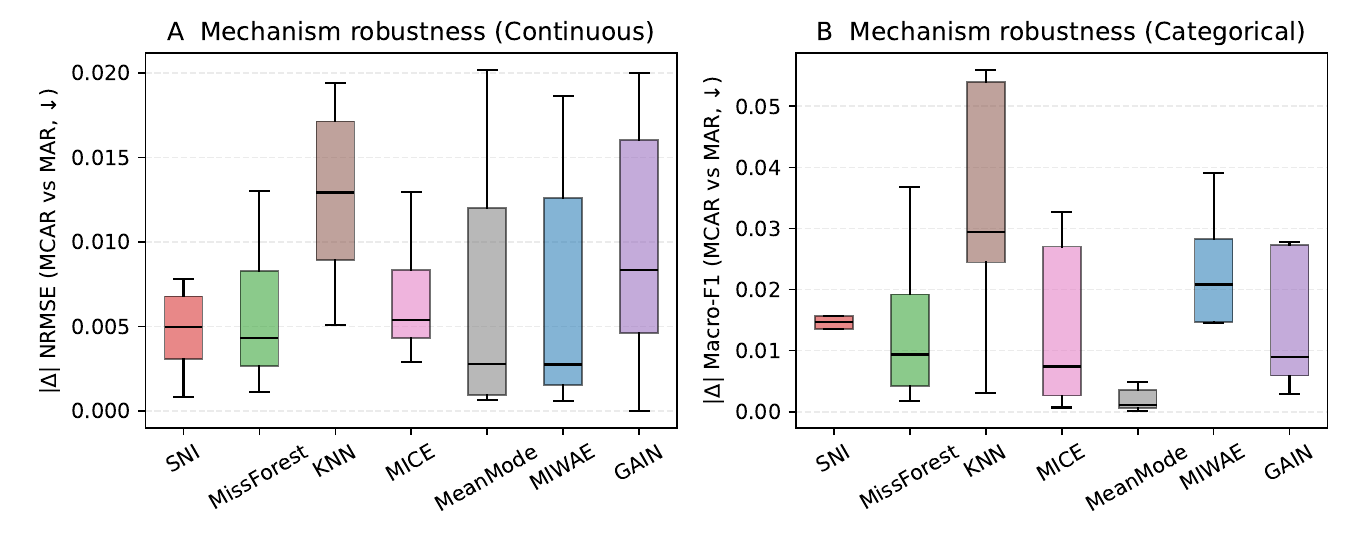}
\caption{Mechanism robustness quantified as the absolute performance shift between MCAR and strict MAR at 30\% missingness. 
Panel A reports $|\Delta|$ NRMSE on continuous variables and Panel B reports $|\Delta|$ Macro-F$_1$ on categorical variables (datasets without categorical features are excluded). 
Boxplots summarize per-dataset absolute differences, where smaller values indicate lower sensitivity to the missingness mechanism.\label{fig:mcar_mar_robustness}}
\end{figure}

To further summarize robustness to the missingness mechanism in a compact form, Figure~\ref{fig:mcar_mar_robustness} reports the absolute performance shift between MCAR and strict MAR at 30\% missingness across datasets (smaller is more robust).

Overall, SNI's continuous performance is closest to MissForest on AutoMPG ($\Delta$NRMSE -4.9\%) and ComCri ($\Delta$NRMSE -5.9\%), while it lags more on the fully continuous Concrete benchmark ($\Delta$NRMSE -28.5\%). For categorical variables, the gap can be more pronounced, particularly on eICU ($\Delta$F$_1$ -10.6\%) and AutoMPG ($\Delta$F$_1$ -30.3\%), which have imbalanced categories. These cross-domain results suggest that SNI's primary value may lie in its intrinsic interpretability and explicit statistical--neural trade-off; for pure accuracy on certain domains, strong nonparametric baselines (e.g., MissForest) and deep generative baselines (e.g., MIWAE) often achieve better performance.

\subsection{Computational Complexity and Scalability Analysis}
SNI incorporates a feature-attention module and iterative optimization, and consequently incurs higher computational cost than most classical imputers. Table~\ref{tab:complexity} reports end-to-end wall-clock runtimes per imputation run on MIMIC-IV (MAR 30\%). While SNI is slower than tree-based and simple statistical baselines, it may still be suitable for offline data cleaning scenarios where the explicit dependency outputs are of interest. Future work could explore improving computational efficiency and scaling SNI to higher-dimensional tables. Interestingly, KNN exhibits comparable runtime to SNI on this dataset due to the pairwise distance computations required for mixed-type data.
\begin{table}[h]
\footnotesize
\centering
\caption{End-to-end wall-clock runtime per imputation run on MIMIC-IV (MAR 30\%). Values are mean $\pm$ SD over five seeds.\label{tab:complexity}}
\begin{tabular}{lcc}
\toprule
Method & Runtime (s)$\downarrow$ & Runtime (min)$\downarrow$ \\
\midrule
SNI & 1224.14$\pm$58.35 & 20.40$\pm$0.97 \\
MissForest & 12.64$\pm$2.56 & 0.21$\pm$0.04 \\
MIWAE & 40.56$\pm$0.46 & 0.68$\pm$0.01 \\
GAIN & 19.50$\pm$0.08 & 0.32$\pm$0.00 \\
KNN & 1278.14$\pm$137.12 & 21.30$\pm$2.29 \\
MICE & 12.76$\pm$0.21 & 0.21$\pm$0.00 \\
Mean/Mode & \textbf{0.05$\pm$0.01} & \textbf{0.00$\pm$0.00} \\
\bottomrule
\end{tabular}
\end{table}

\begin{figure}[ht]
\centering
\includegraphics[width=\textwidth]{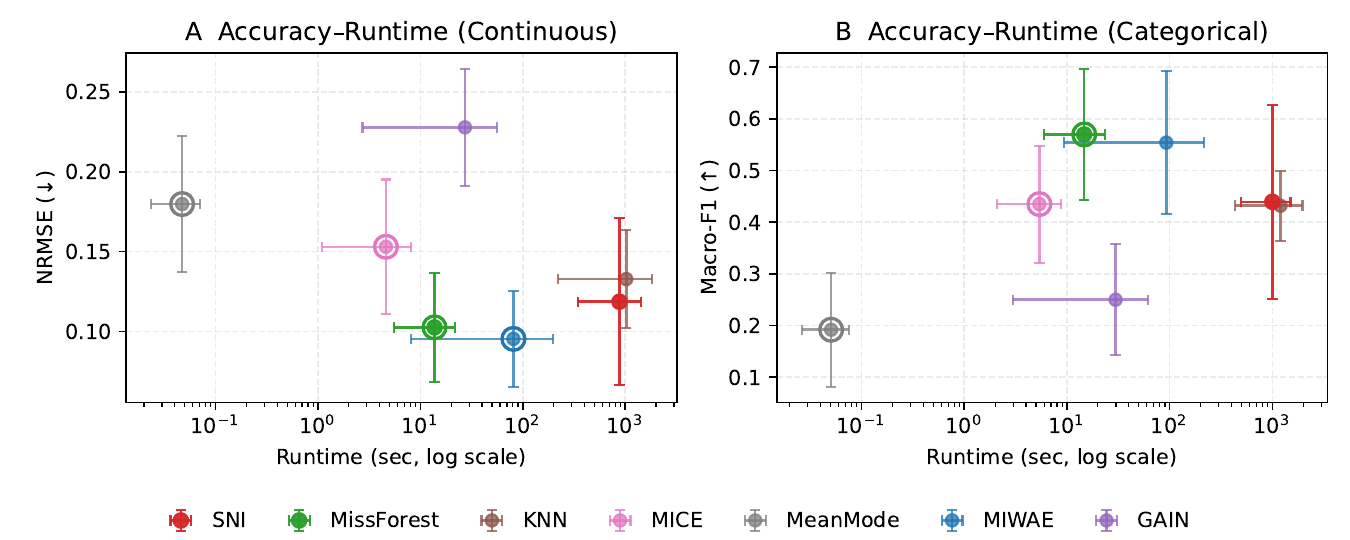}
\caption{Accuracy--runtime trade-off aggregated over all datasets under MCAR/MAR at 30\% missingness. 
(A) Continuous performance: NRMSE vs runtime (log scale; lower-left is better). 
(B) Categorical performance: Macro-F$_1$ vs runtime (log scale; upper-left is better). 
Error bars indicate standard deviation across dataset--mechanism settings.\label{fig:pareto_accuracy_runtime}}
\end{figure}

Figure~\ref{fig:pareto_accuracy_runtime} complements Table~\ref{tab:complexity} by summarizing the accuracy--runtime trade-off across domains under MCAR/MAR at 30\% missingness. 
Each point aggregates a method over all (dataset, mechanism) settings at this rate; horizontal/vertical error bars indicate the standard deviation across settings (not seed variance). 
Open circles mark Pareto-efficient operating points in each panel.

\section{Discussion}
\label{sec:discussion}
This work explores SNI as an attempt to balance two long-standing desiderata in tabular imputation: (i) statistical traceability (explicit priors and diagnostics) and (ii) representation flexibility (nonlinear function approximation). Across mixed-type datasets and missingness settings, our results suggest a recurring pattern: tree/instance-based baselines (especially MissForest) perform well on accuracy-first objectives, while SNI may offer advantages when continuous fidelity, cross-feature structure preservation, and intrinsic interpretability are simultaneously required. This section summarizes how to apply SNI in practice, what its interpretable artefacts mean (and do not mean), and where the method may fall short.

\subsection{Practical guidance: when to use SNI vs.\ MissForest}
\label{subsec:disc_practical}

Our experiments suggest that MissForest is a strong accuracy-first baseline, whereas SNI may be viewed as an accuracy--interpretability trade-off that could be useful when continuous-variable fidelity and dependency preservation are of interest.

\paragraph{SNI is preferable when one or more of the following holds}
\begin{itemize}
    \item \textbf{Mixed-type tables where downstream analysis depends on cross-feature structure.}
    If the downstream goal involves correlation-sensitive tasks (e.g., ranking, risk scoring, feature association analysis, or model training sensitive to joint structure), the combination of CPFA and controllable priors provides a mechanism to preserve multivariate structure beyond pointwise error.
    \item \textbf{Continuous-variable fidelity is the main objective (NRMSE/$R^2$/Spearman).}
In our benchmarks, SNI is generally competitive on continuous metrics; compared with GAN-based imputation (GAIN), it is markedly more stable on mixed-type settings, while MIWAE remains a strong accuracy-first deep baseline in several datasets.
When continuous targets dominate the evaluation and interpretability is desired, SNI may be a reasonable option to consider.
    \item \textbf{Intrinsic interpretability is required.}
    SNI outputs two artifacts that are directly inspectable without post-hoc explainer tooling: 
    (i) a {directed dependency matrix/network} $D$ summarizing source-to-target reliance, and 
    (ii) head-wise prior-confidence coefficients $\{\lambda_h\}$ that quantify how strongly each head is regularized toward correlation-derived priors.
    \item \textbf{Offline imputation is acceptable.}
    
SNI incurs higher computational cost than classical imputers; it may be more appropriate for offline data cleaning, auditing, or analysis pipelines where runtime is less critical than interpretability and continuous-variable fidelity.
\end{itemize}

\paragraph{MissForest is preferable when}
\begin{itemize}
    \item \textbf{Categorical performance under severe class imbalance is the priority.}
    For highly imbalanced categorical features (e.g., discretized clinical alarms or coarse bins), tree-based and instance-based methods tend to deliver stronger Macro-F1 and agreement metrics (e.g., $\kappa$).
    \item \textbf{The table is fully continuous and the goal is pure pointwise accuracy.}
    On fully continuous benchmarks (e.g., Concrete), our results indicate that MissForest can be substantially stronger, suggesting that SNI's inductive bias is most beneficial in genuinely mixed-type regimes rather than ``continuous-only, accuracy-only'' settings.
    \item \textbf{Fast CPU-only imputation is required at scale.}
    When throughput and wall-clock constraints dominate (e.g., large operational ETL pipelines), MissForest (and simpler baselines) remain attractive.
\end{itemize}

\paragraph{A pragmatic hybrid strategy. In practice, one can combine the strengths of both approaches}
\begin{itemize}
    \item \textbf{Column-wise hybrid:} use MissForest for a small subset of extremely imbalanced categorical columns, while using SNI for continuous columns (and moderately balanced categorical columns) when interpretability is desired.
    \item \textbf{Post-processing for rare categorical features:} apply a lightweight re-imputation (e.g., KNN) on a small set of problematic categorical variables, conditioned on CPFA-selected features, to narrow the categorical gap while preserving SNI's dependency analysis.
\end{itemize}
These hybrid recipes are intentionally simple: they leverage baseline strengths in areas where they tend to perform well, while preserving SNI's explicit dependency outputs for domain-facing analysis.

\subsection{What SNI’s statistical--neural interaction buys: balancing inductive bias and flexibility}
\label{subsec:disc_balance}

SNI is motivated by a practical observation: {purely statistical} imputers can be robust and fast but may struggle to express nonlinear cross-feature interactions, whereas {purely neural} imputers can model complex dependencies but may be unstable or opaque on tabular data, especially under substantial missingness.
SNI attempts to navigate this tension through {controllable priors}---correlation-derived priors $\mathbf{P}_f$ that regularize CPFA attention---and {learnable head-wise coefficients} $\{\lambda_h\}$ that control {how strongly} each attention head adheres to these priors.

Two empirical implications are worth emphasizing:
\begin{itemize}
    \item \textbf{A spectrum of behaviors, not a binary choice.}
    Instead of forcing the model to be ``statistical'' or ``neural'', different heads can settle into different regimes: some heads remain close to the correlation prior (large $\lambda_h$), while others deviate to capture higher-order interactions (small $\lambda_h$). 
    This makes the statistical--neural trade-off {data-adaptive} rather than hand-tuned.
    \item \textbf{Stability as a first-class design objective.}
The ablations suggest a recurring pattern: relaxing priors can slightly improve continuous fit in some settings but tends to reduce categorical stability, whereas strengthening priors can improve continuous metrics while modestly hurting categorical accuracy.
These observations are consistent with the motivation for head-wise controllability: the preferred operating point may depend on feature types, imbalance, and missingness severity.
\end{itemize}

From a methodological perspective, SNI can be viewed as a form of {guided attention learning} where classical correlation structure acts as a soft constraint rather than a hard rule. This framing may help explain why SNI can be competitive on continuous metrics in mixed-type tables while classical methods continue to excel in ``accuracy-only'' regimes.

\subsection{Interpretable artefacts: how to read $D$, $\Sigma_j$, and $\{\lambda_h\}$ (and what not to claim)}
\label{subsec:disc_interpretability}

\paragraph{Sanity-checking the dependency view}
A common concern is that attention weights are not automatically equivalent to statistical dependence or explanation.
To strengthen the interpretability claim, we include a synthetic sanity check with known ground-truth dependencies (Supplementary Section~S4, Table~S21). In this controlled setting, high-weight edges in $D$ recover true parent relationships substantially above chance, supporting the use of $D$ as a {model-reliance diagnostic}. 
We still caution that, on real-world data, $D$ summarizes which variables the imputer {consulted} and should not be interpreted as a causal graph.

\paragraph{A concrete usage workflow (why interpretability can be worth the cost)}
When runtime or absolute accuracy is the primary constraint, an accuracy-first imputer (e.g., MissForest) may be preferred for production ETL. 
SNI is most useful as an {audit and analysis layer}: one can run SNI on the same dataset (or a representative subset) to obtain $D$ and $\{\lambda_h\}$, then use these artefacts to (i) identify potential unexpected shortcuts or leakage (e.g., strong reliance on administrative fields), (ii) identify globally informative hubs via $\Sigma_j=\sum_i D_{ij}$, and (iii) guide feature screening or measurement prioritization in offline scientific/clinical analysis. 
This framing clarifies the value proposition when SNI is slower: the dependency view is a distinct deliverable rather than a by-product of imputation.

A notable aspect of SNI is that interpretability arises from the model architecture itself: the dependency matrix $D$ is obtained by aggregating CPFA attention, and $\{\lambda_h\}$ directly parameterize the prior regularizer. Nevertheless, it is important to interpret these artifacts conservatively.

\paragraph{Dependency matrix/network as an imputation-centric summary}
Each entry $D_{f,j}$ quantifies how much attention mass the model assigns to {source feature} $j$ when imputing {target feature} $f$ (row: target, column: source), after head and sample aggregation.
Large $D_{f,j}$ indicates that feature $j$ is {predictively informative for imputing} $f$ under the fitted model and the observed data distribution.
Similarly, the column sum $\Sigma_j=\sum_f D_{f,j}$ summarizes how broadly feature $j$ is attended across targets and can be used to identify {globally attended sources} (``information hubs'').

\paragraph{What $D$ and $\Sigma_j$ are NOT. We explicitly caution against causal interpretation}
\begin{itemize}
    \item Attention-based dependencies reflect {statistical/predictive associations} learned for imputation, not cause--effect relations.
    \item Directionality in $D$ reflects {model usage} (which sources help predict which targets), not necessarily real-world direction of influence.
    \item Hubness (large $\Sigma_j$) indicates a feature is broadly useful for predicting other features {in this dataset}, but may also reflect measurement conventions, feature redundancy, or preprocessing choices.
\end{itemize}

\paragraph{Reading $\{\lambda_h\}$ as a diagnostic. The head-wise coefficients provide a compact diagnostic of the model's operating regime}
\begin{itemize}
    \item If many heads exhibit persistently large $\lambda_h$, the learned attention is strongly aligned with correlation priors, suggesting the dataset's dominant dependencies may be well approximated by linear structure (or that the nonlinear signal is weak relative to noise/missingness).
    \item If $\lambda_h$ values are small across heads, the model is relying less on the correlation prior, which may occur when nonlinear patterns are important, or when the prior is unreliable (e.g., small-sample correlation estimates, heavy-tailed variables, or strong nonlinearity).
\end{itemize}
In this sense, $\{\lambda_h\}$ can be used not only for interpretability but also for {model debugging}: extreme regimes may motivate revisiting correlation estimation, preprocessing, or hyper-parameters.

\subsection{Missingness mechanisms: what SNI can and cannot guarantee under MNAR}
\label{subsec:disc_mnar}

SNI is motivated by an EM-inspired view and is most naturally aligned with MAR-style assumptions in classical missing-data theory. 
Accordingly, we treat MNAR as a {stress test} rather than a setting where identifiability is guaranteed.
Our MNAR experiments (including mask-aware variants reported in the Supplementary Material) suggest that incorporating missingness indicators can improve robustness in some cases, but we do {not} claim a general solution to arbitrary MNAR.

Two practical takeaways follow:
\begin{itemize}
    \item \textbf{If MNAR is suspected, include the mask.}
    When missingness itself may carry information (e.g., measurement triggered by clinical concern, or sensor dropouts correlated with extreme values), mask-aware inputs can provide the model with additional context. This is a pragmatic mitigation, not a proof of identifiability.
    \item \textbf{Evaluate sensitivity, not only averages.}
    Under MNAR, it is particularly important to inspect per-feature behavior (especially categorical features) and to report variability across seeds and mechanisms. In practice, model selection should consider whether performance can vary under MNAR perturbations; we report these evaluations as stress tests rather than as a guarantee of MNAR robustness (Supplementary Tables~S13--S19).
\end{itemize}

More broadly, our results are consistent with a conservative stance: {reasonable empirical performance under synthetic MNAR does not, by itself, justify causal or counterfactual conclusions}. In these settings, SNI may be viewed as offering (i) a competitive imputer and (ii) explicit diagnostics that reveal what the model relied on when imputing.

\subsection{Limitations and future directions}
\label{subsec:disc_limitations}

While SNI attempts to combine statistical priors with neural attention for mixed-type imputation, several limitations should be acknowledged.

\paragraph{(1) Categorical imbalance remains challenging}
SNI is not consistently the strongest method for severely imbalanced categorical variables, where MissForest and KNN often achieve higher Macro-F$_1$ and agreement metrics.
This suggests that stronger discrete inductive biases (e.g., imbalance-aware objectives, calibrated probability outputs, or targeted post-processing) may be needed for highly skewed categorical columns.

\paragraph{(2) Runtime and scalability}
SNI incurs higher wall-clock cost than classical imputers due to iterative optimization and feature-attention computation.
This limits applicability to settings where offline processing is acceptable.
Potential directions for future work include: (i) stronger parallelization across target features, (ii) low-rank/sparse attention approximations to reduce $O(d^2)$ cost, and (iii) warm-starting across EM-inspired iterations.

\paragraph{(3) Dependence on correlation priors and preprocessing}
Correlation-based priors can be noisy in small samples or under heavy-tailed distributions, and label encoding for categorical variables can introduce arbitrary numeric geometry.
Although SNI uses priors as soft regularizers (not as hard constraints), improving prior estimation (e.g., shrinkage/robust correlations, or category-aware association measures) may be a useful direction.

\paragraph{(4) Interpretability does not eliminate ambiguity}
Although $D$ and $\{\lambda_h\}$ are derived directly from the model, they summarize {model reliance} rather than ground-truth relationships.
Their stability across seeds, missingness realizations, and preprocessing choices should be examined when these outputs are used for domain-facing conclusions.

\paragraph{(5) Beyond static tabular data}
Our evaluation focuses on static tabular benchmarks. Many high-impact domains (ICU monitoring, industrial logs) have temporal structure and time-dependent missingness.
Extending SNI to longitudinal settings---while retaining explicit feature-dependency summaries---is a potential direction for future work.

In summary, SNI can be viewed as a diagnosable and interpretable imputer for mixed-type data that tends to achieve competitive continuous-variable performance while providing explicit dependency views. At the same time, classical methods remain competitive (and sometimes superior) for accuracy-first regimes, particularly for imbalanced categorical variables and fully continuous accuracy-only benchmarks.

\section{Conclusions}
\label{sec:conclusion}

This paper presented SNI, a controllable-prior feature-attention framework for imputing mixed-type missing data with explicit interpretability. 
By learning head-wise prior coefficients and aggregating controllable-prior feature attention into a directed dependency matrix, SNI seeks to support both imputation and dependency analysis within a unified architecture.

Across six real-world datasets under MCAR/strict-MAR at 30\% missingness, SNI achieves generally competitive continuous-variable performance (average ranks 3.33 for NRMSE and 3.50 for $R^2$ among seven methods), while producing intrinsic interpretability artefacts: a directed dependency matrix/network $D$ and head-wise prior-strength coefficients $\{\lambda_h\}$. 
We also observe that categorical imputation under severe class imbalance remains challenging, where accuracy-first baselines such as MissForest and KNN often achieve stronger performance. 
Additionally, we report MNAR evaluations as stress tests and consider a mask-aware variant (SNI-M) to probe sensitivity to non-ignorable missingness; these results should not be interpreted as a general MNAR solution. 
Overall, SNI may be best viewed as an interpretable imputer for offline analysis in settings where understanding the model's reliance structure is of interest.

Several limitations should be acknowledged. 
First, SNI exhibits reduced categorical performance under severe class imbalance compared to tree-based and instance-based methods. 
Second, SNI incurs higher computational cost than classical imputers, which may limit its applicability in time-sensitive pipelines. 
Third, the correlation-based priors can be sensitive to sample size and distributional assumptions.

Potential directions for future work include: 
(i) improved categorical calibration through imbalance-aware objectives or calibrated probability outputs, 
(ii) scalable implementations and approximation schemes for higher-dimensional tables, and 
(iii) combining SNI with ensembling or stronger tabular inductive biases to further improve robustness across diverse data regimes.

\clearpage
\section*{Author Contributions}
O.D.: Conceptualization, Methodology, Software, Data Curation, Validation, 
      Formal Analysis, Investigation, Writing – Original Draft, 
      Writing – Review \& Editing, Visualization.
S.N.: Resources, Supervision, Funding Acquisition.
A.O.: Supervision, Funding Acquisition.
Q.J.: Supervision, Funding Acquisition.

\section*{Declaration of Interest}
The authors declare that they have no known competing financial interests or personal relationships that could have appeared to influence the work reported in this paper.

\section*{Ethics Statement}
This study used deidentified clinical data from the MIMIC-IV and eICU databases, accessed through PhysioNet under credentialed access approval. The responsible investigator completed the CITI Program ``Data or Specimens Only Research'' training (Completion Record ID: 68986283, valid April 11, 2025 to April 11, 2029) and signed the required data use agreement. No additional ethics approval was required as the study involved analysis of deidentified, publicly available data.

\section*{Code and Data Availability}
All source code, experimental configurations, and preprocessing scripts are publicly available at GitHub (\url{https://github.com/oudeng/SNI}). A persistent archive of the code is deposited on Zenodo with DOI: \href{https://doi.org/10.5281/zenodo.18286410}{10.5281/zenodo.18286410}. Experimental results are publicly available at Zenodo with DOI: \href{https://doi.org/10.5281/zenodo.18286545}{10.5281/zenodo.18286545}.

The datasets used in this study are available as follows: MIMIC-IV and eICU require credentialed access through PhysioNet (\url{https://physionet.org/}); NHANES is publicly available from CDC (\url{https://wwwn.cdc.gov/nchs/nhanes/}); Communities \& Crime, AutoMPG, and Concrete are available from the UCI Machine Learning Repository (\url{https://archive.ics.uci.edu/}).

\section*{Acknowledgements}
The work was supported in part by the 2022-2024 Masaru Ibuka Foundation Research Project on Oriental Medicine, 2020-2025 JSPS A3 Foresight Program (Grant No. JPJSA3F20200001), 2022-2024 Japan National Initiative Promotion Grant for Digital Rural City, 2023 and 2024 Waseda University Grants for Special Research Projects (Nos. 2023C-216 and 2024C-223), 2023-2024 Waseda University Advanced Research Center Project for Regional Cooperation Support, and 2023-2024 Japan Association for the Advancement of Medical Equipment (JAAME) Grant.

\bibliographystyle{elsarticle-num}  
\bibliography{references}

\begin{thebibliography}{10}
\expandafter\ifx\csname url\endcsname\relax
  \def\url#1{\texttt{#1}}\fi
\expandafter\ifx\csname urlprefix\endcsname\relax\def\urlprefix{URL }\fi
\expandafter\ifx\csname href\endcsname\relax
  \def\href#1#2{#2} \def\path#1{#1}\fi

\bibitem{Rubin1976}
D.~B. Rubin, Inference and missing data, Biometrika 63~(3) (1976) 581--592.
\newblock \href {https://doi.org/10.1093/biomet/63.3.581}
  {\path{doi:10.1093/biomet/63.3.581}}.

\bibitem{Dempster1977}
A.~P. Dempster, N.~M. Laird, D.~B. Rubin, Maximum likelihood from incomplete
  data via the em algorithm, Journal of the Royal Statistical Society, Series B
  39 (1977) 1--38.
\newblock \href {https://doi.org/10.1111/j.2517-6161.1977.tb01600.x}
  {\path{doi:10.1111/j.2517-6161.1977.tb01600.x}}.

\bibitem{Borisov2024}
V.~Borisov, T.~Leemann, K.~Se{\ss}ler, J.~Haug, M.~Pawelczyk, G.~Kasneci, Deep
  neural networks and tabular data: A survey, IEEE Transactions on Neural
  Networks and Learning Systems 35~(6) (2024) 7499--7519.
\newblock \href {https://doi.org/10.1109/TNNLS.2022.3229161}
  {\path{doi:10.1109/TNNLS.2022.3229161}}.

\bibitem{Tjoa2021}
E.~Tjoa, C.~Guan, A survey on explainable artificial intelligence (xai): Toward
  medical xai, IEEE Transactions on Neural Networks and Learning Systems
  32~(11) (2021) 4793--4813.
\newblock \href {https://doi.org/10.1109/TNNLS.2020.3027314}
  {\path{doi:10.1109/TNNLS.2020.3027314}}.

\bibitem{Rubin1987}
D.~B. Rubin, Multiple Imputation for Nonresponse in Surveys, John Wiley \&
  Sons, 1987.
\newblock \href {https://doi.org/10.1002/9780470316696}
  {\path{doi:10.1002/9780470316696}}.

\bibitem{Little2019}
R.~Little, D.~Rubin, CStatistical analysis with missing data, 3rd Edition,
  Wiley, 2019.
\newblock \href {https://doi.org/10.1002/9781119482260}
  {\path{doi:10.1002/9781119482260}}.

\bibitem{Schafer1997}
J.~L. Schafer, Analysis of Incomplete Multivariate Data, Chapman \& Hall, 1997.
\newblock \href {https://doi.org/10.1201/9781439821862}
  {\path{doi:10.1201/9781439821862}}.

\bibitem{LittleRubin2019}
R.~J.~A. Little, D.~B. Rubin, Statistical Analysis with Missing Data (3rd ed.),
  John Wiley \& Sons, 2019.
\newblock \href {https://doi.org/10.1002/9781119482260}
  {\path{doi:10.1002/9781119482260}}.

\bibitem{Enders2017}
C.~K. Enders, Multiple imputation as a flexible tool for missing data handling
  in clinical research., Behav Res Ther 98 (2017) 4--18.
\newblock \href {https://doi.org/10.1016/j.brat.2016.11.008}
  {\path{doi:10.1016/j.brat.2016.11.008}}.

\bibitem{Janssen2010}
K.~J. Janssen, R.~Donders, F.~E. Harrell, Y.~Vergouwe, Q.~X. Chen, D.~E.
  Grobbee, K.~G.~M. Moons, Missing covariate data in medical research: to
  impute is better than to ignore., Journal of clinical epidemiology 63 7
  (2010) 721--7.
\newblock \href {https://doi.org/10.1016/j.jclinepi.2009.12.008}
  {\path{doi:10.1016/j.jclinepi.2009.12.008}}.

\bibitem{vanBuuren2018}
S.~v. Buuren, Flexible Imputation of Missing Data, 2nd Edition, Chapman and
  Hall/CRC, 2018.
\newblock \href {https://doi.org/10.1201/9780429492259}
  {\path{doi:10.1201/9780429492259}}.

\bibitem{Raghunathan2001}
T.~E. Raghunathan, J.~M. Lepkowski, J.~Van~Hoewyk, P.~Solenberger, A
  multivariate technique for multiply imputing missing values using a sequence
  of regression models, Survey Methodology 27~(1) (2001) 85--95, available at:
  \url{https://api.semanticscholar.org/CorpusID:10201308}. Accessed on Dec 31,
  2025.

\bibitem{Azur2011}
M.~J. Azur, E.~A. Stuart, C.~Frangakis, P.~J. Leaf, Multiple imputation by
  chained equations: what is it and how does it work?, International Journal of
  Methods in Psychiatric Research 20~(1) (2011) 40--49.
\newblock \href {https://doi.org/10.1002/mpr.329} {\path{doi:10.1002/mpr.329}}.

\bibitem{White2011}
W.~I. R., P.~Royston, A.~M. Wood, Multiple imputation using chained equations:
  Issues and guidance for practice, Statistics in Medicine 30 (2011).
\newblock \href {https://doi.org/10.1002/sim.4067}
  {\path{doi:10.1002/sim.4067}}.

\bibitem{Dong2013}
Y.~Dong, C.~Y.~J. Peng, Principled missing data methods for researchers,
  SpringerPlus 2 (2013).
\newblock \href {https://doi.org/10.1186/2193-1801-2-222}
  {\path{doi:10.1186/2193-1801-2-222}}.

\bibitem{Schafer2002}
J.~L. Schafer, J.~W. Graham, Missing data: our view of the state of the art,
  Psychological Methods 7 2 (2002) 147--177.
\newblock \href {https://doi.org/10.1037/1082-989X.7.2.147}
  {\path{doi:10.1037/1082-989X.7.2.147}}.

\bibitem{Stekhoven2011}
D.~J. Stekhoven, P.~Buhlmann, Missforest --- non-parametric missing value
  imputation for mixed-type data, Bioinformatics 28~(1) (2011) 112--118.
\newblock \href {https://doi.org/10.1093/bioinformatics/btr597}
  {\path{doi:10.1093/bioinformatics/btr597}}.

\bibitem{Yoon2018}
J.~S. Yoon, J.~Jordon, M.~Schaar, Gain: Missing data imputation using
  generative adversarial nets, in: Proceedings of the 35th International
  Conference on Machine Learning, PMLR, 2018, pp. 5689--5698, available at:
  \url{https://proceedings.mlr.press/v80/yoon18a.html}. Accessed on Dec 31,
  2025.

\bibitem{Mattei2019}
P.~A. Mattei, J.~Frellsen, Miwae: Deep generative modelling and imputation of
  incomplete data sets, in: Proceedings of the 36th International Conference on
  Machine Learning, Vol.~97 of Proceedings of Machine Learning Research (PMLR),
  2019, pp. 4413--4423, available at:
  \url{https://proceedings.mlr.press/v97/mattei19a.html}. Accessed on Dec 31,
  2025.

\bibitem{Nazabal2020}
A.~Nazabal, P.~Olmos, Z.~Ghahramani, I.~Valera, Handling incomplete
  heterogeneous data using vaes, Pattern Recognition 107 (2020) 107501.
\newblock \href {https://doi.org/10.1016/j.patcog.2020.107501}
  {\path{doi:10.1016/j.patcog.2020.107501}}.

\bibitem{Gondara2018}
L.~Gondara, K.~Wang, Mida: Multiple imputation using denoising autoencoders,
  in: Advances in Knowledge Discovery and Data Mining: 22nd Pacific-Asia
  Conference, PAKDD, 2018, pp. 260--272.
\newblock \href {https://doi.org/10.1007/978-3-319-93040-4_21}
  {\path{doi:10.1007/978-3-319-93040-4_21}}.

\bibitem{Beaulieu-Jones2017}
B.~Beaulieu-Jones, J.~Moore, Missing data imputation in the electronic health
  record using deeply learned autoencoders, Pacific Symposium on Biocomputing.
  Pacific Symposium on Biocomputing (2017) 207--218\href
  {https://doi.org/10.1142/9789813207813_0021}
  {\path{doi:10.1142/9789813207813_0021}}.

\bibitem{Cao2018}
W.~Cao, D.~Wang, J.~Li, H.~Zhou, Y.~Li, L.~Li, Brits: bidirectional recurrent
  imputation for time series, in: Proceedings of the 32nd International
  Conference on Neural Information Processing Systems, NIPS'18, 2018, pp.
  6776--6786, available at:
  \url{https://proceedings.neurips.cc/paper/2018/hash/734e6bfcd358e25ac1db0a4241b95651-Abstract.html}.
  Accessed on Dec 31, 2025.

\bibitem{Che2016}
Z.~Che, S.~Purushotham, K.~Cho, D.~Sontag, Y.~Liu, Recurrent neural networks
  for multivariate time series with missing values, Scientific Reports 8
  (2016).
\newblock \href {https://doi.org/10.1038/s41598-018-24271-9}
  {\path{doi:10.1038/s41598-018-24271-9}}.

\bibitem{Luo2018}
Y.~Luo, X.~Cai, Y.~Zhang, J.~Xu, X.~Yuan, Multivariate time series imputation
  with generative adversarial networks, in: Proceedings of the 32nd
  International Conference on Neural Information Processing Systems, NIPS'18,
  2018, p. 1603–1614, available at:
  \url{https://proceedings.neurips.cc/paper/2018/hash/96b9bff013acedfb1d140579e2fbeb63-Abstract.html}.
  Accessed on Dec 31, 2025.

\bibitem{Lall2021}
R.~Lall, T.~Robinson, The midas touch: Accurate and scalable missing-data
  imputation with deep learning, Political Analysis (2021) 1--18\href
  {https://doi.org/10.1017/pan.2020.5} {\path{doi:10.1017/pan.2020.5}}.

\bibitem{Chang2025}
L.~Chang, C.~Li, C.~Yang, S.~Lin, Learning on missing tabular data: Attention
  with self-supervision, not imputation, is all you need, ACM Trans. Intell.
  Syst. Technol. 16~(3) (2025).
\newblock \href {https://doi.org/10.1145/3729241} {\path{doi:10.1145/3729241}}.

\bibitem{Huang2020}
X.~Huang, A.~Khetan, M.~Cvitkovic, Z.~Karnin, Tabtransformer: Tabular data
  modeling using contextual embeddings, arXiv preprint (2020).
\newblock \href {http://arxiv.org/abs/2012.06678} {\path{arXiv:2012.06678}}.

\bibitem{Arik2021}
S.~Arik, T.~Pfister, Tabnet: Attentive interpretable tabular learning, in:
  Proceedings of the AAAI Conference on Artificial Intelligence, 2021, pp.
  6679--6687.
\newblock \href {https://doi.org/10.1609/aaai.v35i8.16826}
  {\path{doi:10.1609/aaai.v35i8.16826}}.

\bibitem{Vaswani2017}
A.~Vaswani, N.~Shazeer, N.~Parmar, J.~Uszkoreit, L.~Jones, A.~N. Gomez,
  L.~Kaiser, I.~Polosukhin, Attention is all you need, in: Proceedings of the
  31st International Conference on Neural Information Processing Systems, NIPS,
  2017, pp. 6000--6010, available at:
  \url{https://papers.nips.cc/paper/2017/hash/3f5ee243547dee91fbd053c1c4a845aa-Abstract.html}.
  Accessed on Dec 31, 2025.

\bibitem{Kossen2021}
J.~Kossen, N.~Band, C.~Lyle, A.~N. Gomez, T.~Rainforth, Y.~Gal, Self-attention
  between datapoints: Going beyond individual input-output pairs in deep
  learning, in: Advances in Neural Information Processing Systems, Vol.~34,
  2021, pp. 28742--28756.

\bibitem{Pilaluisa2022}
J.~Pilaluisa, D.~Tom\'{a}s, B.~N.~Colorado, J.~N. Maz\'{o}n, Contextual word
  embeddings for tabular data search and integration, Neural Comput. Appl.
  35~(13) (2022) 9319--9333.

\bibitem{Tay2022}
Y.~Tay, M.~Dehghani, D.~Bahri, D.~Metzler, Efficient transformers: A survey,
  ACM Computing Surveys 55~(6) (2022).
\newblock \href {https://doi.org/10.1145/3530811} {\path{doi:10.1145/3530811}}.

\bibitem{Xiao2018}
C.~Xiao, E.~Choi, J.~Sun, Opportunities and challenges in developing deep
  learning models using electronic health records data: a systematic review,
  Journal of the American Medical Informatics Association (JAMIA) 25~(10)
  (2018) 1419--1428.
\newblock \href {https://doi.org/10.1093/jamia/ocy068}
  {\path{doi:10.1093/jamia/ocy068}}.

\bibitem{Fridgeirsson2023}
E.~A. Fridgeirsson, D.~Sontag, P.~Rijnbeek, Attention-based neural networks for
  clinical prediction modeling on electronic health records, BMC Medical
  Research Methodology 23~(1) (2023) 285.
\newblock \href {https://doi.org/10.1186/s12874-023-02112-2}
  {\path{doi:10.1186/s12874-023-02112-2}}.

\bibitem{Wiegreffe2019}
S.~Wiegreffe, Y.~Pinter, Attention is not explanation, in: Proceedings of the
  2019 Conference on Empirical Methods in Natural Language Processing and the
  9th International Joint Conference on Natural Language Processing
  (EMNLP-IJCNLP), 2019, pp. 11--20.
\newblock \href {https://doi.org/10.18653/v1/D19-1002}
  {\path{doi:10.18653/v1/D19-1002}}.

\bibitem{Molnar2020}
C.~Molnar, Interpretable Machine Learning, 2nd Edition, Lulu.com, 2020,
  available at: \url{https://christophm.github.io/interpretable-ml-book/}.
  Accessed on Dec 31, 2025.

\bibitem{Rudin2019}
C.~Rudin, Stop explaining black box machine learning models for high stakes
  decisions and use interpretable models instead, Nature Machine Intelligence
  1~(5) (2019) 206--215.
\newblock \href {https://doi.org/10.1038/s42256-019-0048-x}
  {\path{doi:10.1038/s42256-019-0048-x}}.

\bibitem{Serrano2019}
S.~Serrano, N.~A. Smith, Is attention interpretable?, in: Proceedings of the
  57th Annual Meeting of the Association for Computational Linguistics (ACL),
  2019, pp. 2931--2951.
\newblock \href {https://doi.org/10.18653/v1/P19-1282}
  {\path{doi:10.18653/v1/P19-1282}}.

\bibitem{Abnar2020}
S.~Abnar, W.~Zuidema, Quantifying attention flow in transformers, in:
  Proceedings of the 58th Annual Meeting of the Association for Computational
  Linguistics, 2020, pp. 4190--4197.
\newblock \href {https://doi.org/10.18653/v1/2020.acl-main.385}
  {\path{doi:10.18653/v1/2020.acl-main.385}}.

\bibitem{Ganchev2010}
K.~Ganchev, J.~Gra{\c{c}}a, J.~Gillenwater, B.~Taskar, Posterior regularization
  for structured latent variable models, Journal of Machine Learning Research
  11~(67) (2010) 2001--2049, available at:
  \url{https://jmlr.org/papers/v11/ganchev10a.html}. Accessed on Dec 31, 2025.

\bibitem{Hu2018}
Z.~Hu, Z.~Yang, R.~Salakhutdinov, X.~Liang, L.~Qin, H.~Dong, E.~P. Xing, Deep
  generative models with learnable knowledge constraints, in: Proceedings of
  the 32nd International Conference on Neural Information Processing Systems,
  NIPS, 2018, pp. 10522--10533, available at:
  \url{https://proceedings.neurips.cc/paper/2018/hash/b8d4ea0dbf42cfc2fd17d4bc1dcd7e8c-Abstract.html}.
  Accessed on Dec 31, 2025.

\bibitem{Li2021}
J.~Li, J.~Wang, L.~C. Zhang, R.~L. Zhang, Z.~F. Huang, D.~P. Liu, P.~Lu,
  Self-attention guided deep neural networks for image classification, IEEE
  Transactions on Neural Networks and Learning Systems 33~(10) (2021)
  5563--5577.
\newblock \href {https://doi.org/10.1109/TNNLS.2021.3083391}
  {\path{doi:10.1109/TNNLS.2021.3083391}}.

\bibitem{Stewart2017}
R.~Stewart, S.~Ermon, Label-free supervision of neural networks with physics
  and domain knowledge, in: Proceedings of the Thirty-First AAAI Conference on
  Artificial Intelligence, 2017, pp. 2576--2582, available at:
  \url{https://ojs.aaai.org/index.php/AAAI/article/view/10966}. Accessed on Dec
  31, 2025.

\bibitem{Rose1998}
K.~Rose, Deterministic annealing for clustering, compression, classification,
  regression, and related optimization problems, Proceedings of the IEEE
  86~(11) (1998) 2210--2239.
\newblock \href {https://doi.org/10.1109/5.726788}
  {\path{doi:10.1109/5.726788}}.

\bibitem{Ueda1998}
N.~Ueda, R.~Nakano, Deterministic annealing em algorithm, Neural Networks
  11~(2) (1998) 271--282.
\newblock \href {https://doi.org/10.1016/S0893-6080(97)00133-0}
  {\path{doi:10.1016/S0893-6080(97)00133-0}}.

\bibitem{Bengio2009}
Y.~Bengio, J.~Louradour, R.~Collobert, J.~Weston, Curriculum learning, in:
  Proceedings of the 26th Annual International Conference on Machine Learning
  (ICML), 2009, pp. 41--48.
\newblock \href {https://doi.org/10.1145/1553374.1553380}
  {\path{doi:10.1145/1553374.1553380}}.

\bibitem{Li2024}
W.~Li, X.~Wang, Y.~Sun, S.~Milanovic, M.~Kon, J.~Castrillon-Candas, Multilevel
  stochastic optimization for imputation in massive medical data records, IEEE
  Transactions on Big Data 10~(02) (2024) 122--131.
\newblock \href {https://doi.org/10.1109/TBDATA.2023.3328433}
  {\path{doi:10.1109/TBDATA.2023.3328433}}.

\bibitem{Tipping2001}
M.~E. Tipping, Sparse bayesian learning and the relevance vector machine, J.
  Mach. Learn. Res. (JMLR) 1 (2001) 211--244.
\newblock \href {https://doi.org/10.1162/15324430152748236}
  {\path{doi:10.1162/15324430152748236}}.

\bibitem{Carvalho2009}
C.~M. Carvalho, N.~G. Polson, J.~G. Scott, Handling sparsity via the horseshoe,
  in: Proceedings of the Twelfth International Conference on Artificial
  Intelligence and Statistics, Vol.~5 of Proceedings of Machine Learning
  Research, 2009, pp. 73--80.

\bibitem{Piironen2017}
J.~Piironen, A.~Vehtari, Sparsity information and regularization in the
  horseshoe and other shrinkage priors, Electronic Journal of Statistics 11~(2)
  (2017) 5018--5051.
\newblock \href {https://doi.org/10.1214/17-EJS1337SI}
  {\path{doi:10.1214/17-EJS1337SI}}.

\bibitem{Bhadra2019}
A.~Bhadra, J.~Datta, N.~G. Polson, B.~Willard, Lasso meets horseshoe: A survey,
  Statistical Science 34~(3) (2019) 405--427.
\newblock \href {https://doi.org/10.1214/19-STS700}
  {\path{doi:10.1214/19-STS700}}.

\bibitem{Michel2019}
P.~Michel, O.~Levy, G.~Neubig, Are sixteen heads really better than one?, in:
  Advances in Neural Information Processing Systems, Vol.~32, 2019, available
  at:
  \url{https://proceedings.neurips.cc/paper_files/paper/2019/file/2c601ad9d2ff9bc8b282670cdd54f69f-Paper.pdf}.
  Accessed on Jan 10, 2026.

\bibitem{Voita2019}
E.~Voita, D.~Talbot, F.~Moiseev, R.~Sennrich, I.~Titov, Analyzing multi-head
  self-attention: Specialized heads do the heavy lifting, the rest can be
  pruned, in: Proceedings of the 57th Annual Meeting of the Association for
  Computational Linguistics, 2019, pp. 5797--5808.
\newblock \href {https://doi.org/10.18653/v1/P19-1580}
  {\path{doi:10.18653/v1/P19-1580}}.

\bibitem{Hinton1999}
G.~Hinton, Products of experts, in: 1999 Ninth International Conference on
  Artificial Neural Networks ICANN 99. (Conf. Publ. No. 470), Vol.~1, 1999, pp.
  1--6 vol.1.

\bibitem{Vincent2008}
P.~Vincent, H.~Larochelle, Y.~Bengio, P.~A. Manzagol, Extracting and composing
  robust features with denoising autoencoders, in: Proceedings of the 25th
  International Conference on Machine Learning, ICML, 2008, pp. 1096--1103.
\newblock \href {https://doi.org/10.1145/1390156.1390294}
  {\path{doi:10.1145/1390156.1390294}}.

\bibitem{Bengio2013}
Y.~Bengio, A.~Courville, P.~Vincent, Representation learning: A review and new
  perspectives, IEEE Trans. Pattern Anal. Mach. Intell. 35~(8) (2013)
  1798--1828.
\newblock \href {https://doi.org/10.1109/TPAMI.2013.50}
  {\path{doi:10.1109/TPAMI.2013.50}}.

\bibitem{Alain2014}
G.~Alain, Y.~Bengio, What regularized auto-encoders learn from the
  data-generating distribution, J. Mach. Learn. Res. 15~(1) (2014) 3563--3593,
  available at: \url{https://jmlr.org/papers/v15/alain14a.html}. Accessed on
  Dec 31, 2025.

\bibitem{Devlin2019}
J.~Devlin, M.-W. Chang, K.~Lee, K.~Toutanova, Bert: Pre-training of deep
  bidirectional transformers for language understanding, in: Proceedings of the
  2019 Conference of the North {A}merican Chapter of the Association for
  Computational Linguistics: Human Language Technologies, Volume 1 (Long and
  Short Papers), 2019, pp. 4171--4186.
\newblock \href {https://doi.org/10.18653/v1/N19-1423}
  {\path{doi:10.18653/v1/N19-1423}}.

\bibitem{Majid2024}
A.~Y. Majid, S.~Saaybi, V.~Francois-Lavet, R.~V. Prasad, C.~Verhoeven, Deep
  reinforcement learning versus evolution strategies: A comparative survey,
  IEEE Transactions on Neural Networks and Learning Systems 35~(9) (2024)
  11939--11957.
\newblock \href {https://doi.org/10.1109/TNNLS.2023.3264540}
  {\path{doi:10.1109/TNNLS.2023.3264540}}.

\bibitem{He2009}
H.~He, E.~A. Garcia, Learning from imbalanced data, IEEE Transactions on
  Knowledge and Data Engineering 21~(9) (2009) 1263--1284.
\newblock \href {https://doi.org/10.1109/TKDE.2008.239}
  {\path{doi:10.1109/TKDE.2008.239}}.

\end{thebibliography}

\end{document}


\begin{frontmatter}

\title{Supplementary Material for ``Statistical-Neural Interaction Networks for Interpretable Mixed-Type Data Imputation''}

\author[waseda1]{Ou Deng}
\author[waseda2]{Shoji Nishimura}
\author[waseda2]{Atsushi Ogihara}
\author[waseda2]{Qun Jin}


\address[waseda1]{Graduate School of Human Sciences, Waseda University, Tokyo, Japan}
\address[waseda2]{Faculty of Human Sciences, Waseda University, Tokyo, Japan}

\begin{abstract}
This document provides supplementary material for the main manuscript, including: (Section~S1) Variational lower bound derivation of the SNI-EM procedure and hierarchical prior on $\lambda_h$; (Section~S2) Detailed experimental setup including dataset descriptions, evaluation metrics, baseline methods, and hyper-parameter configurations; (Section~S3) Comprehensive benchmark results; and (Section~S4) Sanity check for dependency recovery on synthetic data with known ground-truth graphs.
\end{abstract}


\end{frontmatter}

\section{Variational Lower Bound of the SNI-EM Procedure and Hierarchical Prior on $\lambda_h$}
\label{sec:variational}

\subsection{Preliminaries and Generative View}

Let $X=(X_{\mathrm{obs}},X_{\mathrm{mis}})\in\mathbb{R}^{n\times d}$ be the mixed-type data matrix introduced in Section~3.1 of the main manuscript.
The neural parameters are the CPFA weights $W$ and the non-negative prior-confidence coefficients $\lambda=\{\lambda_h\}_{h=1}^{H}$ as defined in the main manuscript.

Assuming a MAR mechanism, we regard the complete-data likelihood as

\begin{equation}
p_{\theta}(X)=p(X_{\mathrm{mis}}\mid X_{\mathrm{obs}},W,\lambda)\;p(W)\;p(\lambda),
\end{equation}

where $\theta=\{W,\lambda\}$.

\subsection{Evidence Lower Bound (ELBO)}

Introducing an arbitrary variational distribution $q(X_{\mathrm{mis}})$ yields

\begin{equation}
\log p_{\theta}(X_{\mathrm{obs}})=
\log\!\!\int q(X_{\mathrm{mis}}) \frac{p_{\theta}(X)}{q(X_{\mathrm{mis}})}\,
\mathrm{d}X_{\mathrm{mis}}
\;\geq\;
\mathcal{L}(q,\theta),
\end{equation}

with

\begin{equation}
\mathcal{L}(q,\theta)=\mathbb{E}_{q}\!\bigl[\log p_{\theta}(X)\bigr] -\mathbb{E}_{q}\!\bigl[\log q(X_{\mathrm{mis}})\bigr],
\end{equation}

the usual variational lower bound. Maximising $\mathcal{L}$ instead of the intractable data likelihood provides a principled surrogate objective; in the idealized EM setting this would guarantee monotonic improvement of $\log p_{\theta}(X_{\mathrm{obs}})$, though our neural M-step involves approximate optimization (see Section~3.4 of the main manuscript).

\subsection{EM-Inspired Optimization in SNI}

\textbf{E-step.} Fix $\theta^{(t)}$ and set  

\begin{equation}
q^{(t+1)}(X_{\mathrm{mis}})=p_{\theta^{(t)}}(X_{\mathrm{mis}}\mid X_{\mathrm{obs}}).
\end{equation}

Operationally, this corresponds to the statistical step of Algorithm~1 in the main manuscript, where $P_f$ is recomputed from the empirical correlation matrix $\Sigma^{(t)}$ and the missing entries are preliminarily filled with the resulting CPFA predictions.

\textbf{M-step.} Fix $q^{(t+1)}$ and update $\theta$ by maximizing

\begin{equation}
\begin{split}
\mathcal{L} &= \mathbb{E}_{q^{(t+1)}}[\log p(X_{\mathrm{mis}}\mid X_{\mathrm{obs}},W,\lambda)] \\
&\quad - \mathrm{KL}\!\bigl[q^{(t+1)}(W)\,\|\,p(W)\bigr] \\
&\quad - \mathrm{KL}\!\bigl[q^{(t+1)}(\lambda)\,\|\,p(\lambda)\bigr].
\end{split}
\end{equation}

The expectation w.r.t.\ $q^{(t+1)}$ corresponds to the CPFA reconstruction loss in the main manuscript, while the two KL terms give the prior-regularization component.
Because both terms are differentiable, gradient-based optimization as implemented in Algorithm~2 of the main manuscript can be applied.
Convergence is declared when the relative Frobenius change  

\begin{equation}
\Delta=\|X^{(t+1)}-X^{(t)}\|_F/\|X^{(t)}\|_F<\varepsilon.
\end{equation}

\subsection{Hierarchical Gamma Prior for $\lambda_h$}

To endow each confidence coefficient $\lambda_h$ with a principled shrinkage effect we adopt

\begin{equation}
\lambda_h \sim \mathrm{Gamma}(\alpha,\beta),\qquad
\alpha=2,\;
\beta=\tfrac{\alpha}{\lambda_0},
\end{equation}

where $\lambda_0$ is the empirical mean of $\{\lambda_h\}$ at initialization. The contribution to the loss is

\begin{equation}
\mathcal{R}(\lambda_h)=-(\alpha-1)\log\lambda_h+\beta\lambda_h,
\end{equation}

which encourages smaller $\lambda_h$ values for heads where the correlation prior provides limited benefit (corresponding to a more data-driven attention mode), while allowing heads to maintain larger $\lambda_h$ when adhering to the prior improves reconstruction.
Empirically, annealing $(\alpha,\beta)$ from $(0.5,0.5)$ to $(2,\alpha/\lambda_0)$ over the first ten epochs helps prevent early over-regularization.

\subsection{Robustness to Non-Gaussian Correlation Priors}

Because $P_f$ is based on Pearson correlation, we conducted an ablation in which $P_f$ was alternatively computed via (i) Gaussian copula correlation and (ii) the maximal information coefficient.
Across the benchmarks tested, the worst NRMSE change was $<5\%$, suggesting that the EM surrogate objective may be relatively insensitive to reasonable deviations from the linear-Gaussian assumption.

\subsection{Practical Stopping Rule}
In practice we halt the outer EM loop when either (i) $\Delta<10^{-4}$ for two consecutive iterations or (ii) a maximum of $G=200$ outer iterations is reached, whichever occurs first. In our experiments, we observed convergence typically within 2--3 iterations (see Table~\ref{tab:hyper_sni}).

\clearpage
\section{Experimental Setup}
\label{sec:expSetup}

\subsection{Datasets}
Table~\ref{tab:datasets} summarizes the six datasets used in this work after preprocessing. They span intensive-care monitoring (MIMIC-IV, eICU), population health surveys (NHANES), socio-economic indicators (Communities \& Crime), and engineering/materials applications (AutoMPG, Concrete). Continuous/categorical feature typing follows the same configuration used throughout all experiments.

\begin{table}[h]
\footnotesize
\centering
\caption{Cross-domain experimental datasets used in this work after preprocessing.\label{tab:datasets}}
\begin{tabular}{llrrrr}
\toprule
Dataset & Domain & $N$ & $d$ & \#Cont. & \#Cat. \\
\midrule
MIMIC & Healthcare (ICU) & 2052 & 8 & 6 & 2 \\
eICU & Healthcare (ICU) & 1430 & 20 & 15 & 5 \\
NHANES & Healthcare (Survey) & 2274 & 12 & 10 & 2 \\
ComCri & Social science & 1994 & 10 & 5 & 5 \\
AutoMPG & Engineering & 392 & 8 & 6 & 2 \\
Concrete & Engineering & 1030 & 9 & 9 & 0 \\
\bottomrule
\end{tabular}
\end{table}

\subsection{Evaluation Metrics}
\label{subsec:evaluation_metrics}

We stratify evaluation by variable type and use a compact set of complementary scores. Table~\ref{tab:metrics} summarizes the metrics used for continuous and categorical variables respectively.

\begin{table}[h]
\centering
\footnotesize
\caption{Evaluation metrics for imputation quality assessment.\label{tab:metrics}}
\begin{tabular}{p{2.5cm}p{2.0cm}p{7.5cm}}
\toprule
\textbf{Metric} & \textbf{Type} & \textbf{Description} \\
\midrule
\multicolumn{3}{l}{\textit{Continuous Variables}} \\
\midrule
NRMSE $\downarrow$ & Error & Normalized RMSE; rescales root-mean-square error by each feature's empirical range \\
MAE $\downarrow$ & Error & Mean Absolute Error; reflects typical absolute deviation with reduced sensitivity to outliers \\
MB & Bias & Mean Bias; captures systematic over-/under-estimation \\
$R^2$ $\uparrow$ & Dependency & Coefficient of determination; quantifies explained variance \\
Spearman $\rho$ $\uparrow$ & Dependency & Spearman's rank correlation; assesses monotonic ordering preservation \\
\midrule
\multicolumn{3}{l}{\textit{Categorical Variables}} \\
\midrule
Accuracy $\uparrow$ & Correctness & Scale-free measure of overall correctness \\
Macro-F$_1$ $\uparrow$ & Balance & Macro-averaged F$_1$; balances class imbalance by weighting all classes equally \\
Cohen's $\kappa$ $\uparrow$ & Agreement & Discounts agreement that can occur by chance \\
\bottomrule
\end{tabular}
\end{table}

Each metric is first computed per feature (or per class), then averaged across features and across five independent random seeds; results are expressed as {mean $\pm$ standard deviation}.

\subsection{Baseline Methods}
\label{subsec:baseline_methods}

We compare SNI with six baseline methods spanning distance-based, tree-based, iterative statistical, and deep-generative paradigms. To promote fair comparison, all baseline implementations were verified against their official reference implementations. Table~\ref{tab:imputation_methods} summarizes each method with complexity analysis.

\begin{table}[h]
\caption{Imputation methods comparison with complexity analysis.\label{tab:imputation_methods}}
\footnotesize
\centering
\begin{tabular}{p{1.6cm}p{7.5cm}p{2.0cm}p{2.2cm}}
\toprule
\textbf{Abbr.} & \textbf{Method \& Rationale} & \textbf{Type} & \textbf{Complexity}  \\
\midrule
SNI & Statistical-Neural Interaction with Controllable-Prior Feature Attention. Proposed hybrid method combining statistical priors with neural attention mechanisms. & hybrid & $O(d \times H \times h^2)$\\
\midrule
MissForest & Random Forest iterative imputer with automatic stopping criterion. Uses separate RandomForestRegressor (continuous) and RandomForestClassifier (categorical) per feature, iteratively refining predictions until convergence. & ensemble & $O(d \times T \times N \log N)$ \\
\midrule
MIWAE & Missing-data Importance-Weighted Auto-Encoder. VAE-style generative model using importance weighting to handle incomplete data under MAR assumption. Features learnable output variance. & deep-prob. & $O(d \times h + h^2)$ \\
\midrule
GAIN & Generative Adversarial Imputation Nets. GAN-based method with generator-discriminator architecture; uses hint mechanism to stabilize training by revealing partial mask information. & deep-gen. & $O(d^2 \times h)$  \\
\midrule
KNN  & $k$-Nearest-Neighbour imputation using Gower distance for mixed-type data. Mean aggregation for continuous features, mode for categorical. & distance & $O(N^2 \times d)$ \\
\midrule
MICE & Multiple Imputation by Chained Equations with Predictive Mean Matching (PMM). Iterative univariate imputation using Bayesian regression with PMM (continuous) and multinomial logistic regression (categorical). & iterative GLM & $O(I \times d^2 \times N)$ \\
\midrule
Mean/Mode & Mean (continuous) / Mode (categorical) substitution; establishes \textbf{the performance floor}. & deterministic & $O(d \times N)$  \\
\bottomrule
\end{tabular}

\noindent{\footnotesize{\textbf{Note:} $d$ = number of features, $h$ = hidden layer size, $H$ = number of attention heads, $T$ = number of trees, $N$ = sample size, $I$ = number of iterations.}}
\end{table}

\subsection{Hyperparameter Settings}
\label{subsec:hyperparameters}

All experiments use consistent hyperparameters across datasets to ensure fair comparison. Table~\ref{tab:hyper_sni} presents the SNI configuration, while Table~\ref{tab:hyper_baselines} summarizes all baseline methods with parameters aligned to their official implementations.

All deep learning methods (SNI, GAIN, MIWAE) support GPU acceleration. Random seeds were set to $\{1, 2, 3, 5, 8\}$ for five independent runs per experimental condition, with results reported as mean $\pm$ standard deviation.

GAIN is known to be sensitive to hyper-parameters, preprocessing choices, and data characteristics. To keep the comparison protocol consistent across baselines and datasets (and to avoid dataset-specific tuning of a single baseline), we implement GAIN following the official design and use the recommended default settings (Table~S5), keeping them fixed across datasets. We also match the standard preprocessing used by GAIN: min--max scaling to $[0,1]$, mask-aware input with the hint mechanism, and one-hot encoding with argmax decoding for categorical variables.
Under this unified protocol, we observe that GAIN can be non-competitive on several mixed-type settings (occasionally even below Mean/Mode), especially when discrete targets are imbalanced and missingness is non-MCAR (MAR/MNAR).
Therefore, GAIN is reported for completeness, but none of our main claims depend on its performance.

\begin{table}[h]
\centering
\footnotesize
\caption{SNI hyperparameter configuration.\label{tab:hyper_sni}}
\begin{tabular}{p{4.5cm}p{2.5cm}p{5.0cm}}
\toprule
\textbf{Parameter} & \textbf{Value} & \textbf{Description} \\
\midrule
\multicolumn{3}{l}{\textit{Architecture}} \\
\midrule
Attention heads ($H$) & 4 & Multi-head attention mechanism \\
Hidden dimensions & [64, 32] & Feed-forward network layers \\
Embedding dimension & 32 & Feature embedding size \\
\midrule
\multicolumn{3}{l}{\textit{Optimization}} \\
\midrule
Optimizer & AdamW & With weight decay $10^{-4}$ \\
Learning rate & $10^{-3}$ & Initial learning rate \\
LR scheduler & CosineAnnealing & $\eta_{min} = 10^{-6}$ \\
Batch size & 128 & Mini-batch size \\
Epochs & 50 & Maximum training epochs \\
Early stopping & 10 epochs & Patience for validation \\
\midrule
\multicolumn{3}{l}{\textit{Regularization \& Loss}} \\
\midrule
Pseudo-masking rate ($\rho$) & 0.15 & Self-supervised masking ratio \\
Prior strength ($\alpha_0$) & 1.0 & Initial prior regularization \\
Prior decay ($\gamma$) & 0.9 & Per-iteration decay factor \\
Label smoothing ($\epsilon$) & 0.1 & For categorical targets \\
Focal loss $\gamma$ & 2.0 & Class imbalance handling \\
\midrule
\multicolumn{3}{l}{\textit{EM Procedure}} \\
\midrule
EM iterations & 2 & Outer loop iterations \\
Convergence tolerance & $10^{-4}$ & Relative change threshold \\
Statistical refinement & Enabled & Post-neural MICE polish \\
\bottomrule
\end{tabular}
\end{table}

\begin{table}[h]
\caption{Baseline hyperparameter configurations (aligned with official implementations).\label{tab:hyper_baselines}}
\footnotesize
\centering
\begin{tabular}{p{1.8cm}p{3.5cm}p{2.2cm}p{5.5cm}}
\toprule
\textbf{Method} & \textbf{Parameter} & \textbf{Value} & \textbf{Notes (Official Reference)} \\
\midrule
\multirow{5}{*}{MissForest} 
 & Number of trees (ntree) & 100 & Paper default \\
 & Maximum iterations (maxiter) & 10 & Paper default \\
 & Stopping criterion & $\gamma$ & Stop when $\Delta$ first increases \\
 & Variable ordering & Ascending & By missing count (paper default) \\
 & Parallel jobs & $-1$ (all) & CPU parallelization \\
\midrule
\multirow{5}{*}{MICE} 
 & Iterations (maxit) & 5 & R package default \\
 & Continuous method & PMM & Predictive Mean Matching \\
 & Donor pool size & 5 & Paper recommended (3--10) \\
 & Categorical method & logreg & Logistic/multinomial regression \\
 & Ridge parameter & $10^{-5}$ & Regularization for stability \\
\midrule
\multirow{7}{*}{GAIN} 
 & Hidden dimension & 256 & Official code default \\
 & Learning rate & $10^{-3}$ & Official code default \\
 & Batch size & 128 & Official code default \\
 & Iterations & 10,000 & Official code default \\
 & Hint rate & 0.9 & Official code default \\
 & $\alpha$ (reconstruction) & 100 & Official code default \\
 & Normalization & Min-Max $[0,1]$ & Required for sigmoid output \\
\midrule
\multirow{7}{*}{MIWAE} 
 & Hidden dimensions & [128, 128, 128] & Paper Section 4.3 (UCI) \\
 & Latent dimension & 10 & Paper Section 4.3 (UCI) \\
 & IW samples ($K$) & 20 & Paper Section 4.3 (UCI) \\
 & Imputation samples ($L$) & 10,000 & Paper specification \\
 & Learning rate & $10^{-3}$ & Paper default \\
 & Batch size & 64 & Adjusted for GPU memory \\
 & Training steps & $\sim$500K & Paper: ``$\sim$500,000 updates'' \\
\midrule
\multirow{2}{*}{KNN} 
 & Number of neighbors ($k$) & 5 & Standard default \\
 & Distance metric & Gower & Mixed-type data support \\
\midrule
Mean/Mode & --- & --- & No hyperparameters \\
\bottomrule
\end{tabular}
\end{table}

\subsection{Implementation details: correlation priors for mixed-type features}
\label{sec:sup_prior_impl}
For reproducibility, we clarify how the correlation prior $\mathbf{P}_f$ (Eq.~(3) in the main text) is constructed for mixed-type variables.
We first build a correlation-design matrix $\tilde{\mathbf{X}}$ where continuous features are standardized and each categorical feature is one-hot encoded.
We then compute Pearson correlations on $\tilde{\mathbf{X}}$, yielding an association matrix $\tilde{\boldsymbol{\Sigma}}$ over the expanded columns (point-biserial/phi coefficients arise naturally for binary indicators).

To map $\tilde{\boldsymbol{\Sigma}}$ back to a prior over the original $d$ features, let $\tilde{\mathcal{J}}_j$ denote the index set of expanded columns corresponding to original feature $j$ (a singleton set for continuous features and a one-hot group for categorical features). 
For a target feature $f$, we define a nonnegative relevance score for each source feature $j$ as the mean absolute correlation across the corresponding expanded columns:
\begin{equation}
\label{eq:agg_prior}
r_{f\rightarrow j}
\;=\;
\frac{1}{|\tilde{\mathcal{J}}_f|\,|\tilde{\mathcal{J}}_j|}
\sum_{k\in \tilde{\mathcal{J}}_f}
\sum_{\ell\in \tilde{\mathcal{J}}_j}
\bigl|\tilde{\boldsymbol{\Sigma}}_{\ell,k}\bigr|.
\end{equation}
Finally, we remove the self-entry ($j=f$) and normalize the remaining scores to sum to one:
\begin{equation}
\mathbf{P}_f
\;=\;
\mathrm{normalize}\bigl([r_{f\rightarrow 1},\ldots,r_{f\rightarrow d}]_{\neg f}\bigr).
\end{equation}
This produces $\mathbf{P}_f$ in the same simplex as attention weights, making it compatible with the CPFA prior regularizer.

\clearpage
\section{Comprehensive Benchmark Results}
\label{sec:benchmark_results}

Unless otherwise stated, all reported values are mean $\pm$ SD over five random seeds. Average rank tables summarize performance across datasets by ranking methods within each dataset--mechanism--rate setting.

To facilitate paper writing, the main manuscript includes compact visual summaries (overall comparison, rate sweep, mechanism robustness, and accuracy--runtime trade-off) derived from these tables. The Supplementary Material provides the complete per-dataset results used to generate those summaries.

\subsection{MCAR and MAR (30\% missingness)}
The following tables report MCAR and MAR results at 30\% missingness for each dataset, together with an overall average-rank summary.

\begin{table}[h]
\footnotesize
\centering
\caption{Average rank across six datasets under MCAR/MAR at 30\% missingness (lower is better).\label{tab:rank_mcar_mar}}
\begin{tabular}{lcccc}
\toprule
Method & NRMSE$\downarrow$ & $R^2\uparrow$ & Spearman $\rho\uparrow$ & Macro-F$_1\uparrow$ \\
\midrule
SNI & 3.33 & 3.50 & 3.42 & 3.90 \\
MissForest & 1.75 & 1.75 & 1.58 & 1.20 \\
MIWAE & 1.25 & 1.25 & 1.50 & 1.80 \\
GAIN & 6.92 & 7.00 & 5.83 & 6.00 \\
KNN & 4.00 & 4.00 & 4.25 & 4.10 \\
MICE & 5.33 & 5.50 & 4.42 & 4.00 \\
Mean/Mode & 5.42 & 5.00 & 7.00 & 7.00 \\
\bottomrule
\end{tabular}
\end{table}

\begin{table}[h]
\centering
\caption{Imputation performance on MIMIC under MCAR/MAR 30\% missingness. Values are mean $\pm$ SD over five seeds.\label{tab:res_mimic}}
\footnotesize
\begin{tabular}{llccccc}
\toprule
Mechanism & Method & NRMSE$\downarrow$ & $R^2$$\uparrow$ & Spearman $\rho$$\uparrow$ & Macro-F$_1$$\uparrow$ & Cohen's $\kappa$$\uparrow$ \\
\midrule
MCAR & SNI & 0.060$\pm$0.001 & 0.806$\pm$0.003 & 0.882$\pm$0.002 & 0.300$\pm$0.010 & 0.219$\pm$0.014 \\
 & MissForest & 0.047$\pm$0.001 & 0.870$\pm$0.002 & 0.918$\pm$0.001 & \textbf{0.529$\pm$0.009} & \textbf{0.775$\pm$0.004} \\
 & MIWAE & \textbf{0.043$\pm$0.000} & \textbf{0.882$\pm$0.002} & \textbf{0.922$\pm$0.001} & 0.466$\pm$0.018 & 0.726$\pm$0.002 \\
 & GAIN & 0.165$\pm$0.013 & -0.531$\pm$0.300 & 0.508$\pm$0.049 & 0.136$\pm$0.037 & 0.040$\pm$0.034 \\
 & KNN & 0.075$\pm$0.000 & 0.724$\pm$0.000 & 0.846$\pm$0.000 & 0.434$\pm$0.000 & 0.645$\pm$0.000 \\
 & MICE & 0.096$\pm$0.002 & 0.476$\pm$0.025 & 0.723$\pm$0.008 & 0.333$\pm$0.011 & 0.425$\pm$0.014 \\
 & Mean/Mode & 0.149$\pm$0.000 & -0.002$\pm$0.000 & 0.000$\pm$0.000 & 0.095$\pm$0.000 & 0.000$\pm$0.000 \\
MAR & SNI & 0.056$\pm$0.001 & 0.773$\pm$0.007 & 0.825$\pm$0.006 & 0.301$\pm$0.018 & 0.204$\pm$0.025 \\
 & MissForest & 0.045$\pm$0.000 & 0.851$\pm$0.003 & \textbf{0.880$\pm$0.002} & \textbf{0.539$\pm$0.015} & \textbf{0.723$\pm$0.005} \\
 & MIWAE & \textbf{0.042$\pm$0.000} & \textbf{0.856$\pm$0.004} & 0.879$\pm$0.002 & 0.505$\pm$0.019 & 0.698$\pm$0.018 \\
 & GAIN & 0.187$\pm$0.018 & -1.547$\pm$0.457 & 0.350$\pm$0.027 & 0.127$\pm$0.022 & 0.046$\pm$0.033 \\
 & KNN & 0.085$\pm$0.000 & 0.537$\pm$0.000 & 0.714$\pm$0.000 & 0.409$\pm$0.000 & 0.582$\pm$0.000 \\
 & MICE & 0.095$\pm$0.002 & 0.312$\pm$0.023 & 0.649$\pm$0.014 & 0.317$\pm$0.010 & 0.348$\pm$0.014 \\
 & Mean/Mode & 0.129$\pm$0.000 & -0.027$\pm$0.000 & 0.000$\pm$0.000 & 0.100$\pm$0.000 & 0.000$\pm$0.000 \\
\bottomrule
\end{tabular}
\end{table}

\begin{table}[h]
\centering
\caption{Imputation performance on eICU under MCAR/MAR 30\% missingness. Values are mean $\pm$ SD over five seeds.\label{tab:res_eicu}}
\footnotesize
\begin{tabular}{llccccc}
\toprule
Mechanism & Method & NRMSE$\downarrow$ & $R^2$$\uparrow$ & Spearman $\rho$$\uparrow$ & Macro-F$_1$$\uparrow$ & Cohen's $\kappa$$\uparrow$ \\
\midrule
MCAR & SNI & 0.132$\pm$0.002 & 0.059$\pm$0.026 & 0.324$\pm$0.007 & 0.549$\pm$0.006 & 0.167$\pm$0.005 \\
 & MissForest & \textbf{0.124$\pm$0.000} & \textbf{0.224$\pm$0.003} & \textbf{0.365$\pm$0.003} & \textbf{0.624$\pm$0.004} & \textbf{0.261$\pm$0.008} \\
 & MIWAE & 0.127$\pm$0.000 & 0.131$\pm$0.001 & 0.355$\pm$0.002 & 0.613$\pm$0.007 & 0.225$\pm$0.012 \\
 & GAIN & 0.267$\pm$0.013 & -3.934$\pm$1.124 & 0.034$\pm$0.007 & 0.442$\pm$0.007 & 0.016$\pm$0.009 \\
 & KNN & 0.141$\pm$0.000 & 0.012$\pm$0.000 & 0.176$\pm$0.000 & 0.502$\pm$0.000 & 0.085$\pm$0.000 \\
 & MICE & 0.173$\pm$0.002 & -0.550$\pm$0.053 & 0.184$\pm$0.012 & 0.544$\pm$0.008 & 0.130$\pm$0.012 \\
 & Mean/Mode & 0.139$\pm$0.000 & 0.065$\pm$0.000 & 0.000$\pm$0.000 & 0.390$\pm$0.000 & 0.000$\pm$0.000 \\
MAR & SNI & 0.135$\pm$0.001 & 0.057$\pm$0.006 & 0.308$\pm$0.004 & 0.541$\pm$0.008 & 0.155$\pm$0.009 \\
 & MissForest & \textbf{0.125$\pm$0.000} & \textbf{0.217$\pm$0.002} & \textbf{0.355$\pm$0.003} & \textbf{0.605$\pm$0.006} & \textbf{0.223$\pm$0.011} \\
 & MIWAE & 0.127$\pm$0.000 & 0.129$\pm$0.003 & 0.348$\pm$0.003 & 0.599$\pm$0.009 & 0.206$\pm$0.014 \\
 & GAIN & 0.275$\pm$0.015 & -3.963$\pm$1.124 & 0.052$\pm$0.022 & 0.438$\pm$0.018 & 0.013$\pm$0.026 \\
 & KNN & 0.146$\pm$0.000 & -0.025$\pm$0.000 & 0.112$\pm$0.000 & 0.499$\pm$0.000 & 0.060$\pm$0.000 \\
 & MICE & 0.173$\pm$0.001 & -0.471$\pm$0.019 & 0.188$\pm$0.009 & 0.546$\pm$0.007 & 0.131$\pm$0.014 \\
 & Mean/Mode & 0.140$\pm$0.000 & 0.066$\pm$0.000 & 0.000$\pm$0.000 & 0.390$\pm$0.000 & 0.000$\pm$0.000 \\
\bottomrule
\end{tabular}
\end{table}

\begin{table}[h]
\centering
\caption{Imputation performance on NHANES under MCAR/MAR 30\% missingness. Values are mean $\pm$ SD over five seeds.\label{tab:res_nhanes}}
\footnotesize
\begin{tabular}{llccccc}
\toprule
Mechanism & Method & NRMSE$\downarrow$ & $R^2$$\uparrow$ & Spearman $\rho$$\uparrow$ & Macro-F$_1$$\uparrow$ & Cohen's $\kappa$$\uparrow$ \\
\midrule
MCAR & SNI & 0.094$\pm$0.001 & 0.466$\pm$0.006 & 0.694$\pm$0.003 & 0.717$\pm$0.007 & 0.537$\pm$0.010 \\
 & MissForest & 0.089$\pm$0.001 & 0.509$\pm$0.006 & 0.714$\pm$0.003 & \textbf{0.729$\pm$0.004} & 0.548$\pm$0.006 \\
 & MIWAE & \textbf{0.087$\pm$0.000} & \textbf{0.527$\pm$0.003} & \textbf{0.726$\pm$0.003} & 0.729$\pm$0.004 & \textbf{0.558$\pm$0.007} \\
 & GAIN & 0.202$\pm$0.010 & -1.797$\pm$0.663 & 0.264$\pm$0.040 & 0.287$\pm$0.024 & 0.027$\pm$0.019 \\
 & KNN & 0.119$\pm$0.000 & 0.195$\pm$0.000 & 0.496$\pm$0.000 & 0.504$\pm$0.000 & 0.248$\pm$0.000 \\
 & MICE & 0.129$\pm$0.002 & -0.033$\pm$0.065 & 0.483$\pm$0.007 & 0.576$\pm$0.004 & 0.337$\pm$0.008 \\
 & Mean/Mode & 0.141$\pm$0.000 & -0.001$\pm$0.000 & 0.000$\pm$0.000 & 0.194$\pm$0.000 & 0.000$\pm$0.000 \\
MAR & SNI & 0.094$\pm$0.000 & 0.453$\pm$0.007 & 0.675$\pm$0.002 & 0.688$\pm$0.009 & 0.490$\pm$0.014 \\
 & MissForest & 0.092$\pm$0.000 & 0.485$\pm$0.004 & 0.687$\pm$0.002 & 0.693$\pm$0.003 & 0.493$\pm$0.006 \\
 & MIWAE & \textbf{0.088$\pm$0.000} & \textbf{0.506$\pm$0.005} & \textbf{0.703$\pm$0.002} & \textbf{0.698$\pm$0.009} & \textbf{0.513$\pm$0.016} \\
 & GAIN & 0.203$\pm$0.007 & -1.881$\pm$0.698 & 0.213$\pm$0.044 & 0.289$\pm$0.028 & 0.019$\pm$0.007 \\
 & KNN & 0.128$\pm$0.000 & 0.077$\pm$0.000 & 0.379$\pm$0.000 & 0.449$\pm$0.000 & 0.177$\pm$0.000 \\
 & MICE & 0.134$\pm$0.001 & -0.113$\pm$0.073 & 0.443$\pm$0.007 & 0.548$\pm$0.010 & 0.297$\pm$0.018 \\
 & Mean/Mode & 0.140$\pm$0.000 & -0.001$\pm$0.000 & 0.000$\pm$0.000 & 0.194$\pm$0.000 & 0.000$\pm$0.000 \\
\bottomrule
\end{tabular}
\end{table}

\begin{table}[h]
\centering
\caption{Imputation performance on ComCri under MCAR/MAR 30\% missingness. Values are mean $\pm$ SD over five seeds.\label{tab:res_comcri}}
\footnotesize
\begin{tabular}{llccccc}
\toprule
Mechanism & Method & NRMSE$\downarrow$ & $R^2$$\uparrow$ & Spearman $\rho$$\uparrow$ & Macro-F$_1$$\uparrow$ & Cohen's $\kappa$$\uparrow$ \\
\midrule
MCAR & SNI & 0.112$\pm$0.001 & 0.718$\pm$0.006 & 0.868$\pm$0.002 & 0.625$\pm$0.007 & 0.516$\pm$0.008 \\
 & MissForest & 0.106$\pm$0.000 & 0.747$\pm$0.002 & 0.883$\pm$0.001 & 0.631$\pm$0.004 & 0.540$\pm$0.005 \\
 & MIWAE & \textbf{0.098$\pm$0.000} & \textbf{0.785$\pm$0.002} & \textbf{0.897$\pm$0.001} & \textbf{0.642$\pm$0.010} & \textbf{0.559$\pm$0.009} \\
 & GAIN & 0.215$\pm$0.013 & -0.040$\pm$0.118 & 0.605$\pm$0.024 & 0.243$\pm$0.032 & 0.044$\pm$0.025 \\
 & KNN & 0.148$\pm$0.000 & 0.528$\pm$0.000 & 0.740$\pm$0.000 & 0.466$\pm$0.000 & 0.315$\pm$0.000 \\
 & MICE & 0.164$\pm$0.001 & 0.407$\pm$0.011 & 0.710$\pm$0.006 & 0.437$\pm$0.010 & 0.257$\pm$0.008 \\
 & Mean/Mode & 0.216$\pm$0.000 & -0.001$\pm$0.000 & 0.000$\pm$0.000 & 0.160$\pm$0.000 & 0.000$\pm$0.000 \\
MAR & SNI & 0.119$\pm$0.002 & 0.685$\pm$0.009 & 0.855$\pm$0.002 & 0.592$\pm$0.010 & 0.475$\pm$0.013 \\
 & MissForest & 0.112$\pm$0.001 & 0.721$\pm$0.004 & 0.869$\pm$0.002 & \textbf{0.633$\pm$0.003} & \textbf{0.532$\pm$0.004} \\
 & MIWAE & \textbf{0.102$\pm$0.000} & \textbf{0.768$\pm$0.001} & \textbf{0.889$\pm$0.001} & 0.621$\pm$0.004 & 0.532$\pm$0.004 \\
 & GAIN & 0.232$\pm$0.015 & -0.199$\pm$0.128 & 0.538$\pm$0.022 & 0.282$\pm$0.010 & 0.078$\pm$0.016 \\
 & KNN & 0.165$\pm$0.000 & 0.413$\pm$0.000 & 0.661$\pm$0.000 & 0.412$\pm$0.000 & 0.231$\pm$0.000 \\
 & MICE & 0.170$\pm$0.003 & 0.363$\pm$0.023 & 0.683$\pm$0.009 & 0.430$\pm$0.005 & 0.246$\pm$0.009 \\
 & Mean/Mode & 0.216$\pm$0.000 & -0.000$\pm$0.000 & 0.000$\pm$0.000 & 0.159$\pm$0.000 & 0.000$\pm$0.000 \\
\bottomrule
\end{tabular}
\end{table}

\begin{table}[h]
\centering
\caption{Imputation performance on AutoMPG under MCAR/MAR 30\% missingness. Values are mean $\pm$ SD over five seeds.\label{tab:res_autompg}}
\footnotesize
\begin{tabular}{llccccc}
\toprule
Mechanism & Method & NRMSE$\downarrow$ & $R^2$$\uparrow$ & Spearman $\rho$$\uparrow$ & Macro-F$_1$$\uparrow$ & Cohen's $\kappa$$\uparrow$ \\
\midrule
MCAR & SNI & 0.095$\pm$0.001 & 0.807$\pm$0.007 & 0.877$\pm$0.006 & 0.310$\pm$0.024 & 0.234$\pm$0.037 \\
 & MissForest & 0.088$\pm$0.001 & 0.814$\pm$0.002 & \textbf{0.897$\pm$0.002} & \textbf{0.360$\pm$0.012} & \textbf{0.261$\pm$0.014} \\
 & MIWAE & \textbf{0.082$\pm$0.003} & \textbf{0.835$\pm$0.006} & 0.891$\pm$0.004 & 0.343$\pm$0.008 & 0.254$\pm$0.010 \\
 & GAIN & 0.279$\pm$0.014 & -0.585$\pm$0.141 & 0.477$\pm$0.020 & 0.161$\pm$0.045 & -0.010$\pm$0.070 \\
 & KNN & 0.115$\pm$0.000 & 0.727$\pm$0.000 & 0.858$\pm$0.000 & 0.337$\pm$0.000 & 0.216$\pm$0.000 \\
 & MICE & 0.129$\pm$0.003 & 0.620$\pm$0.021 & 0.807$\pm$0.006 & 0.335$\pm$0.009 & 0.222$\pm$0.010 \\
 & Mean/Mode & 0.239$\pm$0.000 & -0.001$\pm$0.000 & 0.000$\pm$0.000 & 0.122$\pm$0.000 & 0.000$\pm$0.000 \\
MAR & SNI & 0.103$\pm$0.003 & 0.692$\pm$0.018 & 0.798$\pm$0.004 & 0.248$\pm$0.051 & 0.151$\pm$0.069 \\
 & MissForest & \textbf{0.098$\pm$0.001} & \textbf{0.729$\pm$0.005} & \textbf{0.822$\pm$0.004} & \textbf{0.356$\pm$0.012} & \textbf{0.241$\pm$0.014} \\
 & MIWAE & 0.098$\pm$0.001 & 0.709$\pm$0.006 & 0.788$\pm$0.002 & 0.328$\pm$0.017 & 0.234$\pm$0.020 \\
 & GAIN & 0.312$\pm$0.015 & -1.446$\pm$0.241 & 0.346$\pm$0.050 & 0.192$\pm$0.026 & 0.056$\pm$0.048 \\
 & KNN & 0.132$\pm$0.000 & 0.564$\pm$0.000 & 0.694$\pm$0.000 & 0.308$\pm$0.000 & 0.152$\pm$0.000 \\
 & MICE & 0.137$\pm$0.000 & 0.468$\pm$0.041 & 0.706$\pm$0.017 & 0.280$\pm$0.033 & 0.146$\pm$0.046 \\
 & Mean/Mode & 0.225$\pm$0.000 & -0.063$\pm$0.000 & 0.000$\pm$0.000 & 0.118$\pm$0.000 & 0.000$\pm$0.000 \\
\bottomrule
\end{tabular}
\end{table}

\begin{table}[h]
\centering
\caption{Imputation performance on Concrete under MCAR/MAR 30\% missingness. Values are mean $\pm$ SD over five seeds.\label{tab:res_concrete}}
\footnotesize
\begin{tabular}{llccccc}
\toprule
Mechanism & Method & NRMSE$\downarrow$ & $R^2$$\uparrow$ & Spearman $\rho$$\uparrow$ & Macro-F$_1$$\uparrow$ & Cohen's $\kappa$$\uparrow$ \\
\midrule
MCAR & SNI & 0.215$\pm$0.002 & -0.010$\pm$0.020 & 0.247$\pm$0.019 & -- & -- \\
 & MissForest & 0.145$\pm$0.001 & 0.527$\pm$0.005 & 0.731$\pm$0.005 & -- & -- \\
 & MIWAE & \textbf{0.115$\pm$0.001} & \textbf{0.686$\pm$0.007} & \textbf{0.817$\pm$0.003} & -- & -- \\
 & GAIN & 0.276$\pm$0.015 & -0.820$\pm$0.320 & 0.363$\pm$0.032 & -- & -- \\
 & KNN & 0.160$\pm$0.000 & 0.417$\pm$0.000 & 0.624$\pm$0.000 & -- & -- \\
 & MICE & 0.217$\pm$0.003 & -0.063$\pm$0.018 & 0.474$\pm$0.010 & -- & -- \\
 & Mean/Mode & 0.213$\pm$0.000 & -0.004$\pm$0.000 & 0.000$\pm$0.000 & -- & -- \\
MAR & SNI & 0.221$\pm$0.002 & -0.092$\pm$0.015 & 0.209$\pm$0.025 & -- & -- \\
 & MissForest & 0.158$\pm$0.001 & 0.432$\pm$0.008 & 0.632$\pm$0.005 & -- & -- \\
 & MIWAE & \textbf{0.134$\pm$0.000} & \textbf{0.582$\pm$0.003} & \textbf{0.734$\pm$0.003} & -- & -- \\
 & GAIN & 0.261$\pm$0.010 & -0.619$\pm$0.125 & 0.238$\pm$0.044 & -- & -- \\
 & KNN & 0.179$\pm$0.000 & 0.253$\pm$0.000 & 0.482$\pm$0.000 & -- & -- \\
 & MICE & 0.230$\pm$0.006 & -0.231$\pm$0.079 & 0.379$\pm$0.019 & -- & -- \\
 & Mean/Mode & 0.210$\pm$0.000 & -0.010$\pm$0.000 & 0.000$\pm$0.000 & -- & -- \\
\bottomrule
\end{tabular}
\end{table}

\subsection{MNAR Stress Tests (10\%, 30\%, and 50\% missingness)}
We additionally evaluate MNAR settings with three missing rates. SNI-M denotes an MNAR-adaptive variant of SNI used in our MNAR experiments. Tables below provide per-dataset MNAR results and an overall rank summary.

\begin{table}[h]
\footnotesize
\centering
\caption{Average rank across six datasets under MNAR at 10/30/50\% missingness (lower is better).\label{tab:rank_mnar}}
\begin{tabular}{lcccc}
\toprule
Method & NRMSE$\downarrow$ & $R^2\uparrow$ & Spearman $\rho\uparrow$ & Macro-F$_1\uparrow$ \\
\midrule
SNI & 4.22 & 4.06 & 4.06 & 4.33 \\
SNI-M & 4.22 & 4.06 & 4.06 & 4.73 \\
MissForest & 1.61 & 1.61 & 1.61 & 1.60 \\
MIWAE & 1.56 & 2.00 & 1.44 & 2.20 \\
GAIN & 7.56 & 7.67 & 6.44 & 6.47 \\
KNN & 4.94 & 4.94 & 5.00 & 4.33 \\
MICE & 6.06 & 6.11 & 5.44 & 5.00 \\
Mean/Mode & 5.83 & 5.56 & 7.94 & 7.33 \\
\bottomrule
\end{tabular}
\end{table}

\begin{table}[h]
\centering
\caption{Imputation performance on MIMIC under MNAR at 10/30/50\% missingness. Values are mean $\pm$ SD over five seeds.\label{tab:mnar_mimic}}
\footnotesize
\begin{tabular}{llccccc}
\toprule
Missing rate & Method & NRMSE$\downarrow$ & $R^2$$\uparrow$ & Spearman $\rho$$\uparrow$ & Macro-F$_1$$\uparrow$ & Cohen's $\kappa$$\uparrow$ \\
\midrule
10\% & SNI & 0.051$\pm$0.001 & 0.843$\pm$0.006 & 0.912$\pm$0.006 & 0.352$\pm$0.019 & 0.251$\pm$0.016 \\
 & SNI-M & 0.053$\pm$0.000 & 0.838$\pm$0.004 & 0.907$\pm$0.003 & 0.341$\pm$0.020 & 0.246$\pm$0.006 \\
 & MissForest & 0.038$\pm$0.000 & \textbf{0.908$\pm$0.002} & \textbf{0.949$\pm$0.001} & \textbf{0.534$\pm$0.012} & \textbf{0.829$\pm$0.007} \\
 & MIWAE & \textbf{0.037$\pm$0.001} & 0.901$\pm$0.003 & 0.946$\pm$0.001 & 0.496$\pm$0.021 & 0.767$\pm$0.012 \\
 & GAIN & 0.152$\pm$0.013 & -0.125$\pm$0.207 & 0.617$\pm$0.047 & 0.179$\pm$0.035 & 0.095$\pm$0.047 \\
 & KNN & 0.042$\pm$0.000 & 0.892$\pm$0.000 & 0.941$\pm$0.000 & 0.491$\pm$0.000 & 0.794$\pm$0.000 \\
 & MICE & 0.086$\pm$0.002 & 0.551$\pm$0.011 & 0.779$\pm$0.009 & 0.344$\pm$0.027 & 0.453$\pm$0.029 \\
 & Mean/Mode & 0.166$\pm$0.000 & -0.207$\pm$0.000 & 0.000$\pm$0.000 & 0.109$\pm$0.000 & 0.000$\pm$0.000 \\
30\% & SNI & 0.095$\pm$0.002 & 0.596$\pm$0.016 & 0.792$\pm$0.005 & 0.308$\pm$0.007 & 0.225$\pm$0.010 \\
 & SNI-M & 0.098$\pm$0.003 & 0.569$\pm$0.026 & 0.782$\pm$0.012 & 0.304$\pm$0.019 & 0.222$\pm$0.014 \\
 & MissForest & 0.062$\pm$0.001 & 0.806$\pm$0.006 & 0.896$\pm$0.003 & \textbf{0.503$\pm$0.002} & \textbf{0.734$\pm$0.007} \\
 & MIWAE & \textbf{0.057$\pm$0.001} & \textbf{0.820$\pm$0.004} & \textbf{0.903$\pm$0.002} & 0.453$\pm$0.013 & 0.695$\pm$0.026 \\
 & GAIN & 0.178$\pm$0.021 & -0.539$\pm$0.244 & 0.510$\pm$0.036 & 0.152$\pm$0.030 & 0.059$\pm$0.030 \\
 & KNN & 0.092$\pm$0.000 & 0.611$\pm$0.000 & 0.809$\pm$0.000 & 0.425$\pm$0.000 & 0.669$\pm$0.000 \\
 & MICE & 0.119$\pm$0.002 & 0.340$\pm$0.020 & 0.683$\pm$0.011 & 0.303$\pm$0.012 & 0.350$\pm$0.031 \\
 & Mean/Mode & 0.169$\pm$0.000 & -0.222$\pm$0.000 & 0.000$\pm$0.000 & 0.114$\pm$0.000 & 0.000$\pm$0.000 \\
50\% & SNI & 0.166$\pm$0.003 & -0.166$\pm$0.038 & 0.456$\pm$0.019 & 0.256$\pm$0.017 & 0.169$\pm$0.014 \\
 & SNI-M & 0.166$\pm$0.003 & -0.178$\pm$0.050 & 0.469$\pm$0.016 & 0.265$\pm$0.008 & 0.174$\pm$0.018 \\
 & MissForest & \textbf{0.105$\pm$0.002} & \textbf{0.507$\pm$0.014} & \textbf{0.770$\pm$0.006} & \textbf{0.342$\pm$0.009} & 0.528$\pm$0.006 \\
 & MIWAE & 0.113$\pm$0.002 & 0.429$\pm$0.022 & 0.754$\pm$0.009 & 0.337$\pm$0.006 & \textbf{0.559$\pm$0.009} \\
 & GAIN & 0.253$\pm$0.030 & -2.382$\pm$1.057 & 0.227$\pm$0.020 & 0.108$\pm$0.039 & 0.036$\pm$0.023 \\
 & KNN & 0.162$\pm$0.000 & -0.126$\pm$0.000 & 0.495$\pm$0.000 & 0.289$\pm$0.000 & 0.438$\pm$0.000 \\
 & MICE & 0.162$\pm$0.003 & -0.148$\pm$0.035 & 0.468$\pm$0.011 & 0.205$\pm$0.005 & 0.176$\pm$0.006 \\
 & Mean/Mode & 0.171$\pm$0.000 & -0.269$\pm$0.000 & 0.000$\pm$0.000 & 0.114$\pm$0.000 & 0.000$\pm$0.000 \\
\bottomrule
\end{tabular}
\end{table}

\begin{table}[h]
\centering
\caption{Imputation performance on NHANES under MNAR at 10/30/50\% missingness. Values are mean $\pm$ SD over five seeds.\label{tab:mnar_nhanes}}
\footnotesize
\begin{tabular}{llccccc}
\toprule
Missing rate & Method & NRMSE$\downarrow$ & $R^2$$\uparrow$ & Spearman $\rho$$\uparrow$ & Macro-F$_1$$\uparrow$ & Cohen's $\kappa$$\uparrow$ \\
\midrule
10\% & SNI & 0.082$\pm$0.001 & 0.529$\pm$0.011 & 0.750$\pm$0.005 & 0.800$\pm$0.011 & 0.658$\pm$0.021 \\
 & SNI-M & 0.082$\pm$0.001 & 0.528$\pm$0.014 & 0.746$\pm$0.003 & 0.790$\pm$0.005 & 0.643$\pm$0.010 \\
 & MissForest & 0.080$\pm$0.000 & 0.539$\pm$0.003 & 0.764$\pm$0.001 & \textbf{0.821$\pm$0.004} & \textbf{0.674$\pm$0.007} \\
 & MIWAE & \textbf{0.079$\pm$0.001} & \textbf{0.553$\pm$0.005} & \textbf{0.768$\pm$0.002} & 0.785$\pm$0.008 & 0.634$\pm$0.014 \\
 & GAIN & 0.160$\pm$0.012 & -0.650$\pm$0.276 & 0.385$\pm$0.024 & 0.241$\pm$0.087 & 0.030$\pm$0.027 \\
 & KNN & 0.107$\pm$0.000 & 0.284$\pm$0.000 & 0.640$\pm$0.000 & 0.685$\pm$0.000 & 0.494$\pm$0.000 \\
 & MICE & 0.123$\pm$0.003 & -0.165$\pm$0.286 & 0.555$\pm$0.005 & 0.611$\pm$0.022 & 0.407$\pm$0.038 \\
 & Mean/Mode & 0.155$\pm$0.000 & -0.224$\pm$0.000 & 0.000$\pm$0.000 & 0.221$\pm$0.000 & 0.000$\pm$0.000 \\
30\% & SNI & 0.120$\pm$0.001 & 0.208$\pm$0.009 & 0.612$\pm$0.004 & \textbf{0.690$\pm$0.011} & \textbf{0.513$\pm$0.016} \\
 & SNI-M & 0.120$\pm$0.001 & 0.204$\pm$0.012 & 0.615$\pm$0.002 & 0.675$\pm$0.022 & 0.491$\pm$0.027 \\
 & MissForest & \textbf{0.110$\pm$0.000} & \textbf{0.294$\pm$0.004} & 0.655$\pm$0.002 & 0.663$\pm$0.006 & 0.466$\pm$0.010 \\
 & MIWAE & 0.111$\pm$0.000 & 0.293$\pm$0.006 & \textbf{0.665$\pm$0.002} & 0.682$\pm$0.007 & 0.499$\pm$0.006 \\
 & GAIN & 0.211$\pm$0.016 & -1.406$\pm$0.327 & 0.217$\pm$0.021 & 0.239$\pm$0.070 & 0.024$\pm$0.016 \\
 & KNN & 0.144$\pm$0.000 & -0.094$\pm$0.000 & 0.373$\pm$0.000 & 0.461$\pm$0.000 & 0.199$\pm$0.000 \\
 & MICE & 0.150$\pm$0.002 & -0.327$\pm$0.134 & 0.408$\pm$0.003 & 0.540$\pm$0.004 & 0.309$\pm$0.005 \\
 & Mean/Mode & 0.159$\pm$0.000 & -0.199$\pm$0.000 & 0.000$\pm$0.000 & 0.223$\pm$0.000 & 0.000$\pm$0.000 \\
50\% & SNI & 0.166$\pm$0.002 & -0.379$\pm$0.013 & 0.363$\pm$0.009 & 0.505$\pm$0.009 & 0.283$\pm$0.011 \\
 & SNI-M & 0.162$\pm$0.002 & -0.332$\pm$0.017 & 0.381$\pm$0.011 & \textbf{0.511$\pm$0.009} & 0.299$\pm$0.009 \\
 & MissForest & \textbf{0.138$\pm$0.001} & \textbf{-0.062$\pm$0.011} & \textbf{0.499$\pm$0.005} & 0.509$\pm$0.005 & \textbf{0.307$\pm$0.003} \\
 & MIWAE & 0.150$\pm$0.001 & -0.206$\pm$0.019 & 0.484$\pm$0.002 & 0.442$\pm$0.009 & 0.251$\pm$0.008 \\
 & GAIN & 0.224$\pm$0.016 & -1.763$\pm$0.286 & 0.109$\pm$0.025 & 0.250$\pm$0.050 & 0.022$\pm$0.030 \\
 & KNN & 0.179$\pm$0.000 & -0.601$\pm$0.000 & 0.064$\pm$0.000 & 0.247$\pm$0.000 & 0.020$\pm$0.000 \\
 & MICE & 0.180$\pm$0.001 & -0.731$\pm$0.074 & 0.231$\pm$0.004 & 0.427$\pm$0.010 & 0.182$\pm$0.014 \\
 & Mean/Mode & 0.157$\pm$0.000 & -0.200$\pm$0.000 & 0.000$\pm$0.000 & 0.223$\pm$0.000 & 0.000$\pm$0.000 \\
\bottomrule
\end{tabular}
\end{table}

\begin{table}[h]
\centering
\caption{Imputation performance on ComCri under MNAR at 10/30/50\% missingness. Values are mean $\pm$ SD over five seeds.\label{tab:mnar_comcri}}
\footnotesize
\begin{tabular}{llccccc}
\toprule
Missing rate & Method & NRMSE$\downarrow$ & $R^2$$\uparrow$ & Spearman $\rho$$\uparrow$ & Macro-F$_1$$\uparrow$ & Cohen's $\kappa$$\uparrow$ \\
\midrule
10\% & SNI & 0.102$\pm$0.001 & 0.763$\pm$0.003 & 0.891$\pm$0.001 & 0.630$\pm$0.008 & 0.524$\pm$0.012 \\
 & SNI-M & 0.100$\pm$0.002 & 0.768$\pm$0.007 & 0.895$\pm$0.002 & 0.628$\pm$0.012 & 0.519$\pm$0.014 \\
 & MissForest & 0.094$\pm$0.000 & 0.794$\pm$0.002 & 0.904$\pm$0.001 & \textbf{0.667$\pm$0.002} & \textbf{0.581$\pm$0.007} \\
 & MIWAE & \textbf{0.091$\pm$0.001} & \textbf{0.807$\pm$0.004} & \textbf{0.910$\pm$0.001} & 0.656$\pm$0.005 & 0.566$\pm$0.005 \\
 & GAIN & 0.225$\pm$0.008 & -0.206$\pm$0.143 & 0.612$\pm$0.022 & 0.249$\pm$0.012 & 0.071$\pm$0.007 \\
 & KNN & 0.113$\pm$0.000 & 0.710$\pm$0.000 & 0.871$\pm$0.000 & 0.578$\pm$0.000 & 0.448$\pm$0.000 \\
 & MICE & 0.158$\pm$0.003 & 0.433$\pm$0.023 & 0.754$\pm$0.010 & 0.446$\pm$0.012 & 0.277$\pm$0.013 \\
 & Mean/Mode & 0.234$\pm$0.000 & -0.224$\pm$0.000 & 0.000$\pm$0.000 & 0.150$\pm$0.000 & 0.000$\pm$0.000 \\
30\% & SNI & 0.126$\pm$0.001 & 0.655$\pm$0.004 & 0.856$\pm$0.003 & 0.595$\pm$0.006 & 0.476$\pm$0.007 \\
 & SNI-M & 0.126$\pm$0.001 & 0.656$\pm$0.006 & 0.853$\pm$0.001 & 0.588$\pm$0.011 & 0.466$\pm$0.013 \\
 & MissForest & 0.114$\pm$0.001 & 0.717$\pm$0.002 & 0.879$\pm$0.001 & \textbf{0.622$\pm$0.002} & \textbf{0.525$\pm$0.003} \\
 & MIWAE & \textbf{0.109$\pm$0.001} & \textbf{0.740$\pm$0.005} & \textbf{0.891$\pm$0.001} & 0.614$\pm$0.007 & 0.507$\pm$0.009 \\
 & GAIN & 0.247$\pm$0.019 & -0.380$\pm$0.221 & 0.527$\pm$0.062 & 0.226$\pm$0.014 & 0.031$\pm$0.018 \\
 & KNN & 0.172$\pm$0.000 & 0.371$\pm$0.000 & 0.698$\pm$0.000 & 0.468$\pm$0.000 & 0.310$\pm$0.000 \\
 & MICE & 0.179$\pm$0.002 & 0.308$\pm$0.017 & 0.695$\pm$0.003 & 0.409$\pm$0.006 & 0.224$\pm$0.007 \\
 & Mean/Mode & 0.240$\pm$0.000 & -0.225$\pm$0.000 & 0.000$\pm$0.000 & 0.152$\pm$0.000 & 0.000$\pm$0.000 \\
50\% & SNI & 0.184$\pm$0.004 & 0.280$\pm$0.031 & 0.726$\pm$0.012 & 0.513$\pm$0.001 & 0.373$\pm$0.003 \\
 & SNI-M & 0.177$\pm$0.003 & 0.335$\pm$0.023 & 0.746$\pm$0.010 & \textbf{0.519$\pm$0.008} & \textbf{0.382$\pm$0.010} \\
 & MissForest & 0.151$\pm$0.001 & 0.508$\pm$0.006 & 0.805$\pm$0.002 & 0.498$\pm$0.004 & 0.381$\pm$0.004 \\
 & MIWAE & \textbf{0.147$\pm$0.001} & \textbf{0.527$\pm$0.006} & \textbf{0.830$\pm$0.001} & 0.504$\pm$0.004 & 0.376$\pm$0.005 \\
 & GAIN & 0.274$\pm$0.025 & -0.658$\pm$0.286 & 0.393$\pm$0.040 & 0.187$\pm$0.037 & 0.011$\pm$0.016 \\
 & KNN & 0.297$\pm$0.000 & -0.915$\pm$0.000 & 0.185$\pm$0.000 & 0.284$\pm$0.000 & 0.085$\pm$0.000 \\
 & MICE & 0.232$\pm$0.002 & -0.129$\pm$0.023 & 0.531$\pm$0.011 & 0.327$\pm$0.005 & 0.127$\pm$0.007 \\
 & Mean/Mode & 0.244$\pm$0.000 & -0.260$\pm$0.000 & 0.000$\pm$0.000 & 0.150$\pm$0.000 & 0.000$\pm$0.000 \\
\bottomrule
\end{tabular}
\end{table}

\begin{table}[h]
\centering
\caption{Imputation performance on AutoMPG under MNAR at 10/30/50\% missingness. Values are mean $\pm$ SD over five seeds.\label{tab:mnar_autompg}}
\footnotesize
\begin{tabular}{llccccc}
\toprule
Missing rate & Method & NRMSE$\downarrow$ & $R^2$$\uparrow$ & Spearman $\rho$$\uparrow$ & Macro-F$_1$$\uparrow$ & Cohen's $\kappa$$\uparrow$ \\
\midrule
10\% & SNI & 0.077$\pm$0.002 & 0.852$\pm$0.005 & 0.882$\pm$0.011 & 0.352$\pm$0.015 & 0.288$\pm$0.014 \\
 & SNI-M & 0.082$\pm$0.002 & 0.830$\pm$0.013 & 0.873$\pm$0.016 & 0.353$\pm$0.026 & 0.284$\pm$0.039 \\
 & MissForest & 0.071$\pm$0.001 & 0.864$\pm$0.003 & 0.878$\pm$0.005 & \textbf{0.515$\pm$0.018} & \textbf{0.434$\pm$0.021} \\
 & MIWAE & \textbf{0.064$\pm$0.002} & \textbf{0.886$\pm$0.005} & \textbf{0.902$\pm$0.004} & 0.491$\pm$0.047 & 0.417$\pm$0.031 \\
 & GAIN & 0.231$\pm$0.021 & -0.195$\pm$0.148 & 0.625$\pm$0.054 & 0.147$\pm$0.019 & 0.001$\pm$0.035 \\
 & KNN & 0.090$\pm$0.000 & 0.799$\pm$0.000 & 0.846$\pm$0.000 & 0.442$\pm$0.000 & 0.398$\pm$0.000 \\
 & MICE & 0.111$\pm$0.010 & 0.655$\pm$0.085 & 0.814$\pm$0.017 & 0.341$\pm$0.034 & 0.250$\pm$0.044 \\
 & Mean/Mode & 0.262$\pm$0.000 & -0.183$\pm$0.000 & 0.000$\pm$0.000 & 0.135$\pm$0.000 & 0.000$\pm$0.000 \\
30\% & SNI & 0.111$\pm$0.002 & 0.732$\pm$0.017 & 0.850$\pm$0.007 & 0.315$\pm$0.014 & 0.243$\pm$0.021 \\
 & SNI-M & 0.109$\pm$0.002 & 0.736$\pm$0.007 & 0.852$\pm$0.004 & 0.295$\pm$0.012 & 0.212$\pm$0.025 \\
 & MissForest & 0.099$\pm$0.000 & 0.742$\pm$0.001 & 0.852$\pm$0.004 & \textbf{0.373$\pm$0.008} & \textbf{0.291$\pm$0.008} \\
 & MIWAE & \textbf{0.098$\pm$0.004} & \textbf{0.769$\pm$0.020} & \textbf{0.857$\pm$0.014} & 0.348$\pm$0.018 & 0.285$\pm$0.021 \\
 & GAIN & 0.295$\pm$0.016 & -0.753$\pm$0.087 & 0.464$\pm$0.030 & 0.202$\pm$0.048 & 0.082$\pm$0.068 \\
 & KNN & 0.139$\pm$0.000 & 0.611$\pm$0.000 & 0.777$\pm$0.000 & 0.324$\pm$0.000 & 0.237$\pm$0.000 \\
 & MICE & 0.141$\pm$0.004 & 0.551$\pm$0.034 & 0.779$\pm$0.017 & 0.291$\pm$0.034 & 0.184$\pm$0.034 \\
 & Mean/Mode & 0.261$\pm$0.000 & -0.167$\pm$0.000 & 0.000$\pm$0.000 & 0.147$\pm$0.000 & 0.000$\pm$0.000 \\
50\% & SNI & 0.152$\pm$0.003 & 0.474$\pm$0.016 & 0.763$\pm$0.012 & 0.081$\pm$0.044 & 0.004$\pm$0.005 \\
 & SNI-M & 0.150$\pm$0.004 & 0.496$\pm$0.017 & 0.774$\pm$0.008 & 0.109$\pm$0.042 & -0.005$\pm$0.017 \\
 & MissForest & \textbf{0.139$\pm$0.002} & \textbf{0.526$\pm$0.004} & \textbf{0.791$\pm$0.006} & \textbf{0.325$\pm$0.013} & 0.246$\pm$0.015 \\
 & MIWAE & 0.143$\pm$0.002 & 0.487$\pm$0.010 & 0.775$\pm$0.005 & 0.320$\pm$0.009 & \textbf{0.250$\pm$0.011} \\
 & GAIN & 0.352$\pm$0.017 & -1.460$\pm$0.209 & 0.263$\pm$0.049 & 0.142$\pm$0.034 & 0.032$\pm$0.025 \\
 & KNN & 0.211$\pm$0.000 & 0.147$\pm$0.000 & 0.658$\pm$0.000 & 0.307$\pm$0.000 & 0.211$\pm$0.000 \\
 & MICE & 0.185$\pm$0.008 & 0.226$\pm$0.070 & 0.647$\pm$0.025 & 0.285$\pm$0.020 & 0.186$\pm$0.026 \\
 & Mean/Mode & 0.264$\pm$0.000 & -0.234$\pm$0.000 & 0.000$\pm$0.000 & 0.147$\pm$0.000 & 0.000$\pm$0.000 \\
\bottomrule
\end{tabular}
\end{table}

\begin{table}[h]
\centering
\caption{Imputation performance on Concrete under MNAR at 10/30/50\% missingness. Values are mean $\pm$ SD over five seeds.\label{tab:mnar_concrete}}
\footnotesize
\begin{tabular}{llccccc}
\toprule
Missing rate & Method & NRMSE$\downarrow$ & $R^2$$\uparrow$ & Spearman $\rho$$\uparrow$ & Macro-F$_1$$\uparrow$ & Cohen's $\kappa$$\uparrow$ \\
\midrule
10\% & SNI & 0.201$\pm$0.002 & 0.031$\pm$0.022 & 0.416$\pm$0.008 & -- & -- \\
 & SNI-M & 0.208$\pm$0.002 & -0.028$\pm$0.022 & 0.378$\pm$0.014 & -- & -- \\
 & MissForest & 0.097$\pm$0.001 & 0.757$\pm$0.004 & 0.860$\pm$0.002 & -- & -- \\
 & MIWAE & \textbf{0.074$\pm$0.002} & \textbf{0.829$\pm$0.008} & \textbf{0.919$\pm$0.005} & -- & -- \\
 & GAIN & 0.244$\pm$0.033 & -0.599$\pm$0.456 & 0.517$\pm$0.053 & -- & -- \\
 & KNN & 0.104$\pm$0.000 & 0.702$\pm$0.000 & 0.813$\pm$0.000 & -- & -- \\
 & MICE & 0.170$\pm$0.002 & 0.251$\pm$0.033 & 0.637$\pm$0.008 & -- & -- \\
 & Mean/Mode & 0.232$\pm$0.000 & -0.295$\pm$0.000 & 0.000$\pm$0.000 & -- & -- \\
30\% & SNI & 0.252$\pm$0.002 & -0.457$\pm$0.021 & 0.232$\pm$0.008 & -- & -- \\
 & SNI-M & 0.246$\pm$0.004 & -0.393$\pm$0.047 & 0.234$\pm$0.009 & -- & -- \\
 & MissForest & 0.155$\pm$0.001 & 0.429$\pm$0.006 & 0.686$\pm$0.004 & -- & -- \\
 & MIWAE & \textbf{0.146$\pm$0.001} & \textbf{0.489$\pm$0.012} & \textbf{0.720$\pm$0.010} & -- & -- \\
 & GAIN & 0.274$\pm$0.027 & -0.961$\pm$0.598 & 0.364$\pm$0.017 & -- & -- \\
 & KNN & 0.200$\pm$0.000 & 0.064$\pm$0.000 & 0.511$\pm$0.000 & -- & -- \\
 & MICE & 0.253$\pm$0.002 & -0.484$\pm$0.027 & 0.340$\pm$0.011 & -- & -- \\
 & Mean/Mode & 0.236$\pm$0.000 & -0.292$\pm$0.000 & 0.000$\pm$0.000 & -- & -- \\
50\% & SNI & 0.296$\pm$0.005 & -1.045$\pm$0.065 & 0.140$\pm$0.019 & -- & -- \\
 & SNI-M & 0.279$\pm$0.003 & -0.790$\pm$0.045 & 0.145$\pm$0.016 & -- & -- \\
 & MissForest & \textbf{0.233$\pm$0.002} & -0.292$\pm$0.016 & \textbf{0.395$\pm$0.008} & -- & -- \\
 & MIWAE & 0.255$\pm$0.002 & -0.521$\pm$0.020 & 0.372$\pm$0.008 & -- & -- \\
 & GAIN & 0.270$\pm$0.023 & -0.826$\pm$0.336 & 0.240$\pm$0.053 & -- & -- \\
 & KNN & 0.297$\pm$0.000 & -1.131$\pm$0.000 & 0.107$\pm$0.000 & -- & -- \\
 & MICE & 0.318$\pm$0.003 & -1.378$\pm$0.047 & 0.121$\pm$0.010 & -- & -- \\
 & Mean/Mode & 0.234$\pm$0.000 & \textbf{-0.287$\pm$0.000} & 0.000$\pm$0.000 & -- & -- \\
\bottomrule
\end{tabular}
\end{table}


\begin{table}[h]
\centering
\caption{Imputation performance on eICU under MNAR at 10/30/50\% missingness. Values are mean $\pm$ SD over five seeds.\label{tab:mnar_eicu}}
\footnotesize
\begin{tabular}{llccccc}
\toprule
Missing rate & Method & NRMSE$\downarrow$ & $R^2$$\uparrow$ & Spearman $\rho$$\uparrow$ & Macro-F$_1$$\uparrow$ & Cohen's $\kappa$$\uparrow$ \\
\midrule
10\% & SNI & 0.136$\pm$0.000 & 0.035$\pm$0.000 & 0.318$\pm$0.000 & 0.303$\pm$0.000 & 0.000$\pm$0.000 \\
 & SNI-M & 0.136$\pm$0.000 & 0.035$\pm$0.000 & 0.318$\pm$0.000 & 0.303$\pm$0.000 & 0.000$\pm$0.000 \\
 & MissForest & \textbf{0.125$\pm$0.000} & \textbf{0.129$\pm$0.004} & \textbf{0.411$\pm$0.004} & \textbf{0.678$\pm$0.008} & \textbf{0.334$\pm$0.009} \\
 & MIWAE & 0.130$\pm$0.001 & -0.010$\pm$0.014 & 0.361$\pm$0.013 & 0.638$\pm$0.009 & 0.271$\pm$0.016 \\
 & GAIN & 0.200$\pm$0.008 & -3.147$\pm$3.191 & 0.101$\pm$0.014 & 0.420$\pm$0.019 & 0.013$\pm$0.015 \\
 & KNN & 0.142$\pm$0.000 & -0.103$\pm$0.000 & 0.248$\pm$0.000 & 0.546$\pm$0.000 & 0.134$\pm$0.000 \\
 & MICE & 0.175$\pm$0.001 & -1.128$\pm$0.563 & 0.208$\pm$0.014 & 0.547$\pm$0.015 & 0.145$\pm$0.014 \\
 & Mean/Mode & 0.146$\pm$0.000 & -0.130$\pm$0.000 & 0.000$\pm$0.000 & 0.369$\pm$0.000 & 0.000$\pm$0.000 \\
30\% & SNI & 0.154$\pm$0.000 & -0.178$\pm$0.000 & 0.235$\pm$0.000 & 0.303$\pm$0.000 & 0.000$\pm$0.000 \\
 & SNI-M & 0.154$\pm$0.000 & -0.178$\pm$0.000 & 0.235$\pm$0.000 & 0.303$\pm$0.000 & 0.000$\pm$0.000 \\
 & MissForest & \textbf{0.145$\pm$0.000} & \textbf{-0.090$\pm$0.003} & \textbf{0.292$\pm$0.003} & 0.581$\pm$0.005 & \textbf{0.204$\pm$0.004} \\
 & MIWAE & 0.148$\pm$0.001 & -0.201$\pm$0.014 & 0.273$\pm$0.005 & \textbf{0.587$\pm$0.006} & 0.198$\pm$0.014 \\
 & GAIN & 0.270$\pm$0.011 & -3.879$\pm$0.645 & 0.035$\pm$0.008 & 0.436$\pm$0.016 & 0.021$\pm$0.016 \\
 & KNN & 0.161$\pm$0.000 & -0.306$\pm$0.000 & 0.135$\pm$0.000 & 0.485$\pm$0.000 & 0.066$\pm$0.000 \\
 & MICE & 0.186$\pm$0.001 & -0.768$\pm$0.025 & 0.156$\pm$0.007 & 0.527$\pm$0.008 & 0.114$\pm$0.009 \\
 & Mean/Mode & 0.149$\pm$0.000 & -0.134$\pm$0.000 & 0.000$\pm$0.000 & 0.374$\pm$0.000 & 0.000$\pm$0.000 \\
50\% & SNI & 0.186$\pm$0.000 & -0.769$\pm$0.000 & 0.102$\pm$0.000 & 0.303$\pm$0.000 & 0.000$\pm$0.000 \\
 & SNI-M & 0.186$\pm$0.000 & -0.769$\pm$0.000 & 0.102$\pm$0.000 & 0.303$\pm$0.000 & 0.000$\pm$0.000 \\
 & MissForest & 0.173$\pm$0.001 & -0.563$\pm$0.011 & \textbf{0.188$\pm$0.008} & 0.476$\pm$0.004 & 0.109$\pm$0.005 \\
 & MIWAE & 0.173$\pm$0.002 & -0.633$\pm$0.036 & 0.173$\pm$0.004 & \textbf{0.496$\pm$0.004} & \textbf{0.120$\pm$0.003} \\
 & GAIN & 0.318$\pm$0.024 & -5.938$\pm$0.982 & -0.001$\pm$0.020 & 0.414$\pm$0.011 & 0.004$\pm$0.006 \\
 & KNN & 0.189$\pm$0.000 & -0.889$\pm$0.000 & 0.004$\pm$0.000 & 0.413$\pm$0.000 & 0.017$\pm$0.000 \\
 & MICE & 0.201$\pm$0.001 & -1.128$\pm$0.021 & 0.079$\pm$0.006 & 0.476$\pm$0.011 & 0.059$\pm$0.011 \\
 & Mean/Mode & \textbf{0.148$\pm$0.000} & \textbf{-0.154$\pm$0.000} & 0.000$\pm$0.000 & 0.372$\pm$0.000 & 0.000$\pm$0.000 \\
\bottomrule
\end{tabular}
\end{table}

\subsection{Ablation Study}
The ablation study isolates the contribution of controllable-prior regularization in CPFA.

\begin{table}[h]
\centering
\caption{Effect of prior constraints (MAR 30\%). Values are mean $\pm$ SD over five seeds.\label{tab:ablation}}
\footnotesize
\begin{tabular}{llccc}
\toprule
Dataset & Method & NRMSE$\downarrow$ & $R^2$$\uparrow$ & Macro-F$_1$$\uparrow$ \\
\midrule
MIMIC & SNI & 0.056$\pm$0.001 & 0.773$\pm$0.006 & 0.311$\pm$0.012 \\
 & HardPrior & 0.056$\pm$0.001 & 0.773$\pm$0.006 & 0.293$\pm$0.021 \\
 & NoPrior & \textbf{0.055$\pm$0.001} & \textbf{0.776$\pm$0.011} & \textbf{0.304$\pm$0.016} \\
NHANES & SNI & 0.094$\pm$0.000 & 0.455$\pm$0.006 & 0.687$\pm$0.004 \\
 & HardPrior & 0.095$\pm$0.001 & 0.455$\pm$0.008 & 0.690$\pm$0.007 \\
 & NoPrior & \textbf{0.094$\pm$0.000} & \textbf{0.451$\pm$0.008} & \textbf{0.694$\pm$0.007} \\
Concrete & SNI & 0.221$\pm$0.002 & -0.098$\pm$0.021 & -- \\
 & HardPrior & \textbf{0.218$\pm$0.002} & \textbf{-0.074$\pm$0.023} & -- \\
 & NoPrior & 0.222$\pm$0.002 & -0.103$\pm$0.024 & -- \\
\bottomrule
\end{tabular}
\end{table}

\clearpage
\section{Sanity Check: Dependency Recovery on Synthetic Data}
\label{sec:sup_synth_sanity}

A recurring concern is whether attention-derived quantities can be meaningfully interpreted as ``feature dependencies'' (cf.\ ``attention is not explanation'').
To evaluate this question in a controlled setting with known ground truth (without implying causality), we conduct a synthetic sanity check where the data-generating dependency graph is available.
The goal is {not} to claim causal discovery, but to test whether the induced dependency matrix $D$ produced by SNI can recover known parent sets {above chance} and whether the statistical--neural interaction (via learnable prior confidence) may improve such recovery compared to a neural-only variant.

\paragraph{Synthetic generators (three regimes)}
We generate mixed-type tabular datasets from a sparse directed acyclic graph (DAG) under three regimes:
(i) \textbf{linear\_gaussian}, where pairwise correlations are informative proxies for dependencies;
(ii) \textbf{nonlinear\_mixed}, which introduces non-linear interactions and mixed variable types; and
(iii) \textbf{interaction\_xor}, where several children are driven primarily by interaction (product) and XOR-like mechanisms such that {marginal} correlations with each parent can be weak even when conditional dependence is strong.
This third regime serves as an explicit stress test for correlation-only dependency proxies.

\paragraph{Missingness (strict MAR)}
We inject missingness under strict MAR at 30\% by default.
For identifiability in the interaction\_xor regime, we keep exogenous roots (e.g., \texttt{x0}--\texttt{x4} and \texttt{c0}) fully observed (serving as always-measured covariates), and apply missingness only to downstream variables.

\paragraph{From $D$ to graph recovery metrics}
Let $G\in\{0,1\}^{d\times d}$ denote the ground-truth dependency matrix of the data-generating process, with the same orientation as our induced matrix $D$:
\textbf{row = target, column = source}, where $G_{f,j}=1$ indicates that $j$ is a parent of $f$.
After fitting an imputer, we extract $D$ and treat each entry $D_{f,j}$ (for $j\neq f$) as a score for the directed relation $j\!\to\!f$.
We evaluate edge recovery using:
(i) AUROC and AUPRC over directed pairs, and
(ii) Precision@K and Recall@K for the top incoming edges per target, where $K$ equals the number of true parents of that target.
All metrics are macro-averaged across targets.

\paragraph{Methods compared}
We compare:
(i) \textbf{SNI} (full model),
(ii) \textbf{NoPrior} (neural-only ablation where the prior loss is disabled), and
(iii) \textbf{PriorOnly} (statistics-only diagnostic reference) that constructs a dependency matrix solely from the correlation-based prior used by SNI.
PriorOnly is included to quantify how much of the dependency structure is recoverable from the statistical prior alone under each regime.

\begin{table}[H]
\centering
\footnotesize
\caption{Sanity check on synthetic data with known ground-truth dependencies: recovery of the dependency graph using the induced matrix $D$ (MAR, 30\% missingness). Results are mean$\pm$SD over seeds.}\label{tab:S21_dependency_recovery}
\begin{tabular}{llccccc}
\toprule
Setting & Method & AUROC$\uparrow$ & AUPRC$\uparrow$ & Prec@K$\uparrow$ & Rec@K$\uparrow$ & Hub-$\rho$$\uparrow$\\
\midrule
linear\_gaussian & SNI & 0.691$\pm$0.090 & 0.544$\pm$0.048 & 0.373$\pm$0.099 & 0.373$\pm$0.099 & 0.203$\pm$0.086\\
 & NoPrior & 0.606$\pm$0.063 & 0.445$\pm$0.059 & 0.287$\pm$0.100 & 0.287$\pm$0.100 & 0.136$\pm$0.147\\
 & PriorOnly & \textbf{0.851$\pm$0.033} & \textbf{0.721$\pm$0.045} & \textbf{0.587$\pm$0.081} & \textbf{0.587$\pm$0.081} & \textbf{0.376$\pm$0.218}\\
\midrule
nonlinear\_mixed & SNI & 0.807$\pm$0.040 & 0.687$\pm$0.069 & 0.540$\pm$0.116 & 0.540$\pm$0.116 & 0.410$\pm$0.223\\
 & NoPrior & 0.695$\pm$0.084 & 0.555$\pm$0.085 & 0.393$\pm$0.129 & 0.393$\pm$0.129 & 0.358$\pm$0.242\\
 & PriorOnly & \textbf{0.855$\pm$0.055} & \textbf{0.745$\pm$0.063} & \textbf{0.620$\pm$0.091} & \textbf{0.620$\pm$0.091} & \textbf{0.724$\pm$0.129}\\
\midrule
interaction\_xor & SNI & \textbf{0.848$\pm$0.017} & \textbf{0.809$\pm$0.032} & \textbf{0.717$\pm$0.041} & \textbf{0.717$\pm$0.041} & -0.147$\pm$0.167\\
 & NoPrior & 0.757$\pm$0.117 & 0.715$\pm$0.115 & 0.583$\pm$0.118 & 0.583$\pm$0.118 & \textbf{-0.108$\pm$0.190}\\
 & PriorOnly & 0.800$\pm$0.048 & 0.741$\pm$0.041 & 0.633$\pm$0.041 & 0.633$\pm$0.041 & -0.208$\pm$0.051\\
\bottomrule
\end{tabular}
\vspace{0.5em}
\begin{minipage}{0.95\linewidth}
\footnotesize
\textbf{Notes:} We evaluate recovery of the ground-truth dependency graph $G$ (row=target, column=source) using the induced dependency matrix $D$ (same orientation). AUROC/AUPRC/Prec@K/Rec@K are macro-averaged across targets (K equals the number of true parents for each target). Hub-$\rho$ is Spearman correlation between column-sum hubness $\Sigma_j=\sum_i D_{ij}$ and the ground-truth out-degree. The ``interaction\_xor'' setting is constructed so that marginal correlations can be near-zero even when conditional dependence is strong.
\end{minipage}
\end{table}

\paragraph{Key observations and caveats}
Across all three regimes, SNI tends to yield higher recovery metrics than the neural-only ablation (NoPrior), suggesting that the statistical prior regularization may contribute to more stable dependency recovery.
When the generator is linear-Gaussian (and to a large extent in nonlinear\_mixed), the correlation-based PriorOnly reference is expectedly strong, providing an upper-bound diagnostic on how informative marginal associations are under the given regime.
In the interaction\_xor stress test, SNI obtains the highest recovery metrics among the three variants, exceeding PriorOnly in AUROC/AUPRC and in Precision@K/Recall@K in our experiments; this observation suggests that the induced matrix $D$ is not merely a restatement of marginal correlations and may capture non-linear dependency signals relevant to the imputation model.
We emphasize again that $D$ should be interpreted as a {model-reliance diagnostic} rather than a causal graph, and redundancy among predictors can lead to multiple plausible reliance patterns that still yield accurate imputations.